\documentclass[11pt]{article}

\usepackage[preprint]{acl}

\usepackage{times}
\usepackage{latexsym}
\usepackage{subcaption}
\usepackage{longtable}
\usepackage{booktabs}
\usepackage{tabularx}
\usepackage{pdflscape}
\usepackage{caption}
\usepackage{adjustbox}
\usepackage{multirow}
\usepackage[table]{xcolor}
\usepackage{graphicx}   % for \rotatebox
\usepackage{placeins}
\usepackage{pdfpages}
\usepackage{fontawesome5}

\usepackage[T1]{fontenc}
\usepackage[utf8]{inputenc}

\usepackage{microtype}

\usepackage{inconsolata}

\usepackage{graphicx}

\usepackage{titletoc}
\usepackage{float}

\title{Translation as a Scalable Proxy for Multilingual Evaluation}

\author{
 \textbf{Sheriff Issaka\textsuperscript{1}},
 \textbf{Erick Rosas Gonzalez\textsuperscript{1}}\footnotemark[1],
 \textbf{Lieqi Liu\textsuperscript{1}}\thanks{Equal contribution.},
  \textbf{Evans Kofi Agyei\textsuperscript{2}},\\
 \textbf{Lucas Bandarkar\textsuperscript{1}},
 \textbf{Nanyun Peng\textsuperscript{1}},
 \textbf{David Ifeoluwa Adelani\textsuperscript{3}},\\
 \textbf{Francisco Guzmán\textsuperscript{4}},
 \textbf{Saadia Gabriel\textsuperscript{1}} \\
 \\
 \textsuperscript{1}University of California, Los Angeles,
 \textsuperscript{2}African Languages Lab,\\
 \textsuperscript{3}Mila - Quebec AI Institute, McGill University \& Canada CIFAR AI Chair,
 \textsuperscript{4}Handshake AI\\
 \small{
   \textbf{Correspondence:} \href{mailto:sheriff@cs.ucla.edu}{sheriff@cs.ucla.edu}
 }
}

\begin{document}
\maketitle

\begin{abstract}
The rapid proliferation of LLMs has created a critical evaluation paradox: while LLMs claim multilingual proficiency, comprehensive non-machine-translated benchmarks exist for fewer than 30 languages, leaving $>$98\% of the world's 7,000 languages in an empirical void. Traditional benchmark construction faces scaling challenges such as cost, scarcity of domain experts, and data contamination. We evaluate the validity of a simpler alternative: \textit{can translation quality alone indicate a model's broader multilingual capabilities}? Through systematic evaluation of 14 models (1B-72B parameters) across 9 diverse benchmarks and 7 translation metrics, we find that translation performance is a good indicator of downstream task success (e.g., Phi-4, median Pearson $r$: \textsc{MetricX} = 0.89, \textsc{xCOMET} = 0.91, \textsc{SSA-COMET} = 0.87). These results suggest that the representational abilities supporting faithful translation overlap with those required for multilingual understanding. Translation quality, thus emerges as a strong, inexpensive first-pass proxy of multilingual performance, enabling a \textit{translation-first screening} with targeted follow-up for specific tasks. \footnote{To promote accessibility, we provide translations of this abstract in 10 different languages in Appendix ~\ref{app:translations}, generated using \href{https://all-lab-portal.com/}{Mansa} and \href{https://openai.com/gpt-5/}{ChatGPT 5}.} \footnote{ {\faGithub} Code \& {\includegraphics[height=1em]{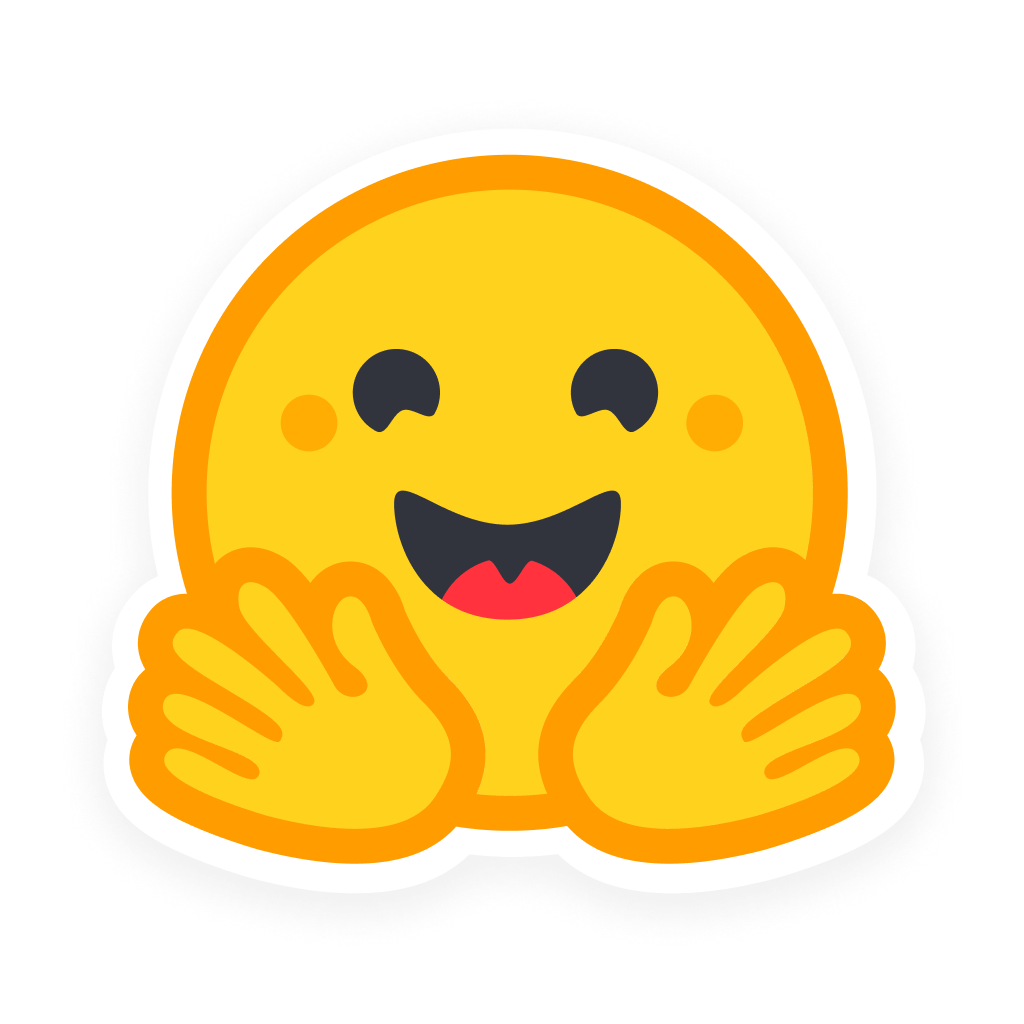}} Data: \url{https://translation-as-multilingual-proxy.github.io/}}

\end{abstract}

\section{Introduction}
The rapid advancement of large language models (LLMs) has created a major technological paradox: while these models are deployed globally and claim multilingual capabilities \cite{ouyang2022, openai2024gpt4technicalreport, touvron2023llamaopenefficientfoundation, workshop2023bloom176bparameteropenaccessmultilingual, xue2021mt5massivelymultilingualpretrained}, we lack the basic infrastructure to assess their abilities. Of the world's 7,000 languages, comprehensive non-machine-translated benchmarks exist for fewer than 30 overlapping languages \cite{choenni2024evaluationpracticesmultilingualnlp, hada-etal-2024-large, adelani-etal-2025-irokobench, singh-etal-2025-global, cobbe2021gsm8k, lewis-etal-2020-mlqa, zellers-etal-2019-hellaswag, lin-etal-2022-truthfulqa}, leaving more than 98\% of linguistic diversity unmeasured. This evaluation vacuum has profound consequences as entire language communities are excluded from AI development, not because their languages are inherently unsuitable for modeling, but because we simply cannot measure whether models work for them at all \cite{zhu2024multilinguallargelanguagemodels, Xu_2025, chang2023surveyevaluationlargelanguage, issaka2025africanlanguageslabcollaborative}. For these low-resourced languages (LRLs), evaluations suffer from a dual failure: they are frequently nonexistent, and when present, systematically overestimate model capabilities \cite{ahuja2022staticmodelstestsets, choenni2024evaluationpracticesmultilingualnlp}.

As a response, this work explores a simple but overlooked question:\begin{quote} \textit{Can translation quality serve as a reliable, scalable, and cost-effective proxy for a model's broader multilingual capabilities?} \end{quote}

The current paradigm of building task-specific benchmarks for each language has proven economically costly, inherently unscalable, and systematically biased \cite{li2024cmmlumeasuringmassivemultitask, yuksel-etal-2024-turkishmmlu, adelani-etal-2025-irokobench}. A recent analysis of over 2,000 benchmarks revealed that despite investments exceeding tens of millions of dollars between 2021 and 2024, coverage remains overwhelmingly concentrated on English and a handful of high-resource languages \cite{wu2025bitterlessonlearned2000}. Even well-intentioned efforts to automatically translate existing benchmarks fail systematically, as these machine translations cannot capture cultural and linguistic nuances, resulting in evaluations that correlate poorly with human judgments and perpetuate biases \cite{wu2025bitterlessonlearned2000, romanou2024includeevaluatingmultilinguallanguage, ramezani2023knowledgeculturalmoralnorms, singh2024aya, aakanksha2024multilingualalignmentprismaligning}.

\begin{figure*}[t!]
    \centering
    \includegraphics[width=\textwidth]{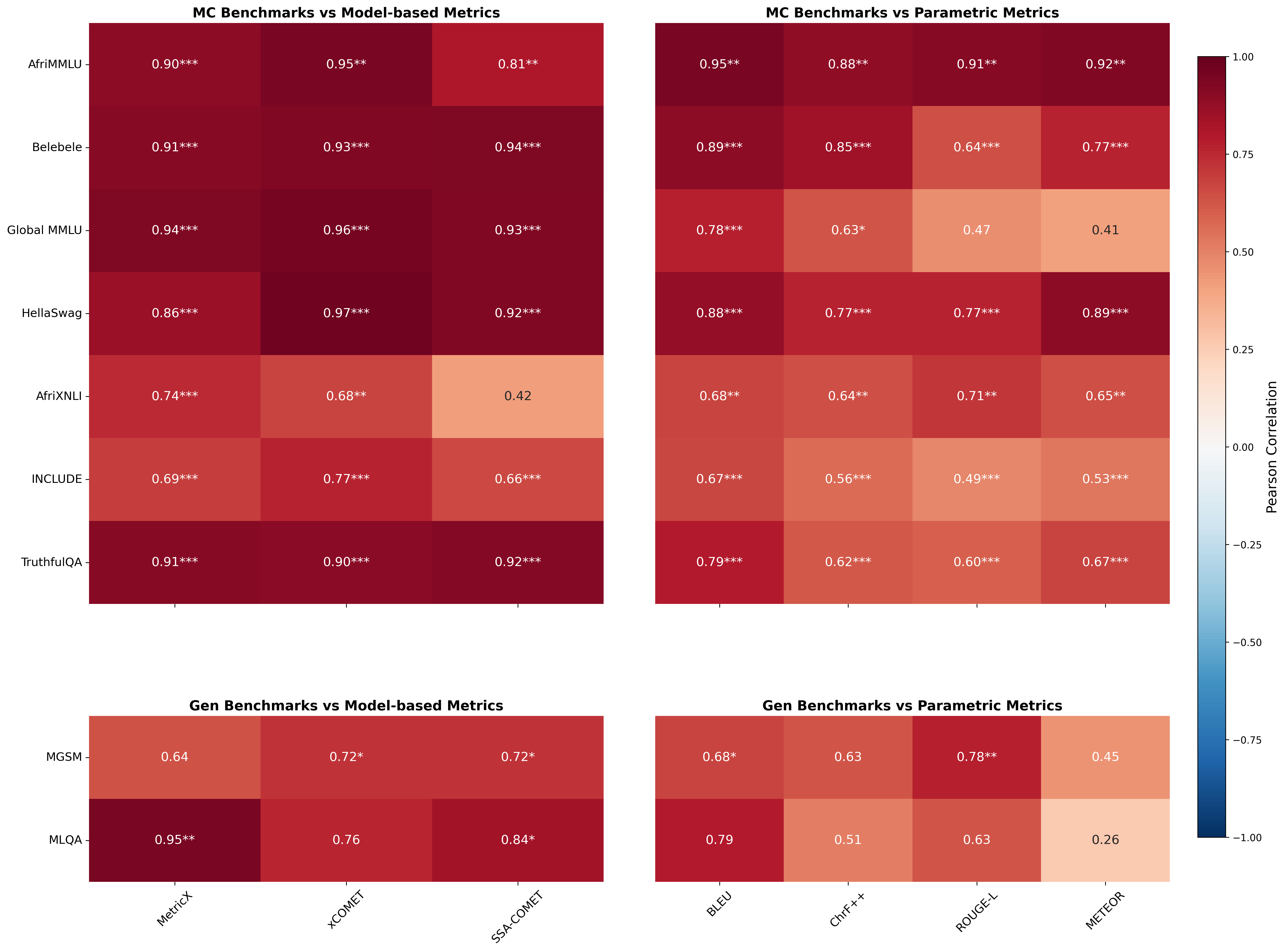}
    \caption{Pearson correlation matrix for the Phi-4 (14B) model, showing correlations between MT quality metrics and multilingual benchmark performance on \textsc{FLORES-200}. Color intensity indicates the strength of correlation, with darker red denoting stronger positive correlations.}
    \label{fig:flagship}
\end{figure*}

We hypothesize that high-quality, human-verified parallel corpora could provide a foundation for evaluating general multilingual capabilities. This claim rests on a central insight: translation is one of the most demanding natural language tasks as it requires bridging non-isomorphic linguistic systems while preserving meaning across semantics, pragmatics, discourse structure, and cultural context \cite{info16090723, liu2025culturally}.

To test this hypothesis, we conduct the most comprehensive correlation analysis to date as described in \S3. We examine 14 models spanning 5 distinct families (Llama \cite{llama3modelcard}, DeepSeek \cite{deepseekai2025deepseekr1}, Qwen \cite{yang2025qwen3technicalreport, qwen2025qwen25technicalreport}, Gemma \cite{gemmateam2025gemma3technicalreport}, Phi \cite{abdin2024phi4technicalreport}), ranging from 1B to 72B parameters. We evaluate their performance across 9 diverse multilingual benchmarks (\textsc{TruthfulQA} \cite{lai-etal-2023-okapi}, \textsc{MLQA} \cite{lewis-etal-2020-mlqa}, \textsc{GlobalMMLU} \cite{singh-etal-2025-global}, \textsc{Belebele} \cite{bandarkar-etal-2024-belebele}, \textsc{MGSM} \cite{shi2022language}, \textsc{AfriMMLU} \cite{adelani-etal-2025-irokobench}, \textsc{HellaSwag} \cite{lai-etal-2023-okapi}, \textsc{AfriXNLI} \cite{adelani-etal-2025-irokobench},  \textsc{INCLUDE} \cite{romanou2024include}) that span the full spectrum of language understanding: knowledge retrieval, logical reasoning, reading comprehension, and natural language inference. These benchmark scores are then systematically correlated with 7 distinct machine translation (MT) quality metrics across 3 translation datasets (\textsc{Flores-200} \cite{nllb2022}, \textsc{WMT24++} \cite{deutsch-etal-2025-wmt24}, \textsc{NTREX} \cite{federmann-etal-2022-ntrex,enevoldsen2025mmtebmassivemultilingualtext, muennighoff2022mteb}), enabling us to examine whether translation quality consistently correlates with performance across different task types, model architectures, and language families.

Our findings in \S4 demonstrate alignment with this hypothesis: translation quality correlates well with downstream task performance across most language-task combinations, yielding a median Pearson of $r$ = 0.89 for \textsc{MetricX}, 0.91 for \textsc{xCOMET}, and 0.87 for \textsc{SSA-COMET}. This suggests that the neural representations required for high-quality translation may be intertwined with those needed for complex reasoning. Translation quality, therefore, offers a low-cost first-pass signal of multilingual performance. While it cannot replace task-specific benchmarks entirely, it enables a practical two-stage evaluation pipeline where translation metrics provide rapid assessment across thousands of languages, followed by targeted benchmark evaluation only where needed.

\section{Related Work}
LLMs have changed natural language processing (NLP) and the space of AI \cite{brown2020languagemodelsfewshotlearners, ouyang2022traininglanguagemodelsfollow, bai2022constitutionalaiharmlessnessai, openai2024gpt4technicalreport, geminiteam2025geminifamilyhighlycapable}.
Consequently, the rapid development of multilingual LLMs has created a significant challenge: how can we reliably and efficiently evaluate their capabilities across a diverse linguistic landscape? The academic community has largely pursued the path of building comprehensive, task-specific benchmarks that aim for exhaustive testing.

\subsection{Multilingual Benchmark}
Recent research has adopted two approaches to multilingual LLM evaluation. The first involved translating existing English evaluation suites into other languages using either human translators or MT systems \cite{shi2022languagemodelsmultilingualchainofthought, lai-etal-2023-okapi, bandarkar-etal-2024-belebele, singh-etal-2025-global, hupkes2025multilokomultilinguallocalknowledge}. Works such as \textsc{XNLI} extended natural language inference to 15 languages, establishing a paradigm that was adopted by benchmarks for question answering (\textsc{MLQA}), commonsense reasoning (\textsc{XCOPA}), and mathematical reasoning (\textsc{MGSM})  \citep{conneau-etal-2018-xnli, lewis-etal-2020-mlqa, ponti-etal-2020-xcopa, shi2022language}.

To address the limitations and potential biases of this translation-centric approach, a second line focuses on creating more culturally-centric and linguistically nuanced benchmarks curated directly in target languages. For example, studies collected human exam questions from specific regions in the target language with examples in Chinese, Indonesian, and Turkish \cite{koto-etal-2023-large, li-etal-2024-cmmlu, yuksel-etal-2024-turkishmmlu}.

Other efforts expanded multitask evaluation with a focus on low-resourced regions, as seen in \textsc{Global MMLU} and \textsc{IrokoBench} \citep{singh-etal-2025-global, adelani-etal-2025-irokobench}. Other studies emphasize the inclusion of culturally specific content in their evaluation, such as \textsc{CulturalBench} \cite{chiu2025culturalbenchrobustdiversechallenging}, \textsc{ArtELingo-28} \cite{mohamed-etal-2024-culture}, and \textsc{CVQA} \cite{NEURIPS2024_1568882b}. 

While these benchmarks provide a thorough assessment of a model's multilingual capabilities, their primary limitation is the difficulty in creating them. They are computationally and financially expensive, time-consuming to build and run, making frequent or large-scale evaluation prohibitive. Thus, the (left-behind) language communities that will need such evaluations the most are those least likely to have the said resources to create such benchmarks \citep{issaka2024ghanaiannlplandscapelook}.

\subsection{Machine Translation Metrics}
The field of MT metrics has evolved significantly from its reliance on lexical overlap methods like \textsc{BLEU} and \textsc{ChrF++}, which often fail to capture semantic accuracy  \citep{papineni-etal-2002-bleu, popovic-2017-chrf}. The current state of the art is dominated by embedding neural metrics trained on human judgments. Models like \textsc{COMET} and \textsc{MetricX} leverage powerful encoders to score translations based on semantic similarity rather than surface-level n-gram matching \citep{rei-etal-2020-comet, juraska-etal-2024-metricx}. This has made translation evaluation on broad coverage datasets like \textsc{FLORES-200} \citep{nllb2022} an attractive proxy for a model's general cross-lingual understanding. In fact, many large-scale evaluation suites now include translation as a core task along with reasoning and knowledge assessments \citep{zhang2025pmmevalparallelmultilingualmultitask}. 

\section{Methods}
\subsection{Evaluation Design}
Our study investigates the relationship between MT quality and multilingual task performance. To this end, we evaluate translation outputs of LLMs and compare them against benchmark performance across diverse categories. The evaluation framework consists of three components: (1) MT for direct measurement of cross-lingual quality, (2) benchmark tasks spanning reasoning, comprehension, and inference, and (3) correlation analysis between translation metrics and benchmark scores.

\subsection{Translation Evaluation}
We adopt \textsc{Flores-200}, \textsc{WMT24++}, and \textsc{NTREX} datasets, which provide translated parallel corpora in about 200 languages, as our primary testbed for translation quality \cite{nllb2022, deutsch-etal-2025-wmt24, federmann-etal-2022-ntrex,enevoldsen2025mmtebmassivemultilingualtext, muennighoff2022mteb}. These datasets cover a broad spectrum of language families and resource levels, ensuring representativeness for both high- and low-resource settings (Table \ref{tab:supported_languages} in Appendix). We use different datasets to ensure that the results we see are not singularly aligned with any corpus. 

For the translation task, we adopt a zero-shot prompting strategy designed to elicit 
faithful translations without additional fine-tuning. We use a direct instruction 
format, structured as follows:
\begin{quote}
    \texttt{Translate the following sentence into \{lang\}. Do not output any other text. \\
    English: \{s\} \\
    \{lang\}:}
\end{quote}

\noindent The resulting outputs are evaluated against the corresponding reference translations. Rather than employing a uniform deterministic setting, we utilize model-specific sampling parameters to align with the recommended configurations for each architecture (Table~\ref{tab:sampling_params}). We maintain a consistent limit of 1,024 maximum tokens across all evaluations. For models with specialized reasoning capabilities, we explicitly disabled those modes during inference. Any residual reasoning traces (e.g., <think> blocks) were strictly filtered out during post-processing to ensure consistency.

\begin{table}[ht]
\centering
\small
\begin{tabular}{lcccc}
\hline
\textbf{Model} & \textbf{temp} & \textbf{top\_p} & \textbf{top\_k} & \textbf{max\_tokens} \\ \hline
Gemma                 & 1.00                & 0.95            & 64              & 1024                 \\
Qwen                  & 0.70                & 0.80            & 20              & 1024                 \\
DeepSeek              & 0.60                & 0.95            & 0.0             & 1024                 \\
Llama                 & 0.60                & 0.90            & 0.0             & 1024                 \\
Mistral               & 0.15                & 1               & 0.0             & 1024                 \\
Phi                   & 0.80                & 0.95            & 0.0             & 1024    \\ \hline
\end{tabular}
\caption{Sampling parameters utilized for each model family during evaluation.}
\label{tab:sampling_params}
\end{table}

Translation quality is assessed using both established surface-level metrics and modern neural evaluation approaches. We employ \textsc{BLEU} \citep{papineni-etal-2002-bleu}, which measures n-gram precision between candidate and reference translations, and \textsc{ChrF++} \citep{popovic-2017-chrf}, a character-level F-score metric that also incorporates word n-grams for improved correlation with human judgments. Additionally, we include \textsc{ROUGE-L} \citep{lin-2004-rouge}, which focuses on longest common subsequence matching, and \textsc{METEOR} \citep{denkowski-lavie-2014-meteor}, which balances precision and recall while accounting for stemming and synonymy. To complement these lexical overlap metrics, we incorporate neural metrics that better capture semantic equivalence: \textsc{xCOMET} \citep{guerreiro-etal-2024-xcomet}, which combines sentence-level evaluation with fine-grained error detection capabilities, \textsc{SSA-COMET} \citep{li2025ssacometllmsoutperformlearned}, an update from \textsc{AfriCOMET} \citep{wang-etal-2024-afrimte} using a large human-annotated dataset, and \textsc{MetricX} \citep{juraska-etal-2024-metricx}, the Google submission to the \textsc{WMT 2024 Metrics Shared Task.} 

This comprehensive evaluation framework allows us to assess both surface-level similarity and deeper semantic preservation across translations. 

\begin{figure*}[t!]
    \centering
    \includegraphics[width=\textwidth]{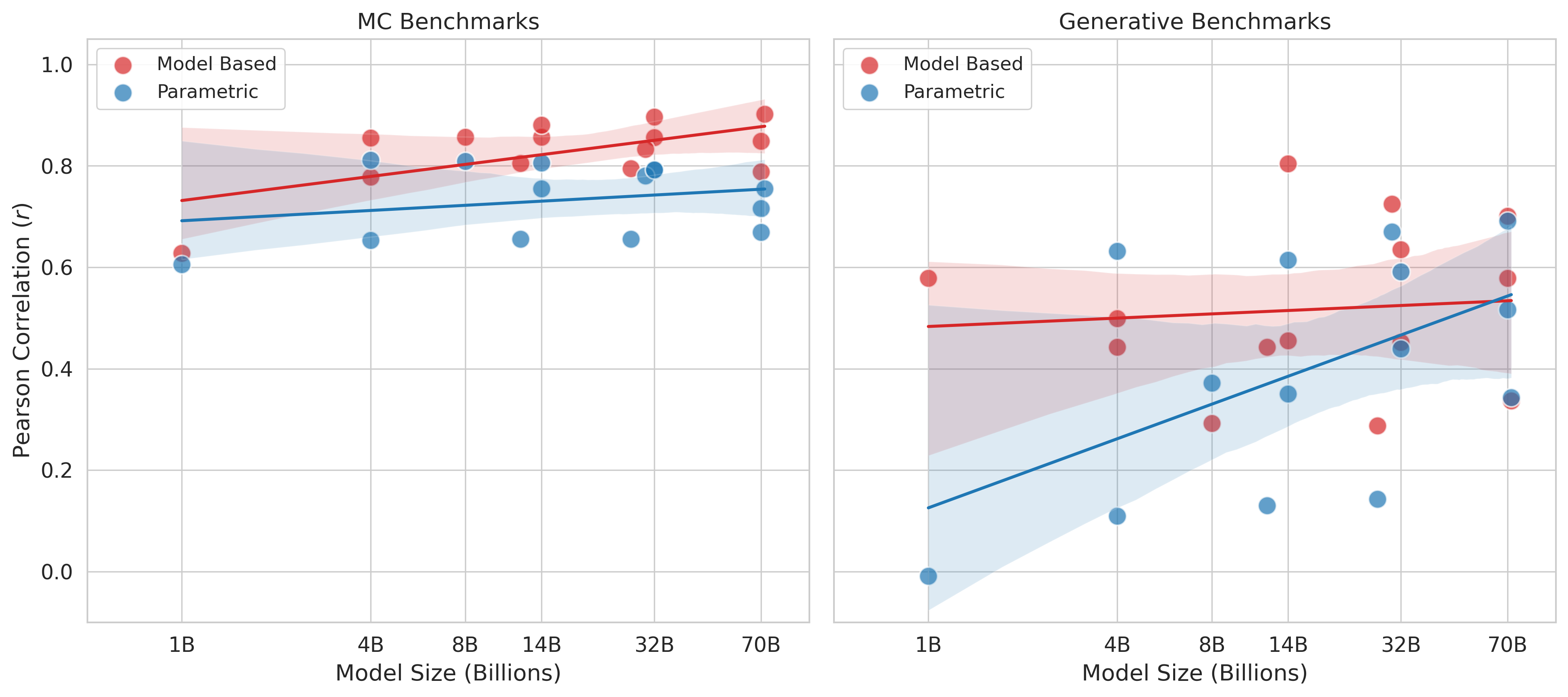}
    \caption{Aggregated Pearson correlations between MT metrics and multilingual benchmarks across 14 models ranging from 1B to 72B parameters on the \textsc{Flores-200} dataset. \textbf{Left:} Correlations with Multiple Choice (MC) benchmarks. \textbf{Right}: Correlations with Generative benchmarks. Shaded areas indicate 95\% confidence intervals. Red lines represent model-based neural MT metrics, while blue lines represent parametric metrics.
    }
    \label{fig:scaling}
\end{figure*}

\subsection{Benchmarks}
We evaluate models across 9 multilingual benchmarks to cover complementary task categories:

\noindent \textbf{Global MMLU}: A multilingual extension of the Massive Multitask Language Understanding benchmark, testing factual knowledge and academic reasoning across languages \citep{hendryckstest2021, singh-etal-2025-global}. We use the \textsc{"Lite"} variant, which contains of 15 languages with fully human-translated or post-edited content. We use accuracy scores in our analysis.

\noindent \textbf{Belebele}: A large-scale reading comprehension benchmark covering 122 languages \citep{bandarkar-etal-2024-belebele}. Each task consists of a short passage from the \textsc{Flores-200} dataset followed by a multiple-choice question. We use accuracy scores in our analysis. 

\noindent \textbf{MLQA}: A cross-lingual extractive question answering benchmark spanning seven languages and designed in the \textsc{SQuAD} format \citep{lewis-etal-2020-mlqa}. We use f1 scores from English to target language in our analysis. 

\noindent \textbf{MGSM}: A multilingual extension of the \textsc{GSM8K} dataset for mathematical reasoning \citep{cobbe2021gsm8k} in ten languages \citep{shi2022language}. We evaluate in the direct prediction setting, where the model generates a final answer without reasoning steps. We use direct exact match scores in our analysis. 

\noindent \textbf{IrokoBench}: A benchmark suite for African languages covering 16 low-resource languages (\citealp{adelani-etal-2025-irokobench}). It includes adaptations of 3 well-known benchmarks. Due to tooling limitations, \textsc{AfriMGSM} was not included in our evaluation.
\begin{itemize}
\item \textbf{AfriMMLU}: An African language adaptation of MMLU, designed to test reasoning in LRLs. We use direct accuracy scores and Prompt 1 in our analysis. 
\item \textbf{AfriXNLI}:  A natural language inference dataset for African languages.  We use accuracy scores and Prompt 2 in our analysis.
\end{itemize}

\noindent \textbf{HellaSwag}: A commonsense reasoning benchmark originally introduced by \citet{zellers-etal-2019-hellaswag}. We use the multilingual version from the Okapi benchmark suite \citep{lai-etal-2023-okapi}, which includes 30 machine-translated languages. We use accuracy scores in our analysis. 

\noindent \textbf{TruthfulQA}: An adversarial benchmark for testing truthfulness and informativeness, originally developed by \citet{lin-etal-2022-truthfulqa}. We use the MC1 variant from the Okapi suite \citep{lai-etal-2023-okapi}. This variant covers 31 machine-translated languages. We use MC2 accuracy scores in our analysis. 

\noindent \textbf{INCLUDE}: A question answering benchmark emphasizing knowledge and reasoning across 44 languages to assess the capabilities of LLMs within the authentic linguistic contexts in which they are intended to operate \citep{romanou2024includeevaluatingmultilinguallanguage}. We use default accuracy scores in our analysis. 

\subsection{Model and Software Infrastructure}
All evaluations are conducted on state-of-the-art multilingual LLMs (Table ~\ref{tab:model-families} in Appendix). We generate outputs via the \textsc{vLLM} framework, which provides efficient distributed inference with high-throughput decoding \cite{kwon2023efficient}.

Benchmark results are obtained through \textsc{LM Eval Harness}, a standardized framework for large-scale model evaluation \cite{eval-harness}. These frameworks increase comparability with prior work and minimize implementation biases.

\subsection{Correlation Analysis}
To quantify the relationship between translation quality and benchmark performance, we compute both non-parametric and parametric correlation coefficients. Specifically, we report \textbf{Pearson’s correlation} ($r$) to capture linear associations, sensitive to magnitude and direction of variation \cite{pearson1895vii}. Also, \textbf{Spearman’s rank correlation} ($\rho$): to capture monotonic relationships, robust to non-linearities and outliers \cite{spearman1961proof}.  

Statistical significance is assessed at $\alpha = 0.05$ using two-tailed permutation tests (\(N=2000\)). This non-parametric approach is well-suited to our multilingual benchmark data as it remains robust under non-normal distributions and small sample sizes. Correlation coefficients are computed via  \texttt{scipy.stats} \cite{2020SciPy-NMeth}, while significance is determined via random shuffling. To aggregate performance across the fixed population of languages and tasks, we apply Fisher’s $z$-transformation to normalize correlation coefficients prior to averaging. Reported mean correlations are then derived via inverse transformation, ensuring a statistically rigorous representation of the central tendency.

\subsection{Reproducibility and Availability}
All experiments were run on NVIDIA H100 GPUs with 80GB memory. Translation and benchmark outputs, evaluation scripts, and correlation analyses are implemented in Python and will be released upon publication to support reproducibility.

\section{Results}
We analyze the relationship between MT quality and downstream multilingual task performance by computing correlations between 7 MT metrics and 9 multilingual benchmarks across 14 LLMs. This section highlights findings from the Phi-4 (14B) model as a representative case, as it exhibits stable behavior across tasks and metrics. Figure~\ref{fig:flagship} provides a granular view of the Pearson correlation matrix for Phi-4 on \textsc{FLORES-200}, with a complete set of its metric-dataset-benchmark correlations provided in Tables ~\ref{tab:transposed_correlations} and ~\ref{tab:phi_raw_scores} in the Appendix.

Beyond this individual case, we observe that the correlational strength of translation metrics scales consistently across model sizes. Figure~\ref{fig:scaling} summarizes these aggregated trends, illustrating how the correlation between MT quality and benchmark success persists across the 1B to 72B parameter range. Exhaustive results across the remaining 13 models can be found in Appendix~\ref{app:additional_models}.

\subsection{Overview of Translation-Performance Alignment}
\begin{figure}[H]
  \centering
  \begin{subfigure}[t]{\linewidth}
    \centering
    \includegraphics[width=\linewidth]{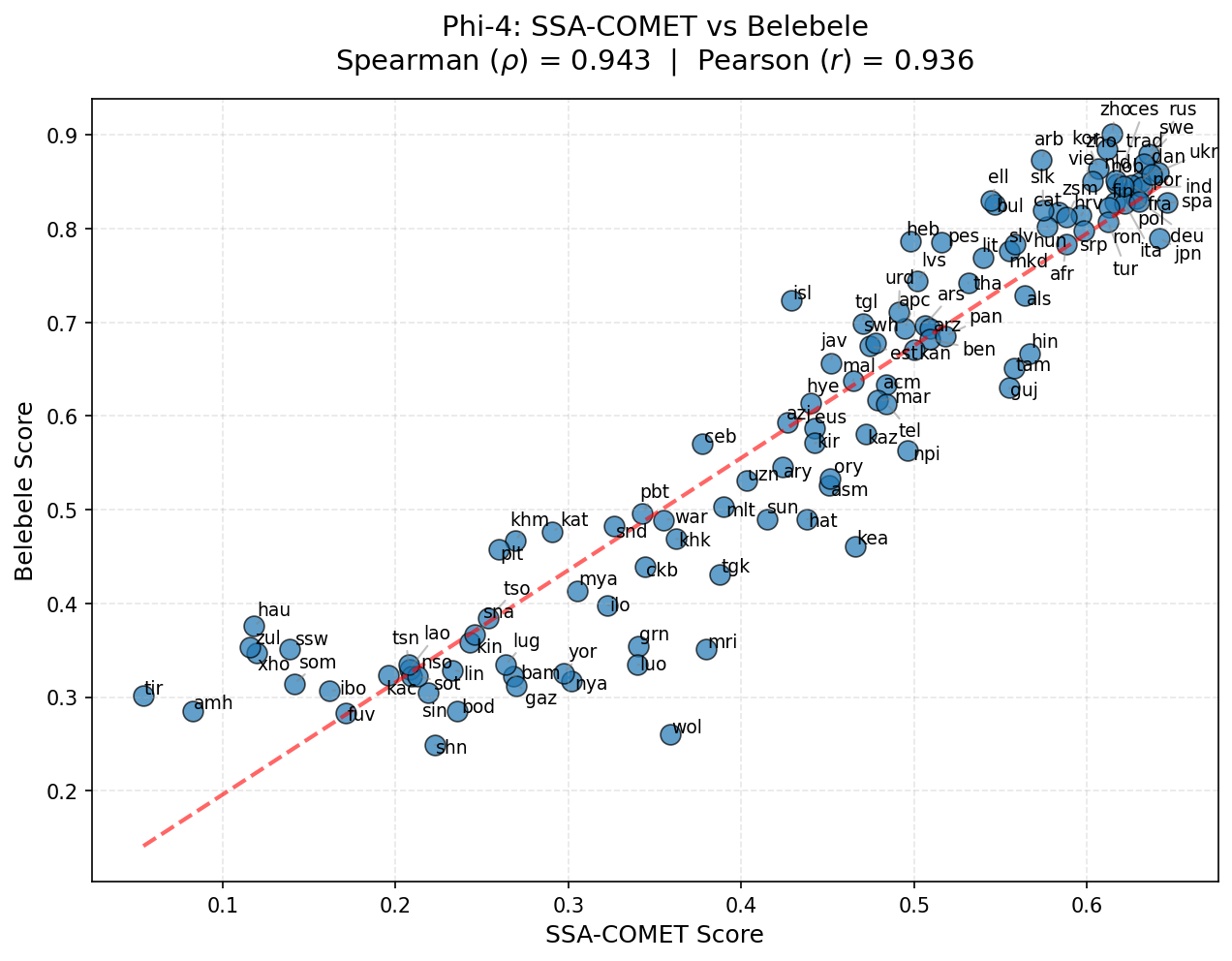}
  \end{subfigure}
  \caption{Correlation between translation quality (\textsc{SSA-COMET}) and reading comprehension (\textsc{Belebele}) for Phi-4 across 115 languages. The high coefficients ($r=0.936, \rho=0.943$) demonstrate a strong linear alignment between translation and NLU tasks.}
  \label{fig:belebele_vs_ssa-comet}
\end{figure}

\begin{figure}[H]
  \centering
  \begin{subfigure}[t]{\linewidth}
    \centering
    \includegraphics[width=\linewidth]{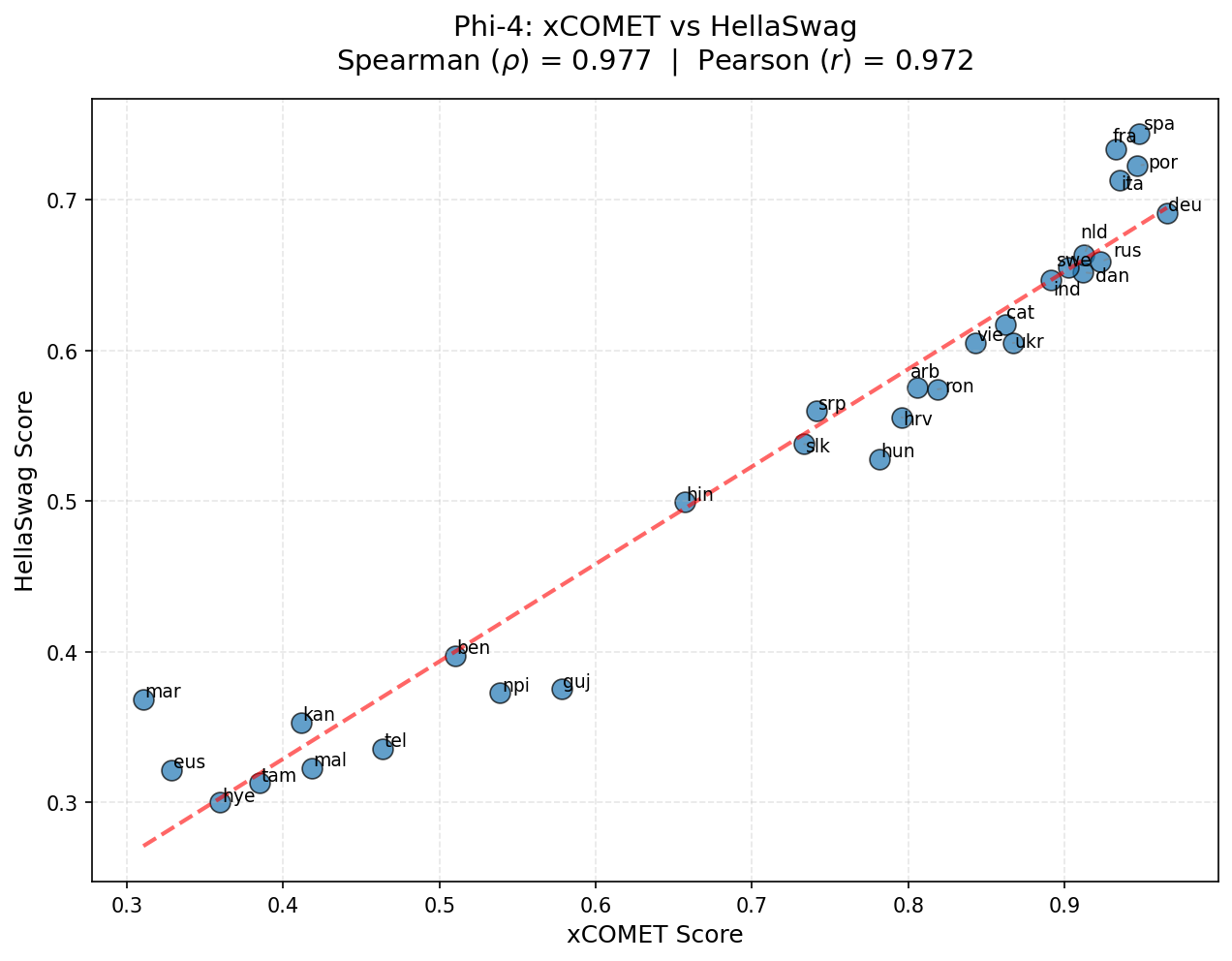}
  \end{subfigure}
  \caption{Correlation between translation quality (\textsc{xCOMET}) and commonsense reasoning (\textsc{HellaSwag}) for Phi-4 across 30 languages. The alignment ($r=0.972, \rho=0.977$) suggests that the semantic representations required for translation and commonsense reasoning are highly congruent.}
  \label{fig:hellaswag_vs_xcomet}
\end{figure}

Across languages and tasks, translation quality correlates strongly with downstream performance. For most metrics, the median correlation exceeds $r=0.80$. Neural metrics are the most consistently correlated: \textsc{xCOMET} achieves a median $r=0.91$, \textsc{MetricX} a median $r=0.89$, and \textsc{SSA-COMET} a median $r=0.87$. This linear relationship is visually apparent in Figure~\ref{fig:belebele_vs_ssa-comet} (reading comprehension) and Figure~\ref{fig:hellaswag_vs_xcomet} (commonsense reasoning).

Importantly, the relationship does not rely exclusively on neural evaluators. Purely lexical metrics retain substantial correlational strength: \textsc{BLEU} attains a median $r=0.79$, with \textsc{ChrF++} (median $r=0.63$) and \textsc{ROUGE-L} (median $r=0.66$) also remaining informative. The competitive performance of \textsc{BLEU} and \textsc{ROUGE-L} on many benchmark-dataset pairs provides a strong control: it suggests that the correlation signal originates in translation quality itself, rather than idiosyncrasies of learned MT evaluators. Overall, translation quality, regardless of the metric family, emerges as a meaningful proxy for multilingual task performance.

We also observe systematic differences by task category. Benchmarks emphasizing semantic understanding and inference show uniformly strong correlations (\textsc{AfriMMLU}: median $r=0.92$; \textsc{HellaSwag}: median $r=0.87$; \textsc{TruthfulQA}: median $r=0.76$). In contrast, tasks requiring more specialized competencies exhibit weaker and more variable alignment (\textsc{INCLUDE}: median $r=0.62$; \textsc{MGSM}: median $r=0.72$; \textsc{MLQA}: median $r=0.76$).

\paragraph{Scaling across model size.}
Figure~\ref{fig:scaling} aggregates correlations over all 14 models and shows that translation-benchmark alignment persists across the 1B-72B range (with 95\% confidence intervals), indicating that the proxy relationship is not specific to a single architecture or scale.

\paragraph{Figure-based exemplars.}
Figure~\ref{fig:belebele_vs_ssa-comet} illustrates the correlation between translation quality (\textsc{SSA-COMET}) and reading comprehension (\textsc{Belebele}) for Phi-4 across 115 languages, with high coefficients ($r=0.936$, $\rho=0.943$), indicating strong linear and rank alignment between translation and NLU performance. Figure~\ref{fig:hellaswag_vs_xcomet} shows a similarly strong relationship between translation quality (\textsc{xCOMET}) and commonsense reasoning (\textsc{HellaSwag}) for Phi-4 across 30 languages ($r=0.972$, $\rho=0.977$), consistent with highly shared semantic representations.

\subsection{Task-Based Analysis}
We next break down alignment by benchmark family to characterize where translation quality is most (and least) efficacious, and how this depends on metric choice and translation dataset.

\subsubsection{Knowledge and Academic Reasoning}
\textsc{Global MMLU} and \textsc{AfriMMLU} both show strong alignment between MT quality and benchmark performance, but with clear metric dependence. On \textsc{Global MMLU}, neural metrics are consistently strongest: \textsc{xCOMET} and \textsc{SSA-COMET} achieve correlations above 0.92 across datasets. \textsc{AfriMMLU} also exhibits high alignment across nearly all metrics, in part due to its limited data points. Traditional lexical metrics are more mixed: \textsc{BLEU} remains moderately to strongly correlated on both benchmarks, whereas \textsc{METEOR} and \textsc{ROUGE-L} degrade notably on \textsc{Global MMLU}, particularly on \textsc{FLORES-200}.

\subsubsection{Natural Language Understanding}
% \textbf{Reading Comprehension:} 

\textsc{Belebele} displays high translation-benchmark alignment (median \(r=0.83\)).
Neural metrics are consistently strong (e.g., \textsc{MetricX} \(r=0.71{-}0.92\), \textsc{SSA-COMET} \(r=0.85{-}0.94\), \textsc{xCOMET} \(r=0.83{-}0.93\)), while overlap metrics can be substantially lower on \textsc{WMT24++} (e.g., \textsc{ChrF++} \(r=0.44\) and \textsc{BLEU} \(r=0.69\) on \textsc{WMT24++}, versus \textsc{ChrF++} \(r=0.77{-}0.85\) and \textsc{BLEU} \(r=0.86{-}0.89\) on \textsc{FLORES-200}/\textsc{NTREX}; Table~\ref{tab:transposed_correlations}).
\textsc{MLQA} (median \(r=0.76\)) shows the largest spread across metrics, including the lowest correlation reported in Table~\ref{tab:transposed_correlations} (\textsc{METEOR} on \textsc{FLORES-200}: \(r=0.26\)).
In contrast, \textsc{MetricX} is consistently high on \textsc{MLQA} (\(r=0.93{-}0.98\)).
\textsc{AfriXNLI} has a lower median (median \(r=0.72\)) and notable metric sensitivity: on \textsc{FLORES-200} and \textsc{NTREX}, \textsc{SSA-COMET} is comparatively low (\(r=0.42\) and \(r=0.37\), respectively), while other metrics remain substantially higher (e.g., \textsc{MetricX} \(r=0.74{-}0.79\); Table~\ref{tab:transposed_correlations}).

\subsubsection{Specialized Reasoning}
\textsc{HellaSwag} shows strong alignment (median \(r=0.87\)), with neural metrics consistently high (e.g., \textsc{xCOMET} \(r=0.97{-}0.98\)) and even lexical metrics remaining informative (e.g., \textsc{BLEU} \(r=0.87{-}0.90\); Table~\ref{tab:transposed_correlations}).
\textsc{MGSM} exhibits weaker and more metric-dependent correlations (median \(r=0.72\)).
Notably, \textsc{ROUGE-L} is strongest for \textsc{MGSM} across datasets (\(r=0.78{-}0.81\)), while \textsc{MetricX} is comparatively lower (\(r=0.64{-}0.65\)).
\textsc{INCLUDE} shows the weakest median correlation (median \(r=0.62\)), with neural metrics providing the strongest signal (e.g., \textsc{xCOMET} \(r=0.73{-}0.78\), \textsc{MetricX} \(r=0.68{-}0.71\)) and overlap-based metrics generally lower (e.g., \textsc{ChrF++} \(r=0.34{-}0.56\); Table~\ref{tab:transposed_correlations}).

\subsection{Metric and Dataset Insights}
\subsubsection{Neural Metrics Performance}
Neural MT metrics are the most reliable indicators overall (median \(r=0.87{-}0.91\); Table~\ref{tab:transposed_correlations}).
Within this family, \textsc{xCOMET} is the strongest aggregate indicator (median \(r=0.91\)), while \textsc{MetricX} is particularly strong on \textsc{MLQA} (\(r=0.93{-}0.98\)) and remains competitive elsewhere.
\textsc{SSA-COMET} is highly correlated on many benchmarks (e.g., \textsc{Belebele}: \(r=0.85{-}0.94\); \textsc{TruthfulQA}: \(r=0.85{-}0.92\)), but can be substantially lower on \textsc{AfriXNLI} under \textsc{FLORES-200}/\textsc{NTREX} (\(r=0.42/0.37\)).

\subsubsection{Traditional Metrics Show Performance}
Among overlap-based metrics, \textsc{BLEU} is the most robust aggregate baseline (median \(r=0.79\)) and performs strongly on several benchmarks (e.g., \textsc{HellaSwag}: \(r=0.87{-}0.90\); \textsc{TruthfulQA}: \(r=0.70{-}0.79\)).
However, overlap metrics exhibit larger task sensitivity: \textsc{ROUGE-L} is unusually strong on \textsc{MGSM} (\(r=0.78{-}0.81\)), whereas \textsc{METEOR} can be very weak on \textsc{MLQA} (\(r=0.26\) on \textsc{FLORES-200}; Table~\ref{tab:transposed_correlations}).

\subsubsection{Dataset Effects}
Across \textsc{FLORES-200}, \textsc{WMT24++}, and \textsc{NTREX}, the high-level pattern is stable: neural metrics dominate, and benchmarks such as \textsc{Belebele} and \textsc{HellaSwag} show strong alignment.
At the same time, correlations for some benchmark-metric pairs are substantially lower on \textsc{WMT24++} (e.g., for \textsc{Belebele} under \textsc{ChrF++}/\textsc{BLEU}/\textsc{MetricX}/\textsc{xCOMET}; Table~\ref{tab:transposed_correlations}), consistent with domain and coverage differences across translation datasets.

\section{Conclusion}
Comprehensive multilingual evaluation remains fundamentally bottlenecked by benchmark availability: for most of the world's languages, there are no reliable, non-machine-translated task suites, and translating existing benchmarks introduces well-known validity issues.
This work asks whether \emph{translation quality}, measured on high-coverage parallel corpora, can serve as a scalable, low-cost proxy signal for a model's broader multilingual competence.

We answer this question with a large correlation study spanning 14 multilingual LLMs (1B--72B parameters), 9 multilingual benchmarks, 3 translation datasets (\textsc{FLORES-200}, \textsc{WMT24++}, \textsc{NTREX}), and 7 MT evaluation metrics.
Across this setting, translation quality is strongly indicative of downstream benchmark performance: neural MT metrics yield high median Pearson correlations, with \textsc{xCOMET} (median \(r=0.91\)), \textsc{MetricX} (median \(r=0.89\)), and \textsc{SSA-COMET} (median \(r=0.87\)).
Importantly, the relationship is not confined to learned evaluators: even overlap-based metrics remain informative (e.g., \textsc{BLEU} median \(r=0.79\)), indicating that the proxy signal is grounded in translation performance itself.
The strength of this alignment persists across model sizes, supporting translation evaluation as a stable first-pass indicator rather than a model-specific artifact.

At the same time, our results clarify the boundary of what translation can proxy.
Correlation strength varies by task family, and correlations are weaker for some specialized benchmarks (e.g., \textsc{INCLUDE} and \textsc{MGSM}), indicating that translation quality cannot replace task-specific evaluation when the target application depends on competencies not well captured by translation.
We therefore view translation not as a substitute for multilingual benchmarks, but as a practical \emph{screening layer}: translation-first evaluation can rapidly surface where models are likely to succeed (or fail) across many languages, enabling targeted downstream benchmarking where it is most needed.

Beyond methodological innovation, this work has practical and ethical implications.
By turning translation evaluation into a first-pass measurement tool, it offers a blueprint for expanding evaluation reach far beyond the small set of languages covered by high-quality, non-translated benchmarks.
In particular, translation-based proxies could help communities and researchers obtain meaningful, comparable signals of multilingual model competence at a fraction of the cost of building and running task suites, supporting fairer and more inclusive progress in multilingual language technology.

Ultimately, our results suggest that broad multilingual assessment may benefit less from endlessly expanding task suites, and more from rethinking what evaluation can be at scale.
Translation, the act of comprehending and re-expressing meaning across linguistic boundaries, serves here as both a scientific lens and a practical tool: it is widely available, cheap to run, and empirically linked to downstream success across many benchmarks, while still leaving clear cases where targeted task evaluation is necessary.
By grounding evaluation in translation quality through a staged, translation-first framework, we move toward a more scalable evaluation ecosystem in which more languages can be meaningfully measured (and therefore improved), without overstating translation as a complete substitute for task-centric benchmarks.

\section{Limitations}
\label{sec:limitations}
While our results provide strong evidence that translation quality can serve as a useful first-pass proxy for multilingual performance, several limitations constrain the scope and interpretation of this study.

\paragraph{Translation datasets are curated and do not span all real-world language use.}
Our proxy signals are computed on \textsc{FLORES-200}, \textsc{WMT24++}, and \textsc{NTREX}, which are high-quality, reference-based parallel corpora.
These datasets primarily cover standardized written text and may under-represent dialectal variation, informal registers, code-switching, and domain-specific language.
As a result, strong translation-benchmark correlations on these corpora do not guarantee equivalent correlation in more diverse, community-specific, or conversational settings.

\paragraph{MT metrics introduce measurement bias, especially for low-resource languages.}
We rely on both lexical metrics and neural MT evaluators (\textsc{xCOMET}, \textsc{SSA-COMET}, \textsc{MetricX}).
Neural metrics are trained to align with human judgments and can be sensitive to the distribution and language coverage of their training data.
Therefore, MT scores may partially reflect metric-specific biases rather than pure underlying model competence, particularly for languages and scripts that are under-represented in the metric training signal.
The fact that overlap metrics remain informative mitigates (but does not eliminate) this concern.

\paragraph{The proxy is incomplete: some tasks depend on competencies not well captured by translation.}
Translation quality is an indirect signal of multilingual capability.
Our own results show weaker median correlations on several benchmarks (e.g., \textsc{INCLUDE} and \textsc{MGSM}; Table~\ref{tab:transposed_correlations}), indicating that translation cannot replace task-centric evaluation when downstream success is gated by specialized reasoning, task format constraints, or domain-specific knowledge.
Consequently, we frame translation as a screening layer, not as a universal substitute for multilingual benchmarks.

\paragraph{Language overlap is sparse, yielding small sample sizes for some correlations.}
Correlation estimates are computed only over the intersection of languages supported by each translation dataset and each benchmark.
This overlap is uneven across tasks and corpora (e.g., some \textsc{WMT24++} correlations use as few as \(n=3\) languages; Table~\ref{tab:transposed_correlations}), which can make estimates statistically fragile and inflate the apparent strength of association.
Similarly, some benchmarks have limited language coverage in our evaluation setup, restricting the stability and generality of per-benchmark conclusions.

\paragraph{Experimental scope: models, prompting, and translation direction.}
We evaluate 14 instruction-tuned LLMs across several model families, but the findings may not transfer to all architectures, pretraining mixes, or decoding regimes.
Translation is elicited via a single, direct, zero-shot prompt and evaluated for English-to-target translation; correlations could differ under alternative prompting strategies, decoding settings, or translation directions.
We also disable explicit ``reasoning modes'' where applicable and filter any residual traces during post-processing; while this improves consistency, it may not reflect how such models are deployed in practice.

\paragraph{Correlational evidence does not establish causation; contamination cannot be fully ruled out.}
Our analysis quantifies association, not mechanism: high translation quality and high downstream performance may co-occur due to shared representations, shared training data, or other confounds.
Because model training corpora are not fully disclosed, we cannot definitively rule out data contamination for either benchmarks or translation test sets.
We mitigate this concern through breadth (multiple datasets, metrics, tasks, and model families) and statistical testing, but controlled causal studies would be needed to establish why translation quality influences downstream success.

Despite these limitations, our results provide a clear empirical foundation: translation metrics can substantially reduce the cost and coverage gap in multilingual evaluation. Future work should extend this approach to low-resource settings, non-standard text types, and causally controlled training regimes to better understand the boundaries of translation as a universal evaluator.
\clearpage
\bibliography{custom}

@misc{li2025ssacometllmsoutperformlearned,
      title={SSA-COMET: Do LLMs Outperform Learned Metrics in Evaluating MT for Under-Resourced African Languages?}, 
      author={Senyu Li and Jiayi Wang and Felermino D. M. A. Ali and Colin Cherry and Daniel Deutsch and Eleftheria Briakou and Rui Sousa-Silva and Henrique Lopes Cardoso and Pontus Stenetorp and David Ifeoluwa Adelani},
      year={2025},
      eprint={2506.04557},
      archivePrefix={arXiv},
      primaryClass={cs.CL},
      url={https://arxiv.org/abs/2506.04557}, 
}

@inproceedings{singh-etal-2025-global,
    title = "Global {MMLU}: Understanding and Addressing Cultural and Linguistic Biases in Multilingual Evaluation",
    author = "Singh, Shivalika  and
      Romanou, Angelika  and
      Fourrier, Cl{\'e}mentine  and
      Adelani, David Ifeoluwa  and
      Ngui, Jian Gang  and
      Vila-Suero, Daniel  and
      Limkonchotiwat, Peerat  and
      Marchisio, Kelly  and
      Leong, Wei Qi  and
      Susanto, Yosephine  and
      Ng, Raymond  and
      Longpre, Shayne  and
      Ruder, Sebastian  and
      Ko, Wei-Yin  and
      Bosselut, Antoine  and
      Oh, Alice  and
      Martins, Andre  and
      Choshen, Leshem  and
      Ippolito, Daphne  and
      Ferrante, Enzo  and
      Fadaee, Marzieh  and
      Ermis, Beyza  and
      Hooker, Sara",
    editor = "Che, Wanxiang  and
      Nabende, Joyce  and
      Shutova, Ekaterina  and
      Pilehvar, Mohammad Taher",
    booktitle = "Proceedings of the 63rd Annual Meeting of the Association for Computational Linguistics (Volume 1: Long Papers)",
    month = jul,
    year = "2025",
    address = "Vienna, Austria",
    publisher = "Association for Computational Linguistics",
    url = "https://aclanthology.org/2025.acl-long.919/",
    doi = "10.18653/v1/2025.acl-long.919",
    pages = "18761--18799",
    ISBN = "979-8-89176-251-0"
}

@inproceedings{conneau-etal-2018-xnli,
    title = "{XNLI}: Evaluating Cross-lingual Sentence Representations",
    author = "Conneau, Alexis  and
      Rinott, Ruty  and
      Lample, Guillaume  and
      Williams, Adina  and
      Bowman, Samuel  and
      Schwenk, Holger  and
      Stoyanov, Veselin",
    editor = "Riloff, Ellen  and
      Chiang, David  and
      Hockenmaier, Julia  and
      Tsujii, Jun{'}ichi",
    booktitle = "Proceedings of the 2018 Conference on Empirical Methods in Natural Language Processing",
    month = oct # "-" # nov,
    year = "2018",
    address = "Brussels, Belgium",
    publisher = "Association for Computational Linguistics",
    url = "https://aclanthology.org/D18-1269/",
    doi = "10.18653/v1/D18-1269",
    pages = "2475--2485"
}

@article{shi2022language,
  title={Language models are multilingual chain-of-thought reasoners},
  author={Shi, Freda and Suzgun, Mirac and Freitag, Markus and Wang, Xuezhi and Srivats, Suraj and Vosoughi, Soroush and Chung, Hyung Won and Tay, Yi and Ruder, Sebastian and Zhou, Denny and others},
  journal={arXiv preprint arXiv:2210.03057},
  year={2022}
}

@inproceedings{ponti-etal-2020-xcopa,
    title = "{XCOPA}: A Multilingual Dataset for Causal Commonsense Reasoning",
    author = "Ponti, Edoardo Maria  and
      Glava{\v{s}}, Goran  and
      Majewska, Olga  and
      Liu, Qianchu  and
      Vuli{\'c}, Ivan  and
      Korhonen, Anna",
    editor = "Webber, Bonnie  and
      Cohn, Trevor  and
      He, Yulan  and
      Liu, Yang",
    booktitle = "Proceedings of the 2020 Conference on Empirical Methods in Natural Language Processing (EMNLP)",
    month = nov,
    year = "2020",
    address = "Online",
    publisher = "Association for Computational Linguistics",
    url = "https://aclanthology.org/2020.emnlp-main.185/",
    doi = "10.18653/v1/2020.emnlp-main.185",
    pages = "2362--2376"
}

@misc{zhang2025pmmevalparallelmultilingualmultitask,
      title={P-MMEval: A Parallel Multilingual Multitask Benchmark for Consistent Evaluation of LLMs}, 
      author={Yidan Zhang and Yu Wan and Boyi Deng and Baosong Yang and Haoran Wei and Fei Huang and Bowen Yu and Junyang Lin and Fei Huang and Jingren Zhou},
      year={2025},
      eprint={2411.09116},
      archivePrefix={arXiv},
      primaryClass={cs.CL},
      url={https://arxiv.org/abs/2411.09116}, 
}

@inproceedings{bandarkar-etal-2024-belebele,
    title = "The Belebele Benchmark: a Parallel Reading Comprehension Dataset in 122 Language Variants",
    author = "Bandarkar, Lucas  and
      Liang, Davis  and
      Muller, Benjamin  and
      Artetxe, Mikel  and
      Shukla, Satya Narayan  and
      Husa, Donald  and
      Goyal, Naman  and
      Krishnan, Abhinandan  and
      Zettlemoyer, Luke  and
      Khabsa, Madian",
    editor = "Ku, Lun-Wei  and
      Martins, Andre  and
      Srikumar, Vivek",
    booktitle = "Proceedings of the 62nd Annual Meeting of the Association for Computational Linguistics (Volume 1: Long Papers)",
    month = aug,
    year = "2024",
    address = "Bangkok, Thailand",
    publisher = "Association for Computational Linguistics",
    url = "https://aclanthology.org/2024.acl-long.44/",
    doi = "10.18653/v1/2024.acl-long.44",
    pages = "749--775"
}

@inproceedings{papineni-etal-2002-bleu,
    title = "{B}leu: a Method for Automatic Evaluation of Machine Translation",
    author = "Papineni, Kishore  and
      Roukos, Salim  and
      Ward, Todd  and
      Zhu, Wei-Jing",
    editor = "Isabelle, Pierre  and
      Charniak, Eugene  and
      Lin, Dekang",
    booktitle = "Proceedings of the 40th Annual Meeting of the Association for Computational Linguistics",
    month = jul,
    year = "2002",
    address = "Philadelphia, Pennsylvania, USA",
    publisher = "Association for Computational Linguistics",
    url = "https://aclanthology.org/P02-1040/",
    doi = "10.3115/1073083.1073135",
    pages = "311--318"
}

@inproceedings{popovic-2017-chrf,
    title = "chr{F}++: words helping character n-grams",
    author = "Popovi{\'c}, Maja",
    editor = "Bojar, Ond{\v{r}}ej  and
      Buck, Christian  and
      Chatterjee, Rajen  and
      Federmann, Christian  and
      Graham, Yvette  and
      Haddow, Barry  and
      Huck, Matthias  and
      Yepes, Antonio Jimeno  and
      Koehn, Philipp  and
      Kreutzer, Julia",
    booktitle = "Proceedings of the Second Conference on Machine Translation",
    month = sep,
    year = "2017",
    address = "Copenhagen, Denmark",
    publisher = "Association for Computational Linguistics",
    url = "https://aclanthology.org/W17-4770/",
    doi = "10.18653/v1/W17-4770",
    pages = "612--618"
}

@inproceedings{rei-etal-2020-comet,
    title = "{COMET}: A Neural Framework for {MT} Evaluation",
    author = "Rei, Ricardo  and
      Stewart, Craig  and
      Farinha, Ana C  and
      Lavie, Alon",
    editor = "Webber, Bonnie  and
      Cohn, Trevor  and
      He, Yulan  and
      Liu, Yang",
    booktitle = "Proceedings of the 2020 Conference on Empirical Methods in Natural Language Processing (EMNLP)",
    month = nov,
    year = "2020",
    address = "Online",
    publisher = "Association for Computational Linguistics",
    url = "https://aclanthology.org/2020.emnlp-main.213/",
    doi = "10.18653/v1/2020.emnlp-main.213",
    pages = "2685--2702"
}

@inproceedings{juraska-etal-2024-metricx,
    title = "{M}etric{X}-24: The {G}oogle Submission to the {WMT} 2024 Metrics Shared Task",
    author = "Juraska, Juraj  and
      Deutsch, Daniel  and
      Finkelstein, Mara  and
      Freitag, Markus",
    editor = "Haddow, Barry  and
      Kocmi, Tom  and
      Koehn, Philipp  and
      Monz, Christof",
    booktitle = "Proceedings of the Ninth Conference on Machine Translation",
    month = nov,
    year = "2024",
    address = "Miami, Florida, USA",
    publisher = "Association for Computational Linguistics",
    url = "https://aclanthology.org/2024.wmt-1.35/",
    doi = "10.18653/v1/2024.wmt-1.35",
    pages = "492--504"
}

@article{nllb2022,
      author={{NLLB Team} and Marta R. Costa-jussà and James Cross and Onur Çelebi and Maha Elbayad and Kenneth Heafield and Kevin Heffernan and Elahe Kalbassi and Janice Lam and Daniel Licht and Jean Maillard and Anna Sun and Skyler Wang and Guillaume Wenzek and Al Youngblood and Bapi Akula and Loic Barrault and Gabriel Mejia Gonzalez and Prangthip Hansanti and John Hoffman and Semarley Jarrett and Kaushik Ram Sadagopan and Dirk Rowe and Shannon Spruit and Chau Tran and Pierre Andrews and Necip Fazil Ayan and Shruti Bhosale and Sergey Edunov and Angela Fan and Cynthia Gao and Vedanuj Goswami and Francisco Guzmán and Philipp Koehn and Alexandre Mourachko and Christophe Ropers and Safiyyah Saleem and Holger Schwenk and Jeff Wang},
title     = {No Language Left Behind: Scaling Human-Centered Machine Translation},
      year      = {2022}
}

@inproceedings{denkowski-lavie-2014-meteor,
    title = "Meteor Universal: Language Specific Translation Evaluation for Any Target Language",
    author = "Denkowski, Michael  and
      Lavie, Alon",
    editor = "Bojar, Ond{\v{r}}ej  and
      Buck, Christian  and
      Federmann, Christian  and
      Haddow, Barry  and
      Koehn, Philipp  and
      Monz, Christof  and
      Post, Matt  and
      Specia, Lucia",
    booktitle = "Proceedings of the Ninth Workshop on Statistical Machine Translation",
    month = jun,
    year = "2014",
    address = "Baltimore, Maryland, USA",
    publisher = "Association for Computational Linguistics",
    url = "https://aclanthology.org/W14-3348/",
    doi = "10.3115/v1/W14-3348",
    pages = "376--380"
}

@inproceedings{lewis-etal-2020-mlqa,
    title = "{MLQA}: Evaluating Cross-lingual Extractive Question Answering",
    author = "Lewis, Patrick  and
      Oguz, Barlas  and
      Rinott, Ruty  and
      Riedel, Sebastian  and
      Schwenk, Holger",
    editor = "Jurafsky, Dan  and
      Chai, Joyce  and
      Schluter, Natalie  and
      Tetreault, Joel",
    booktitle = "Proceedings of the 58th Annual Meeting of the Association for Computational Linguistics",
    month = jul,
    year = "2020",
    address = "Online",
    publisher = "Association for Computational Linguistics",
    url = "https://aclanthology.org/2020.acl-main.653/",
    doi = "10.18653/v1/2020.acl-main.653",
    pages = "7315--7330"
}

@inproceedings{adelani-etal-2025-irokobench,
    title = "{I}roko{B}ench: A New Benchmark for {A}frican Languages in the Age of Large Language Models",
    author = "Adelani, David Ifeoluwa  and
      Ojo, Jessica  and
      Azime, Israel Abebe  and
      Zhuang, Jian Yun  and
      Alabi, Jesujoba Oluwadara  and
      He, Xuanli  and
      Ochieng, Millicent  and
      Hooker, Sara  and
      Bukula, Andiswa  and
      Lee, En-Shiun Annie  and
      Chukwuneke, Chiamaka Ijeoma  and
      Buzaaba, Happy  and
      Sibanda, Blessing Kudzaishe  and
      Kalipe, Godson Koffi  and
      Mukiibi, Jonathan  and
      Kabongo Kabenamualu, Salomon  and
      Yuehgoh, Foutse  and
      Setaka, Mmasibidi  and
      Ndolela, Lolwethu  and
      Odu, Nkiruka  and
      Mabuya, Rooweither  and
      Osei, Salomey  and
      Muhammad, Shamsuddeen Hassan  and
      Samb, Sokhar  and
      Guge, Tadesse Kebede  and
      Sherman, Tombekai Vangoni  and
      Stenetorp, Pontus",
    editor = "Chiruzzo, Luis  and
      Ritter, Alan  and
      Wang, Lu",
    booktitle = "Proceedings of the 2025 Conference of the Nations of the Americas Chapter of the Association for Computational Linguistics: Human Language Technologies (Volume 1: Long Papers)",
    month = apr,
    year = "2025",
    address = "Albuquerque, New Mexico",
    publisher = "Association for Computational Linguistics",
    url = "https://aclanthology.org/2025.naacl-long.139/",
    doi = "10.18653/v1/2025.naacl-long.139",
    pages = "2732--2757",
    ISBN = "979-8-89176-189-6"
}

@inproceedings{lin-2004-rouge,
    title = "{ROUGE}: A Package for Automatic Evaluation of Summaries",
    author = "Lin, Chin-Yew",
    booktitle = "Text Summarization Branches Out",
    month = jul,
    year = "2004",
    address = "Barcelona, Spain",
    publisher = "Association for Computational Linguistics",
    url = "https://aclanthology.org/W04-1013/",
    pages = "74--81"
}

@inproceedings{wang-etal-2024-afrimte,
    title = "{A}fri{MTE} and {A}fri{COMET}: Enhancing {COMET} to Embrace Under-resourced {A}frican Languages",
    author = "Wang, Jiayi  and
      Adelani, David Ifeoluwa  and
      Agrawal, Sweta  and
      Masiak, Marek  and
      Rei, Ricardo  and
      Briakou, Eleftheria  and
      Carpuat, Marine  and
      He, Xuanli  and
      Bourhim, Sofia  and
      Bukula, Andiswa  and
      Mohamed, Muhidin  and
      Olatoye, Temitayo  and
      Adewumi, Tosin  and
      Mokayed, Hamam  and
      Mwase, Christine  and
      Kimotho, Wangui  and
      Yuehgoh, Foutse  and
      Aremu, Anuoluwapo  and
      Ojo, Jessica  and
      Muhammad, Shamsuddeen Hassan  and
      Osei, Salomey  and
      Omotayo, Abdul-Hakeem  and
      Chukwuneke, Chiamaka  and
      Ogayo, Perez  and
      Hourrane, Oumaima  and
      El Anigri, Salma  and
      Ndolela, Lolwethu  and
      Mangwana, Thabiso  and
      Mohamed, Shafie Abdi  and
      Ayinde, Hassan  and
      Awoyomi, Oluwabusayo Olufunke  and
      Alkhaled, Lama  and
      Al-azzawi, Sana  and
      Etori, Naome A.  and
      Ochieng, Millicent  and
      Siro, Clemencia  and
      Kiragu, Njoroge  and
      Muchiri, Eric  and
      Kimotho, Wangari  and
      Wamba Momo, Lyse Naomi  and
      Abolade, Daud  and
      Ajao, Simbiat  and
      Shode, Iyanuoluwa  and
      Macharm, Ricky  and
      Iro, Ruqayya Nasir  and
      Abdullahi, Saheed S.  and
      Moore, Stephen E.  and
      Opoku, Bernard  and
      Akinjobi, Zainab  and
      Afolabi, Abeeb  and
      Obiefuna, Nnaemeka  and
      Ogbu, Onyekachi Raphael  and
      Ochieng{'}, Sam  and
      Otiende, Verrah Akinyi  and
      Mbonu, Chinedu Emmanuel  and
      Toadoum Sari, Sakayo  and
      Lu, Yao  and
      Stenetorp, Pontus",
    editor = "Duh, Kevin  and
      Gomez, Helena  and
      Bethard, Steven",
    booktitle = "Proceedings of the 2024 Conference of the North American Chapter of the Association for Computational Linguistics: Human Language Technologies (Volume 1: Long Papers)",
    month = jun,
    year = "2024",
    address = "Mexico City, Mexico",
    publisher = "Association for Computational Linguistics",
    url = "https://aclanthology.org/2024.naacl-long.334/",
    doi = "10.18653/v1/2024.naacl-long.334",
    pages = "5997--6023"
}

@article{guerreiro-etal-2024-xcomet,
    title = "xcomet: Transparent Machine Translation Evaluation through Fine-grained Error Detection",
    author = "Guerreiro, Nuno M.  and
      Rei, Ricardo  and
      Stigt, Daan van  and
      Coheur, Luisa  and
      Colombo, Pierre  and
      Martins, Andr{\'e} F. T.",
    journal = "Transactions of the Association for Computational Linguistics",
    volume = "12",
    year = "2024",
    address = "Cambridge, MA",
    publisher = "MIT Press",
    url = "https://aclanthology.org/2024.tacl-1.54/",
    doi = "10.1162/tacl_a_00683",
    pages = "979--995"
}

@inproceedings{lai-etal-2023-okapi,
    title = "Okapi: Instruction-tuned Large Language Models in Multiple Languages with Reinforcement Learning from Human Feedback",
    author = "Lai, Viet  and
      Nguyen, Chien  and
      Ngo, Nghia  and
      Nguyen, Thuat  and
      Dernoncourt, Franck  and
      Rossi, Ryan  and
      Nguyen, Thien",
    editor = "Feng, Yansong  and
      Lefever, Els",
    booktitle = "Proceedings of the 2023 Conference on Empirical Methods in Natural Language Processing: System Demonstrations",
    month = dec,
    year = "2023",
    address = "Singapore",
    publisher = "Association for Computational Linguistics",
    url = "https://aclanthology.org/2023.emnlp-demo.28/",
    doi = "10.18653/v1/2023.emnlp-demo.28",
    pages = "318--327"
}

@inproceedings{zellers-etal-2019-hellaswag,
    title = "{H}ella{S}wag: Can a Machine Really Finish Your Sentence?",
    author = "Zellers, Rowan  and
      Holtzman, Ari  and
      Bisk, Yonatan  and
      Farhadi, Ali  and
      Choi, Yejin",
    editor = "Korhonen, Anna  and
      Traum, David  and
      M{\`a}rquez, Llu{\'i}s",
    booktitle = "Proceedings of the 57th Annual Meeting of the Association for Computational Linguistics",
    month = jul,
    year = "2019",
    address = "Florence, Italy",
    publisher = "Association for Computational Linguistics",
    url = "https://aclanthology.org/P19-1472/",
    doi = "10.18653/v1/P19-1472",
    pages = "4791--4800"
}

@inproceedings{lin-etal-2022-truthfulqa,
    title = "{T}ruthful{QA}: Measuring How Models Mimic Human Falsehoods",
    author = "Lin, Stephanie  and
      Hilton, Jacob  and
      Evans, Owain",
    editor = "Muresan, Smaranda  and
      Nakov, Preslav  and
      Villavicencio, Aline",
    booktitle = "Proceedings of the 60th Annual Meeting of the Association for Computational Linguistics (Volume 1: Long Papers)",
    month = may,
    year = "2022",
    address = "Dublin, Ireland",
    publisher = "Association for Computational Linguistics",
    url = "https://aclanthology.org/2022.acl-long.229/",
    doi = "10.18653/v1/2022.acl-long.229",
    pages = "3214--3252"
}

@article{hendryckstest2021,
  title={Measuring Massive Multitask Language Understanding},
  author={Dan Hendrycks and Collin Burns and Steven Basart and Andy Zou and Mantas Mazeika and Dawn Song and Jacob Steinhardt},
  journal={Proceedings of the International Conference on Learning Representations (ICLR)},
  year={2021}
}

@article{cobbe2021gsm8k,
  title={Training Verifiers to Solve Math Word Problems},
  author={Cobbe, Karl and Kosaraju, Vineet and Bavarian, Mohammad and Chen, Mark and Jun, Heewoo and Kaiser, Lukasz and Plappert, Matthias and Tworek, Jerry and Hilton, Jacob and Nakano, Reiichiro and Hesse, Christopher and Schulman, John},
  journal={arXiv preprint arXiv:2110.14168},
  year={2021}
}

@misc{zhu2024multilinguallargelanguagemodels,
      title={Multilingual Large Language Models: A Systematic Survey}, 
      author={Shaolin Zhu and Supryadi and Shaoyang Xu and Haoran Sun and Leiyu Pan and Menglong Cui and Jiangcun Du and Renren Jin and António Branco and Deyi Xiong},
      year={2024},
      eprint={2411.11072},
      archivePrefix={arXiv},
      primaryClass={cs.CL},
      url={https://arxiv.org/abs/2411.11072}, 
}

@inproceedings{deutsch-etal-2025-wmt24,
    title = "{WMT}24++: Expanding the Language Coverage of {WMT}24 to 55 Languages {\&} Dialects",
    author = "Deutsch, Daniel  and
      Briakou, Eleftheria  and
      Caswell, Isaac Rayburn  and
      Finkelstein, Mara  and
      Galor, Rebecca  and
      Juraska, Juraj  and
      Kovacs, Geza  and
      Lui, Alison  and
      Rei, Ricardo  and
      Riesa, Jason  and
      Rijhwani, Shruti  and
      Riley, Parker  and
      Salesky, Elizabeth  and
      Trabelsi, Firas  and
      Winkler, Stephanie  and
      Zhang, Biao  and
      Freitag, Markus",
    editor = "Che, Wanxiang  and
      Nabende, Joyce  and
      Shutova, Ekaterina  and
      Pilehvar, Mohammad Taher",
    booktitle = "Findings of the Association for Computational Linguistics: ACL 2025",
    month = jul,
    year = "2025",
    address = "Vienna, Austria",
    publisher = "Association for Computational Linguistics",
    url = "https://aclanthology.org/2025.findings-acl.634/",
    doi = "10.18653/v1/2025.findings-acl.634",
    pages = "12257--12284",
    ISBN = "979-8-89176-256-5"
}

@inproceedings{federmann-etal-2022-ntrex,
  address = {Online},
  author = {Federmann, Christian and Kocmi, Tom and Xin, Ying},
  booktitle = {Proceedings of the First Workshop on Scaling Up Multilingual Evaluation},
  month = {nov},
  pages = {21--24},
  publisher = {Association for Computational Linguistics},
  title = {{NTREX}-128 {--} News Test References for {MT} Evaluation of 128 Languages},
  url = {https://aclanthology.org/2022.sumeval-1.4},
  year = {2022},
}

@article{enevoldsen2025mmtebmassivemultilingualtext,
  title={MMTEB: Massive Multilingual Text Embedding Benchmark},
  author={Kenneth Enevoldsen and Isaac Chung and Imene Kerboua and Márton Kardos and Ashwin Mathur and David Stap and Jay Gala and Wissam Siblini and Dominik Krzemiński and Genta Indra Winata and Saba Sturua and Saiteja Utpala and Mathieu Ciancone and Marion Schaeffer and Gabriel Sequeira and Diganta Misra and Shreeya Dhakal and Jonathan Rystrøm and Roman Solomatin and Ömer Çağatan and Akash Kundu and Martin Bernstorff and Shitao Xiao and Akshita Sukhlecha and Bhavish Pahwa and Rafał Poświata and Kranthi Kiran GV and Shawon Ashraf and Daniel Auras and Björn Plüster and Jan Philipp Harries and Loïc Magne and Isabelle Mohr and Mariya Hendriksen and Dawei Zhu and Hippolyte Gisserot-Boukhlef and Tom Aarsen and Jan Kostkan and Konrad Wojtasik and Taemin Lee and Marek Šuppa and Crystina Zhang and Roberta Rocca and Mohammed Hamdy and Andrianos Michail and John Yang and Manuel Faysse and Aleksei Vatolin and Nandan Thakur and Manan Dey and Dipam Vasani and Pranjal Chitale and Simone Tedeschi and Nguyen Tai and Artem Snegirev and Michael Günther and Mengzhou Xia and Weijia Shi and Xing Han Lù and Jordan Clive and Gayatri Krishnakumar and Anna Maksimova and Silvan Wehrli and Maria Tikhonova and Henil Panchal and Aleksandr Abramov and Malte Ostendorff and Zheng Liu and Simon Clematide and Lester James Miranda and Alena Fenogenova and Guangyu Song and Ruqiya Bin Safi and Wen-Ding Li and Alessia Borghini and Federico Cassano and Hongjin Su and Jimmy Lin and Howard Yen and Lasse Hansen and Sara Hooker and Chenghao Xiao and Vaibhav Adlakha and Orion Weller and Siva Reddy and Niklas Muennighoff},
  publisher = {arXiv},
  journal={arXiv preprint arXiv:2502.13595},
  year={2025},
  url={https://arxiv.org/abs/2502.13595},
  doi = {10.48550/arXiv.2502.13595},
}

@article{muennighoff2022mteb,
  author = {Muennighoff, Niklas and Tazi, Nouamane and Magne, Lo{\"\i}c and Reimers, Nils},
  title = {MTEB: Massive Text Embedding Benchmark},
  publisher = {arXiv},
  journal={arXiv preprint arXiv:2210.07316},
  year = {2022},
  url = {https://arxiv.org/abs/2210.07316},
  doi = {10.48550/ARXIV.2210.07316},
}

@misc{gemmateam2025gemma3technicalreport,
      title={Gemma 3 Technical Report}, 
      author={Gemma Team and Aishwarya Kamath and Johan Ferret and Shreya Pathak and Nino Vieillard and Ramona Merhej and Sarah Perrin and Tatiana Matejovicova and Alexandre Ramé and Morgane Rivière and Louis Rouillard and Thomas Mesnard and Geoffrey Cideron and Jean-bastien Grill and Sabela Ramos and Edouard Yvinec and Michelle Casbon and Etienne Pot and Ivo Penchev and Gaël Liu and Francesco Visin and Kathleen Kenealy and Lucas Beyer and Xiaohai Zhai and Anton Tsitsulin and Robert Busa-Fekete and Alex Feng and Noveen Sachdeva and Benjamin Coleman and Yi Gao and Basil Mustafa and Iain Barr and Emilio Parisotto and David Tian and Matan Eyal and Colin Cherry and Jan-Thorsten Peter and Danila Sinopalnikov and Surya Bhupatiraju and Rishabh Agarwal and Mehran Kazemi and Dan Malkin and Ravin Kumar and David Vilar and Idan Brusilovsky and Jiaming Luo and Andreas Steiner and Abe Friesen and Abhanshu Sharma and Abheesht Sharma and Adi Mayrav Gilady and Adrian Goedeckemeyer and Alaa Saade and Alex Feng and Alexander Kolesnikov and Alexei Bendebury and Alvin Abdagic and Amit Vadi and András György and André Susano Pinto and Anil Das and Ankur Bapna and Antoine Miech and Antoine Yang and Antonia Paterson and Ashish Shenoy and Ayan Chakrabarti and Bilal Piot and Bo Wu and Bobak Shahriari and Bryce Petrini and Charlie Chen and Charline Le Lan and Christopher A. Choquette-Choo and CJ Carey and Cormac Brick and Daniel Deutsch and Danielle Eisenbud and Dee Cattle and Derek Cheng and Dimitris Paparas and Divyashree Shivakumar Sreepathihalli and Doug Reid and Dustin Tran and Dustin Zelle and Eric Noland and Erwin Huizenga and Eugene Kharitonov and Frederick Liu and Gagik Amirkhanyan and Glenn Cameron and Hadi Hashemi and Hanna Klimczak-Plucińska and Harman Singh and Harsh Mehta and Harshal Tushar Lehri and Hussein Hazimeh and Ian Ballantyne and Idan Szpektor and Ivan Nardini and Jean Pouget-Abadie and Jetha Chan and Joe Stanton and John Wieting and Jonathan Lai and Jordi Orbay and Joseph Fernandez and Josh Newlan and Ju-yeong Ji and Jyotinder Singh and Kat Black and Kathy Yu and Kevin Hui and Kiran Vodrahalli and Klaus Greff and Linhai Qiu and Marcella Valentine and Marina Coelho and Marvin Ritter and Matt Hoffman and Matthew Watson and Mayank Chaturvedi and Michael Moynihan and Min Ma and Nabila Babar and Natasha Noy and Nathan Byrd and Nick Roy and Nikola Momchev and Nilay Chauhan and Noveen Sachdeva and Oskar Bunyan and Pankil Botarda and Paul Caron and Paul Kishan Rubenstein and Phil Culliton and Philipp Schmid and Pier Giuseppe Sessa and Pingmei Xu and Piotr Stanczyk and Pouya Tafti and Rakesh Shivanna and Renjie Wu and Renke Pan and Reza Rokni and Rob Willoughby and Rohith Vallu and Ryan Mullins and Sammy Jerome and Sara Smoot and Sertan Girgin and Shariq Iqbal and Shashir Reddy and Shruti Sheth and Siim Põder and Sijal Bhatnagar and Sindhu Raghuram Panyam and Sivan Eiger and Susan Zhang and Tianqi Liu and Trevor Yacovone and Tyler Liechty and Uday Kalra and Utku Evci and Vedant Misra and Vincent Roseberry and Vlad Feinberg and Vlad Kolesnikov and Woohyun Han and Woosuk Kwon and Xi Chen and Yinlam Chow and Yuvein Zhu and Zichuan Wei and Zoltan Egyed and Victor Cotruta and Minh Giang and Phoebe Kirk and Anand Rao and Kat Black and Nabila Babar and Jessica Lo and Erica Moreira and Luiz Gustavo Martins and Omar Sanseviero and Lucas Gonzalez and Zach Gleicher and Tris Warkentin and Vahab Mirrokni and Evan Senter and Eli Collins and Joelle Barral and Zoubin Ghahramani and Raia Hadsell and Yossi Matias and D. Sculley and Slav Petrov and Noah Fiedel and Noam Shazeer and Oriol Vinyals and Jeff Dean and Demis Hassabis and Koray Kavukcuoglu and Clement Farabet and Elena Buchatskaya and Jean-Baptiste Alayrac and Rohan Anil and Dmitry and Lepikhin and Sebastian Borgeaud and Olivier Bachem and Armand Joulin and Alek Andreev and Cassidy Hardin and Robert Dadashi and Léonard Hussenot},
      year={2025},
      eprint={2503.19786},
      archivePrefix={arXiv},
      primaryClass={cs.CL},
      url={https://arxiv.org/abs/2503.19786}, 
}

@misc{yang2025qwen3technicalreport,
      title={Qwen3 Technical Report}, 
      author={An Yang and Anfeng Li and Baosong Yang and Beichen Zhang and Binyuan Hui and Bo Zheng and Bowen Yu and Chang Gao and Chengen Huang and Chenxu Lv and Chujie Zheng and Dayiheng Liu and Fan Zhou and Fei Huang and Feng Hu and Hao Ge and Haoran Wei and Huan Lin and Jialong Tang and Jian Yang and Jianhong Tu and Jianwei Zhang and Jianxin Yang and Jiaxi Yang and Jing Zhou and Jingren Zhou and Junyang Lin and Kai Dang and Keqin Bao and Kexin Yang and Le Yu and Lianghao Deng and Mei Li and Mingfeng Xue and Mingze Li and Pei Zhang and Peng Wang and Qin Zhu and Rui Men and Ruize Gao and Shixuan Liu and Shuang Luo and Tianhao Li and Tianyi Tang and Wenbiao Yin and Xingzhang Ren and Xinyu Wang and Xinyu Zhang and Xuancheng Ren and Yang Fan and Yang Su and Yichang Zhang and Yinger Zhang and Yu Wan and Yuqiong Liu and Zekun Wang and Zeyu Cui and Zhenru Zhang and Zhipeng Zhou and Zihan Qiu},
      year={2025},
      eprint={2505.09388},
      archivePrefix={arXiv},
      primaryClass={cs.CL},
      url={https://arxiv.org/abs/2505.09388}, 
}

@misc{qwen2025qwen25technicalreport,
      title={Qwen2.5 Technical Report}, 
      author={Qwen and : and An Yang and Baosong Yang and Beichen Zhang and Binyuan Hui and Bo Zheng and Bowen Yu and Chengyuan Li and Dayiheng Liu and Fei Huang and Haoran Wei and Huan Lin and Jian Yang and Jianhong Tu and Jianwei Zhang and Jianxin Yang and Jiaxi Yang and Jingren Zhou and Junyang Lin and Kai Dang and Keming Lu and Keqin Bao and Kexin Yang and Le Yu and Mei Li and Mingfeng Xue and Pei Zhang and Qin Zhu and Rui Men and Runji Lin and Tianhao Li and Tianyi Tang and Tingyu Xia and Xingzhang Ren and Xuancheng Ren and Yang Fan and Yang Su and Yichang Zhang and Yu Wan and Yuqiong Liu and Zeyu Cui and Zhenru Zhang and Zihan Qiu},
      year={2025},
      eprint={2412.15115},
      archivePrefix={arXiv},
      primaryClass={cs.CL},
      url={https://arxiv.org/abs/2412.15115}, 
}

@misc{abdin2024phi4technicalreport,
      title={Phi-4 Technical Report}, 
      author={Marah Abdin and Jyoti Aneja and Harkirat Behl and Sébastien Bubeck and Ronen Eldan and Suriya Gunasekar and Michael Harrison and Russell J. Hewett and Mojan Javaheripi and Piero Kauffmann and James R. Lee and Yin Tat Lee and Yuanzhi Li and Weishung Liu and Caio C. T. Mendes and Anh Nguyen and Eric Price and Gustavo de Rosa and Olli Saarikivi and Adil Salim and Shital Shah and Xin Wang and Rachel Ward and Yue Wu and Dingli Yu and Cyril Zhang and Yi Zhang},
      year={2024},
      eprint={2412.08905},
      archivePrefix={arXiv},
      primaryClass={cs.CL},
      url={https://arxiv.org/abs/2412.08905}, 
}

@misc{deepseekai2025deepseekr1,
      title={DeepSeek-R1: Incentivizing Reasoning Capability in LLMs via Reinforcement Learning}, 
      author={DeepSeek-AI},
      year={2025},
      eprint={2501.12948},
      archivePrefix={arXiv},
      primaryClass={cs.CL},
      url={https://arxiv.org/abs/2501.12948}, 
}

@article{llama3modelcard,
  title={Llama 3 Model Card},
  author={AI@Meta},
  year={2024},
  url = {https://github.com/meta-llama/llama3/blob/main/MODEL_CARD.md},
}

@misc{wu2025bitterlessonlearned2000,
      title={The Bitter Lesson Learned from 2,000+ Multilingual Benchmarks}, 
      author={Minghao Wu and Weixuan Wang and Sinuo Liu and Huifeng Yin and Xintong Wang and Yu Zhao and Chenyang Lyu and Longyue Wang and Weihua Luo and Kaifu Zhang},
      year={2025},
      eprint={2504.15521},
      archivePrefix={arXiv},
      primaryClass={cs.CL},
      url={https://arxiv.org/abs/2504.15521}, 
}

@misc{chang2023surveyevaluationlargelanguage,
      title={A Survey on Evaluation of Large Language Models}, 
      author={Yupeng Chang and Xu Wang and Jindong Wang and Yuan Wu and Linyi Yang and Kaijie Zhu and Hao Chen and Xiaoyuan Yi and Cunxiang Wang and Yidong Wang and Wei Ye and Yue Zhang and Yi Chang and Philip S. Yu and Qiang Yang and Xing Xie},
      year={2023},
      eprint={2307.03109},
      archivePrefix={arXiv},
      primaryClass={cs.CL},
      url={https://arxiv.org/abs/2307.03109}, 
}

@misc{li2024cmmlumeasuringmassivemultitask,
      title={CMMLU: Measuring massive multitask language understanding in Chinese}, 
      author={Haonan Li and Yixuan Zhang and Fajri Koto and Yifei Yang and Hai Zhao and Yeyun Gong and Nan Duan and Timothy Baldwin},
      year={2024},
      eprint={2306.09212},
      archivePrefix={arXiv},
      primaryClass={cs.CL},
      url={https://arxiv.org/abs/2306.09212}, 
}

@inproceedings{yuksel-etal-2024-turkishmmlu,
    title = "{T}urkish{MMLU}: Measuring Massive Multitask Language Understanding in {T}urkish",
    author = {Y{\"u}ksel, Arda  and
      K{\"o}ksal, Abdullatif  and
      Senel, L{\"u}tfi Kerem  and
      Korhonen, Anna  and
      Schuetze, Hinrich},
    editor = "Al-Onaizan, Yaser  and
      Bansal, Mohit  and
      Chen, Yun-Nung",
    booktitle = "Findings of the Association for Computational Linguistics: EMNLP 2024",
    month = nov,
    year = "2024",
    address = "Miami, Florida, USA",
    publisher = "Association for Computational Linguistics",
    url = "https://aclanthology.org/2024.findings-emnlp.413/",
    doi = "10.18653/v1/2024.findings-emnlp.413",
    pages = "7035--7055"
}

@misc{ouyang2022,
      title={Training language models to follow instructions with human feedback}, 
      author={Long Ouyang and Jeff Wu and Xu Jiang and Diogo Almeida and Carroll L. Wainwright and Pamela Mishkin and Chong Zhang and Sandhini Agarwal and Katarina Slama and Alex Ray and John Schulman and Jacob Hilton and Fraser Kelton and Luke Miller and Maddie Simens and Amanda Askell and Peter Welinder and Paul Christiano and Jan Leike and Ryan Lowe},
      year={2022},
      eprint={2203.02155},
      archivePrefix={arXiv},
      primaryClass={cs.CL},
      url={https://arxiv.org/abs/2203.02155}, 
}

@misc{openai2024gpt4technicalreport,
      title={GPT-4 Technical Report}, 
      author={OpenAI and Josh Achiam and Steven Adler and Sandhini Agarwal and Lama Ahmad and Ilge Akkaya and Florencia Leoni Aleman and Diogo Almeida and Janko Altenschmidt and Sam Altman and Shyamal Anadkat and Red Avila and Igor Babuschkin and Suchir Balaji and Valerie Balcom and Paul Baltescu and Haiming Bao and Mohammad Bavarian and Jeff Belgum and Irwan Bello and Jake Berdine and Gabriel Bernadett-Shapiro and Christopher Berner and Lenny Bogdonoff and Oleg Boiko and Madelaine Boyd and Anna-Luisa Brakman and Greg Brockman and Tim Brooks and Miles Brundage and Kevin Button and Trevor Cai and Rosie Campbell and Andrew Cann and Brittany Carey and Chelsea Carlson and Rory Carmichael and Brooke Chan and Che Chang and Fotis Chantzis and Derek Chen and Sully Chen and Ruby Chen and Jason Chen and Mark Chen and Ben Chess and Chester Cho and Casey Chu and Hyung Won Chung and Dave Cummings and Jeremiah Currier and Yunxing Dai and Cory Decareaux and Thomas Degry and Noah Deutsch and Damien Deville and Arka Dhar and David Dohan and Steve Dowling and Sheila Dunning and Adrien Ecoffet and Atty Eleti and Tyna Eloundou and David Farhi and Liam Fedus and Niko Felix and Simón Posada Fishman and Juston Forte and Isabella Fulford and Leo Gao and Elie Georges and Christian Gibson and Vik Goel and Tarun Gogineni and Gabriel Goh and Rapha Gontijo-Lopes and Jonathan Gordon and Morgan Grafstein and Scott Gray and Ryan Greene and Joshua Gross and Shixiang Shane Gu and Yufei Guo and Chris Hallacy and Jesse Han and Jeff Harris and Yuchen He and Mike Heaton and Johannes Heidecke and Chris Hesse and Alan Hickey and Wade Hickey and Peter Hoeschele and Brandon Houghton and Kenny Hsu and Shengli Hu and Xin Hu and Joost Huizinga and Shantanu Jain and Shawn Jain and Joanne Jang and Angela Jiang and Roger Jiang and Haozhun Jin and Denny Jin and Shino Jomoto and Billie Jonn and Heewoo Jun and Tomer Kaftan and Łukasz Kaiser and Ali Kamali and Ingmar Kanitscheider and Nitish Shirish Keskar and Tabarak Khan and Logan Kilpatrick and Jong Wook Kim and Christina Kim and Yongjik Kim and Jan Hendrik Kirchner and Jamie Kiros and Matt Knight and Daniel Kokotajlo and Łukasz Kondraciuk and Andrew Kondrich and Aris Konstantinidis and Kyle Kosic and Gretchen Krueger and Vishal Kuo and Michael Lampe and Ikai Lan and Teddy Lee and Jan Leike and Jade Leung and Daniel Levy and Chak Ming Li and Rachel Lim and Molly Lin and Stephanie Lin and Mateusz Litwin and Theresa Lopez and Ryan Lowe and Patricia Lue and Anna Makanju and Kim Malfacini and Sam Manning and Todor Markov and Yaniv Markovski and Bianca Martin and Katie Mayer and Andrew Mayne and Bob McGrew and Scott Mayer McKinney and Christine McLeavey and Paul McMillan and Jake McNeil and David Medina and Aalok Mehta and Jacob Menick and Luke Metz and Andrey Mishchenko and Pamela Mishkin and Vinnie Monaco and Evan Morikawa and Daniel Mossing and Tong Mu and Mira Murati and Oleg Murk and David Mély and Ashvin Nair and Reiichiro Nakano and Rajeev Nayak and Arvind Neelakantan and Richard Ngo and Hyeonwoo Noh and Long Ouyang and Cullen O'Keefe and Jakub Pachocki and Alex Paino and Joe Palermo and Ashley Pantuliano and Giambattista Parascandolo and Joel Parish and Emy Parparita and Alex Passos and Mikhail Pavlov and Andrew Peng and Adam Perelman and Filipe de Avila Belbute Peres and Michael Petrov and Henrique Ponde de Oliveira Pinto and Michael and Pokorny and Michelle Pokrass and Vitchyr H. Pong and Tolly Powell and Alethea Power and Boris Power and Elizabeth Proehl and Raul Puri and Alec Radford and Jack Rae and Aditya Ramesh and Cameron Raymond and Francis Real and Kendra Rimbach and Carl Ross and Bob Rotsted and Henri Roussez and Nick Ryder and Mario Saltarelli and Ted Sanders and Shibani Santurkar and Girish Sastry and Heather Schmidt and David Schnurr and John Schulman and Daniel Selsam and Kyla Sheppard and Toki Sherbakov and Jessica Shieh and Sarah Shoker and Pranav Shyam and Szymon Sidor and Eric Sigler and Maddie Simens and Jordan Sitkin and Katarina Slama and Ian Sohl and Benjamin Sokolowsky and Yang Song and Natalie Staudacher and Felipe Petroski Such and Natalie Summers and Ilya Sutskever and Jie Tang and Nikolas Tezak and Madeleine B. Thompson and Phil Tillet and Amin Tootoonchian and Elizabeth Tseng and Preston Tuggle and Nick Turley and Jerry Tworek and Juan Felipe Cerón Uribe and Andrea Vallone and Arun Vijayvergiya and Chelsea Voss and Carroll Wainwright and Justin Jay Wang and Alvin Wang and Ben Wang and Jonathan Ward and Jason Wei and CJ Weinmann and Akila Welihinda and Peter Welinder and Jiayi Weng and Lilian Weng and Matt Wiethoff and Dave Willner and Clemens Winter and Samuel Wolrich and Hannah Wong and Lauren Workman and Sherwin Wu and Jeff Wu and Michael Wu and Kai Xiao and Tao Xu and Sarah Yoo and Kevin Yu and Qiming Yuan and Wojciech Zaremba and Rowan Zellers and Chong Zhang and Marvin Zhang and Shengjia Zhao and Tianhao Zheng and Juntang Zhuang and William Zhuk and Barret Zoph},
      year={2024},
      eprint={2303.08774},
      archivePrefix={arXiv},
      primaryClass={cs.CL},
      url={https://arxiv.org/abs/2303.08774}, 
}

@misc{touvron2023llamaopenefficientfoundation,
      title={LLaMA: Open and Efficient Foundation Language Models}, 
      author={Hugo Touvron and Thibaut Lavril and Gautier Izacard and Xavier Martinet and Marie-Anne Lachaux and Timothée Lacroix and Baptiste Rozière and Naman Goyal and Eric Hambro and Faisal Azhar and Aurelien Rodriguez and Armand Joulin and Edouard Grave and Guillaume Lample},
      year={2023},
      eprint={2302.13971},
      archivePrefix={arXiv},
      primaryClass={cs.CL},
      url={https://arxiv.org/abs/2302.13971}, 
}

@article{Xu_2025,
   title={A survey on multilingual large language models: corpora, alignment, and bias},
   volume={19},
   ISSN={2095-2236},
   url={http://dx.doi.org/10.1007/s11704-024-40579-4},
   DOI={10.1007/s11704-024-40579-4},
   number={11},
   journal={Frontiers of Computer Science},
   publisher={Springer Science and Business Media LLC},
   author={Xu, Yuemei and Hu, Ling and Zhao, Jiayi and Qiu, Zihan and Xu, Kexin and Ye, Yuqi and Gu, Hanwen},
   year={2025},
   month=apr 
}

@inproceedings{hada-etal-2024-large,
    title = "Are Large Language Model-based Evaluators the Solution to Scaling Up Multilingual Evaluation?",
    author = "Hada, Rishav  and
      Gumma, Varun  and
      de Wynter, Adrian  and
      Diddee, Harshita  and
      Ahmed, Mohamed  and
      Choudhury, Monojit  and
      Bali, Kalika  and
      Sitaram, Sunayana",
    editor = "Graham, Yvette  and
      Purver, Matthew",
    booktitle = "Findings of the Association for Computational Linguistics: EACL 2024",
    month = mar,
    year = "2024",
    address = "St. Julian{'}s, Malta",
    publisher = "Association for Computational Linguistics",
    url = "https://aclanthology.org/2024.findings-eacl.71/",
    pages = "1051--1070"
}

@misc{brown2020languagemodelsfewshotlearners,
      title={Language Models are Few-Shot Learners}, 
      author={Tom B. Brown and Benjamin Mann and Nick Ryder and Melanie Subbiah and Jared Kaplan and Prafulla Dhariwal and Arvind Neelakantan and Pranav Shyam and Girish Sastry and Amanda Askell and Sandhini Agarwal and Ariel Herbert-Voss and Gretchen Krueger and Tom Henighan and Rewon Child and Aditya Ramesh and Daniel M. Ziegler and Jeffrey Wu and Clemens Winter and Christopher Hesse and Mark Chen and Eric Sigler and Mateusz Litwin and Scott Gray and Benjamin Chess and Jack Clark and Christopher Berner and Sam McCandlish and Alec Radford and Ilya Sutskever and Dario Amodei},
      year={2020},
      eprint={2005.14165},
      archivePrefix={arXiv},
      primaryClass={cs.CL},
      url={https://arxiv.org/abs/2005.14165}, 
}

@misc{ouyang2022traininglanguagemodelsfollow,
      title={Training language models to follow instructions with human feedback}, 
      author={Long Ouyang and Jeff Wu and Xu Jiang and Diogo Almeida and Carroll L. Wainwright and Pamela Mishkin and Chong Zhang and Sandhini Agarwal and Katarina Slama and Alex Ray and John Schulman and Jacob Hilton and Fraser Kelton and Luke Miller and Maddie Simens and Amanda Askell and Peter Welinder and Paul Christiano and Jan Leike and Ryan Lowe},
      year={2022},
      eprint={2203.02155},
      archivePrefix={arXiv},
      primaryClass={cs.CL},
      url={https://arxiv.org/abs/2203.02155}, 
}

@misc{bai2022constitutionalaiharmlessnessai,
      title={Constitutional AI: Harmlessness from AI Feedback}, 
      author={Yuntao Bai and Saurav Kadavath and Sandipan Kundu and Amanda Askell and Jackson Kernion and Andy Jones and Anna Chen and Anna Goldie and Azalia Mirhoseini and Cameron McKinnon and Carol Chen and Catherine Olsson and Christopher Olah and Danny Hernandez and Dawn Drain and Deep Ganguli and Dustin Li and Eli Tran-Johnson and Ethan Perez and Jamie Kerr and Jared Mueller and Jeffrey Ladish and Joshua Landau and Kamal Ndousse and Kamile Lukosuite and Liane Lovitt and Michael Sellitto and Nelson Elhage and Nicholas Schiefer and Noemi Mercado and Nova DasSarma and Robert Lasenby and Robin Larson and Sam Ringer and Scott Johnston and Shauna Kravec and Sheer El Showk and Stanislav Fort and Tamera Lanham and Timothy Telleen-Lawton and Tom Conerly and Tom Henighan and Tristan Hume and Samuel R. Bowman and Zac Hatfield-Dodds and Ben Mann and Dario Amodei and Nicholas Joseph and Sam McCandlish and Tom Brown and Jared Kaplan},
      year={2022},
      eprint={2212.08073},
      archivePrefix={arXiv},
      primaryClass={cs.CL},
      url={https://arxiv.org/abs/2212.08073}, 
}

@misc{geminiteam2025geminifamilyhighlycapable,
      title={Gemini: A Family of Highly Capable Multimodal Models}, 
      author={Gemini Team and Rohan Anil and Sebastian Borgeaud and Jean-Baptiste Alayrac and Jiahui Yu and Radu Soricut and Johan Schalkwyk and Andrew M. Dai and Anja Hauth and Katie Millican and David Silver and Melvin Johnson and Ioannis Antonoglou and Julian Schrittwieser and Amelia Glaese and Jilin Chen and Emily Pitler and Timothy Lillicrap and Angeliki Lazaridou and Orhan Firat and James Molloy and Michael Isard and Paul R. Barham and Tom Hennigan and Benjamin Lee and Fabio Viola and Malcolm Reynolds and Yuanzhong Xu and Ryan Doherty and Eli Collins and Clemens Meyer and Eliza Rutherford and Erica Moreira and Kareem Ayoub and Megha Goel and Jack Krawczyk and Cosmo Du and Ed Chi and Heng-Tze Cheng and Eric Ni and Purvi Shah and Patrick Kane and Betty Chan and Manaal Faruqui and Aliaksei Severyn and Hanzhao Lin and YaGuang Li and Yong Cheng and Abe Ittycheriah and Mahdis Mahdieh and Mia Chen and Pei Sun and Dustin Tran and Sumit Bagri and Balaji Lakshminarayanan and Jeremiah Liu and Andras Orban and Fabian Güra and Hao Zhou and Xinying Song and Aurelien Boffy and Harish Ganapathy and Steven Zheng and HyunJeong Choe and Ágoston Weisz and Tao Zhu and Yifeng Lu and Siddharth Gopal and Jarrod Kahn and Maciej Kula and Jeff Pitman and Rushin Shah and Emanuel Taropa and Majd Al Merey and Martin Baeuml and Zhifeng Chen and Laurent El Shafey and Yujing Zhang and Olcan Sercinoglu and George Tucker and Enrique Piqueras and Maxim Krikun and Iain Barr and Nikolay Savinov and Ivo Danihelka and Becca Roelofs and Anaïs White and Anders Andreassen and Tamara von Glehn and Lakshman Yagati and Mehran Kazemi and Lucas Gonzalez and Misha Khalman and Jakub Sygnowski and Alexandre Frechette and Charlotte Smith and Laura Culp and Lev Proleev and Yi Luan and Xi Chen and James Lottes and Nathan Schucher and Federico Lebron and Alban Rrustemi and Natalie Clay and Phil Crone and Tomas Kocisky and Jeffrey Zhao and Bartek Perz and Dian Yu and Heidi Howard and Adam Bloniarz and Jack W. Rae and Han Lu and Laurent Sifre and Marcello Maggioni and Fred Alcober and Dan Garrette and Megan Barnes and Shantanu Thakoor and Jacob Austin and Gabriel Barth-Maron and William Wong and Rishabh Joshi and Rahma Chaabouni and Deeni Fatiha and Arun Ahuja and Gaurav Singh Tomar and Evan Senter and Martin Chadwick and Ilya Kornakov and Nithya Attaluri and Iñaki Iturrate and Ruibo Liu and Yunxuan Li and Sarah Cogan and Jeremy Chen and Chao Jia and Chenjie Gu and Qiao Zhang and Jordan Grimstad and Ale Jakse Hartman and Xavier Garcia and Thanumalayan Sankaranarayana Pillai and Jacob Devlin and Michael Laskin and Diego de Las Casas and Dasha Valter and Connie Tao and Lorenzo Blanco and Adrià Puigdomènech Badia and David Reitter and Mianna Chen and Jenny Brennan and Clara Rivera and Sergey Brin and Shariq Iqbal and Gabriela Surita and Jane Labanowski and Abhi Rao and Stephanie Winkler and Emilio Parisotto and Yiming Gu and Kate Olszewska and Ravi Addanki and Antoine Miech and Annie Louis and Denis Teplyashin and Geoff Brown and Elliot Catt and Jan Balaguer and Jackie Xiang and Pidong Wang and Zoe Ashwood and Anton Briukhov and Albert Webson and Sanjay Ganapathy and Smit Sanghavi and Ajay Kannan and Ming-Wei Chang and Axel Stjerngren and Josip Djolonga and Yuting Sun and Ankur Bapna and Matthew Aitchison and Pedram Pejman and Henryk Michalewski and Tianhe Yu and Cindy Wang and Juliette Love and Junwhan Ahn and Dawn Bloxwich and Kehang Han and Peter Humphreys and Thibault Sellam and James Bradbury and Varun Godbole and Sina Samangooei and Bogdan Damoc and Alex Kaskasoli and Sébastien M. R. Arnold and Vijay Vasudevan and Shubham Agrawal and Jason Riesa and Dmitry Lepikhin and Richard Tanburn and Srivatsan Srinivasan and Hyeontaek Lim and Sarah Hodkinson and Pranav Shyam and Johan Ferret and Steven Hand and Ankush Garg and Tom Le Paine and Jian Li and Yujia Li and Minh Giang and Alexander Neitz and Zaheer Abbas and Sarah York and Machel Reid and Elizabeth Cole and Aakanksha Chowdhery and Dipanjan Das and Dominika Rogozińska and Vitaliy Nikolaev and Pablo Sprechmann and Zachary Nado and Lukas Zilka and Flavien Prost and Luheng He and Marianne Monteiro and Gaurav Mishra and Chris Welty and Josh Newlan and Dawei Jia and Miltiadis Allamanis and Clara Huiyi Hu and Raoul de Liedekerke and Justin Gilmer and Carl Saroufim and Shruti Rijhwani and Shaobo Hou and Disha Shrivastava and Anirudh Baddepudi and Alex Goldin and Adnan Ozturel and Albin Cassirer and Yunhan Xu and Daniel Sohn and Devendra Sachan and Reinald Kim Amplayo and Craig Swanson and Dessie Petrova and Shashi Narayan and Arthur Guez and Siddhartha Brahma and Jessica Landon and Miteyan Patel and Ruizhe Zhao and Kevin Villela and Luyu Wang and Wenhao Jia and Matthew Rahtz and Mai Giménez and Legg Yeung and James Keeling and Petko Georgiev and Diana Mincu and Boxi Wu and Salem Haykal and Rachel Saputro and Kiran Vodrahalli and James Qin and Zeynep Cankara and Abhanshu Sharma and Nick Fernando and Will Hawkins and Behnam Neyshabur and Solomon Kim and Adrian Hutter and Priyanka Agrawal and Alex Castro-Ros and George van den Driessche and Tao Wang and Fan Yang and Shuo-yiin Chang and Paul Komarek and Ross McIlroy and Mario Lučić and Guodong Zhang and Wael Farhan and Michael Sharman and Paul Natsev and Paul Michel and Yamini Bansal and Siyuan Qiao and Kris Cao and Siamak Shakeri and Christina Butterfield and Justin Chung and Paul Kishan Rubenstein and Shivani Agrawal and Arthur Mensch and Kedar Soparkar and Karel Lenc and Timothy Chung and Aedan Pope and Loren Maggiore and Jackie Kay and Priya Jhakra and Shibo Wang and Joshua Maynez and Mary Phuong and Taylor Tobin and Andrea Tacchetti and Maja Trebacz and Kevin Robinson and Yash Katariya and Sebastian Riedel and Paige Bailey and Kefan Xiao and Nimesh Ghelani and Lora Aroyo and Ambrose Slone and Neil Houlsby and Xuehan Xiong and Zhen Yang and Elena Gribovskaya and Jonas Adler and Mateo Wirth and Lisa Lee and Music Li and Thais Kagohara and Jay Pavagadhi and Sophie Bridgers and Anna Bortsova and Sanjay Ghemawat and Zafarali Ahmed and Tianqi Liu and Richard Powell and Vijay Bolina and Mariko Iinuma and Polina Zablotskaia and James Besley and Da-Woon Chung and Timothy Dozat and Ramona Comanescu and Xiance Si and Jeremy Greer and Guolong Su and Martin Polacek and Raphaël Lopez Kaufman and Simon Tokumine and Hexiang Hu and Elena Buchatskaya and Yingjie Miao and Mohamed Elhawaty and Aditya Siddhant and Nenad Tomasev and Jinwei Xing and Christina Greer and Helen Miller and Shereen Ashraf and Aurko Roy and Zizhao Zhang and Ada Ma and Angelos Filos and Milos Besta and Rory Blevins and Ted Klimenko and Chih-Kuan Yeh and Soravit Changpinyo and Jiaqi Mu and Oscar Chang and Mantas Pajarskas and Carrie Muir and Vered Cohen and Charline Le Lan and Krishna Haridasan and Amit Marathe and Steven Hansen and Sholto Douglas and Rajkumar Samuel and Mingqiu Wang and Sophia Austin and Chang Lan and Jiepu Jiang and Justin Chiu and Jaime Alonso Lorenzo and Lars Lowe Sjösund and Sébastien Cevey and Zach Gleicher and Thi Avrahami and Anudhyan Boral and Hansa Srinivasan and Vittorio Selo and Rhys May and Konstantinos Aisopos and Léonard Hussenot and Livio Baldini Soares and Kate Baumli and Michael B. Chang and Adrià Recasens and Ben Caine and Alexander Pritzel and Filip Pavetic and Fabio Pardo and Anita Gergely and Justin Frye and Vinay Ramasesh and Dan Horgan and Kartikeya Badola and Nora Kassner and Subhrajit Roy and Ethan Dyer and Víctor Campos Campos and Alex Tomala and Yunhao Tang and Dalia El Badawy and Elspeth White and Basil Mustafa and Oran Lang and Abhishek Jindal and Sharad Vikram and Zhitao Gong and Sergi Caelles and Ross Hemsley and Gregory Thornton and Fangxiaoyu Feng and Wojciech Stokowiec and Ce Zheng and Phoebe Thacker and Çağlar Ünlü and Zhishuai Zhang and Mohammad Saleh and James Svensson and Max Bileschi and Piyush Patil and Ankesh Anand and Roman Ring and Katerina Tsihlas and Arpi Vezer and Marco Selvi and Toby Shevlane and Mikel Rodriguez and Tom Kwiatkowski and Samira Daruki and Keran Rong and Allan Dafoe and Nicholas FitzGerald and Keren Gu-Lemberg and Mina Khan and Lisa Anne Hendricks and Marie Pellat and Vladimir Feinberg and James Cobon-Kerr and Tara Sainath and Maribeth Rauh and Sayed Hadi Hashemi and Richard Ives and Yana Hasson and Eric Noland and Yuan Cao and Nathan Byrd and Le Hou and Qingze Wang and Thibault Sottiaux and Michela Paganini and Jean-Baptiste Lespiau and Alexandre Moufarek and Samer Hassan and Kaushik Shivakumar and Joost van Amersfoort and Amol Mandhane and Pratik Joshi and Anirudh Goyal and Matthew Tung and Andrew Brock and Hannah Sheahan and Vedant Misra and Cheng Li and Nemanja Rakićević and Mostafa Dehghani and Fangyu Liu and Sid Mittal and Junhyuk Oh and Seb Noury and Eren Sezener and Fantine Huot and Matthew Lamm and Nicola De Cao and Charlie Chen and Sidharth Mudgal and Romina Stella and Kevin Brooks and Gautam Vasudevan and Chenxi Liu and Mainak Chain and Nivedita Melinkeri and Aaron Cohen and Venus Wang and Kristie Seymore and Sergey Zubkov and Rahul Goel and Summer Yue and Sai Krishnakumaran and Brian Albert and Nate Hurley and Motoki Sano and Anhad Mohananey and Jonah Joughin and Egor Filonov and Tomasz Kępa and Yomna Eldawy and Jiawern Lim and Rahul Rishi and Shirin Badiezadegan and Taylor Bos and Jerry Chang and Sanil Jain and Sri Gayatri Sundara Padmanabhan and Subha Puttagunta and Kalpesh Krishna and Leslie Baker and Norbert Kalb and Vamsi Bedapudi and Adam Kurzrok and Shuntong Lei and Anthony Yu and Oren Litvin and Xiang Zhou and Zhichun Wu and Sam Sobell and Andrea Siciliano and Alan Papir and Robby Neale and Jonas Bragagnolo and Tej Toor and Tina Chen and Valentin Anklin and Feiran Wang and Richie Feng and Milad Gholami and Kevin Ling and Lijuan Liu and Jules Walter and Hamid Moghaddam and Arun Kishore and Jakub Adamek and Tyler Mercado and Jonathan Mallinson and Siddhinita Wandekar and Stephen Cagle and Eran Ofek and Guillermo Garrido and Clemens Lombriser and Maksim Mukha and Botu Sun and Hafeezul Rahman Mohammad and Josip Matak and Yadi Qian and Vikas Peswani and Pawel Janus and Quan Yuan and Leif Schelin and Oana David and Ankur Garg and Yifan He and Oleksii Duzhyi and Anton Älgmyr and Timothée Lottaz and Qi Li and Vikas Yadav and Luyao Xu and Alex Chinien and Rakesh Shivanna and Aleksandr Chuklin and Josie Li and Carrie Spadine and Travis Wolfe and Kareem Mohamed and Subhabrata Das and Zihang Dai and Kyle He and Daniel von Dincklage and Shyam Upadhyay and Akanksha Maurya and Luyan Chi and Sebastian Krause and Khalid Salama and Pam G Rabinovitch and Pavan Kumar Reddy M and Aarush Selvan and Mikhail Dektiarev and Golnaz Ghiasi and Erdem Guven and Himanshu Gupta and Boyi Liu and Deepak Sharma and Idan Heimlich Shtacher and Shachi Paul and Oscar Akerlund and François-Xavier Aubet and Terry Huang and Chen Zhu and Eric Zhu and Elico Teixeira and Matthew Fritze and Francesco Bertolini and Liana-Eleonora Marinescu and Martin Bölle and Dominik Paulus and Khyatti Gupta and Tejasi Latkar and Max Chang and Jason Sanders and Roopa Wilson and Xuewei Wu and Yi-Xuan Tan and Lam Nguyen Thiet and Tulsee Doshi and Sid Lall and Swaroop Mishra and Wanming Chen and Thang Luong and Seth Benjamin and Jasmine Lee and Ewa Andrejczuk and Dominik Rabiej and Vipul Ranjan and Krzysztof Styrc and Pengcheng Yin and Jon Simon and Malcolm Rose Harriott and Mudit Bansal and Alexei Robsky and Geoff Bacon and David Greene and Daniil Mirylenka and Chen Zhou and Obaid Sarvana and Abhimanyu Goyal and Samuel Andermatt and Patrick Siegler and Ben Horn and Assaf Israel and Francesco Pongetti and Chih-Wei "Louis" Chen and Marco Selvatici and Pedro Silva and Kathie Wang and Jackson Tolins and Kelvin Guu and Roey Yogev and Xiaochen Cai and Alessandro Agostini and Maulik Shah and Hung Nguyen and Noah Ó Donnaile and Sébastien Pereira and Linda Friso and Adam Stambler and Adam Kurzrok and Chenkai Kuang and Yan Romanikhin and Mark Geller and ZJ Yan and Kane Jang and Cheng-Chun Lee and Wojciech Fica and Eric Malmi and Qijun Tan and Dan Banica and Daniel Balle and Ryan Pham and Yanping Huang and Diana Avram and Hongzhi Shi and Jasjot Singh and Chris Hidey and Niharika Ahuja and Pranab Saxena and Dan Dooley and Srividya Pranavi Potharaju and Eileen O'Neill and Anand Gokulchandran and Ryan Foley and Kai Zhao and Mike Dusenberry and Yuan Liu and Pulkit Mehta and Ragha Kotikalapudi and Chalence Safranek-Shrader and Andrew Goodman and Joshua Kessinger and Eran Globen and Prateek Kolhar and Chris Gorgolewski and Ali Ibrahim and Yang Song and Ali Eichenbaum and Thomas Brovelli and Sahitya Potluri and Preethi Lahoti and Cip Baetu and Ali Ghorbani and Charles Chen and Andy Crawford and Shalini Pal and Mukund Sridhar and Petru Gurita and Asier Mujika and Igor Petrovski and Pierre-Louis Cedoz and Chenmei Li and Shiyuan Chen and Niccolò Dal Santo and Siddharth Goyal and Jitesh Punjabi and Karthik Kappaganthu and Chester Kwak and Pallavi LV and Sarmishta Velury and Himadri Choudhury and Jamie Hall and Premal Shah and Ricardo Figueira and Matt Thomas and Minjie Lu and Ting Zhou and Chintu Kumar and Thomas Jurdi and Sharat Chikkerur and Yenai Ma and Adams Yu and Soo Kwak and Victor Ähdel and Sujeevan Rajayogam and Travis Choma and Fei Liu and Aditya Barua and Colin Ji and Ji Ho Park and Vincent Hellendoorn and Alex Bailey and Taylan Bilal and Huanjie Zhou and Mehrdad Khatir and Charles Sutton and Wojciech Rzadkowski and Fiona Macintosh and Roopali Vij and Konstantin Shagin and Paul Medina and Chen Liang and Jinjing Zhou and Pararth Shah and Yingying Bi and Attila Dankovics and Shipra Banga and Sabine Lehmann and Marissa Bredesen and Zifan Lin and John Eric Hoffmann and Jonathan Lai and Raynald Chung and Kai Yang and Nihal Balani and Arthur Bražinskas and Andrei Sozanschi and Matthew Hayes and Héctor Fernández Alcalde and Peter Makarov and Will Chen and Antonio Stella and Liselotte Snijders and Michael Mandl and Ante Kärrman and Paweł Nowak and Xinyi Wu and Alex Dyck and Krishnan Vaidyanathan and Raghavender R and Jessica Mallet and Mitch Rudominer and Eric Johnston and Sushil Mittal and Akhil Udathu and Janara Christensen and Vishal Verma and Zach Irving and Andreas Santucci and Gamaleldin Elsayed and Elnaz Davoodi and Marin Georgiev and Ian Tenney and Nan Hua and Geoffrey Cideron and Edouard Leurent and Mahmoud Alnahlawi and Ionut Georgescu and Nan Wei and Ivy Zheng and Dylan Scandinaro and Heinrich Jiang and Jasper Snoek and Mukund Sundararajan and Xuezhi Wang and Zack Ontiveros and Itay Karo and Jeremy Cole and Vinu Rajashekhar and Lara Tumeh and Eyal Ben-David and Rishub Jain and Jonathan Uesato and Romina Datta and Oskar Bunyan and Shimu Wu and John Zhang and Piotr Stanczyk and Ye Zhang and David Steiner and Subhajit Naskar and Michael Azzam and Matthew Johnson and Adam Paszke and Chung-Cheng Chiu and Jaume Sanchez Elias and Afroz Mohiuddin and Faizan Muhammad and Jin Miao and Andrew Lee and Nino Vieillard and Jane Park and Jiageng Zhang and Jeff Stanway and Drew Garmon and Abhijit Karmarkar and Zhe Dong and Jong Lee and Aviral Kumar and Luowei Zhou and Jonathan Evens and William Isaac and Geoffrey Irving and Edward Loper and Michael Fink and Isha Arkatkar and Nanxin Chen and Izhak Shafran and Ivan Petrychenko and Zhe Chen and Johnson Jia and Anselm Levskaya and Zhenkai Zhu and Peter Grabowski and Yu Mao and Alberto Magni and Kaisheng Yao and Javier Snaider and Norman Casagrande and Evan Palmer and Paul Suganthan and Alfonso Castaño and Irene Giannoumis and Wooyeol Kim and Mikołaj Rybiński and Ashwin Sreevatsa and Jennifer Prendki and David Soergel and Adrian Goedeckemeyer and Willi Gierke and Mohsen Jafari and Meenu Gaba and Jeremy Wiesner and Diana Gage Wright and Yawen Wei and Harsha Vashisht and Yana Kulizhskaya and Jay Hoover and Maigo Le and Lu Li and Chimezie Iwuanyanwu and Lu Liu and Kevin Ramirez and Andrey Khorlin and Albert Cui and Tian LIN and Marcus Wu and Ricardo Aguilar and Keith Pallo and Abhishek Chakladar and Ginger Perng and Elena Allica Abellan and Mingyang Zhang and Ishita Dasgupta and Nate Kushman and Ivo Penchev and Alena Repina and Xihui Wu and Tom van der Weide and Priya Ponnapalli and Caroline Kaplan and Jiri Simsa and Shuangfeng Li and Olivier Dousse and Fan Yang and Jeff Piper and Nathan Ie and Rama Pasumarthi and Nathan Lintz and Anitha Vijayakumar and Daniel Andor and Pedro Valenzuela and Minnie Lui and Cosmin Paduraru and Daiyi Peng and Katherine Lee and Shuyuan Zhang and Somer Greene and Duc Dung Nguyen and Paula Kurylowicz and Cassidy Hardin and Lucas Dixon and Lili Janzer and Kiam Choo and Ziqiang Feng and Biao Zhang and Achintya Singhal and Dayou Du and Dan McKinnon and Natasha Antropova and Tolga Bolukbasi and Orgad Keller and David Reid and Daniel Finchelstein and Maria Abi Raad and Remi Crocker and Peter Hawkins and Robert Dadashi and Colin Gaffney and Ken Franko and Anna Bulanova and Rémi Leblond and Shirley Chung and Harry Askham and Luis C. Cobo and Kelvin Xu and Felix Fischer and Jun Xu and Christina Sorokin and Chris Alberti and Chu-Cheng Lin and Colin Evans and Alek Dimitriev and Hannah Forbes and Dylan Banarse and Zora Tung and Mark Omernick and Colton Bishop and Rachel Sterneck and Rohan Jain and Jiawei Xia and Ehsan Amid and Francesco Piccinno and Xingyu Wang and Praseem Banzal and Daniel J. Mankowitz and Alex Polozov and Victoria Krakovna and Sasha Brown and MohammadHossein Bateni and Dennis Duan and Vlad Firoiu and Meghana Thotakuri and Tom Natan and Matthieu Geist and Ser tan Girgin and Hui Li and Jiayu Ye and Ofir Roval and Reiko Tojo and Michael Kwong and James Lee-Thorp and Christopher Yew and Danila Sinopalnikov and Sabela Ramos and John Mellor and Abhishek Sharma and Kathy Wu and David Miller and Nicolas Sonnerat and Denis Vnukov and Rory Greig and Jennifer Beattie and Emily Caveness and Libin Bai and Julian Eisenschlos and Alex Korchemniy and Tomy Tsai and Mimi Jasarevic and Weize Kong and Phuong Dao and Zeyu Zheng and Frederick Liu and Fan Yang and Rui Zhu and Tian Huey Teh and Jason Sanmiya and Evgeny Gladchenko and Nejc Trdin and Daniel Toyama and Evan Rosen and Sasan Tavakkol and Linting Xue and Chen Elkind and Oliver Woodman and John Carpenter and George Papamakarios and Rupert Kemp and Sushant Kafle and Tanya Grunina and Rishika Sinha and Alice Talbert and Diane Wu and Denese Owusu-Afriyie and Cosmo Du and Chloe Thornton and Jordi Pont-Tuset and Pradyumna Narayana and Jing Li and Saaber Fatehi and John Wieting and Omar Ajmeri and Benigno Uria and Yeongil Ko and Laura Knight and Amélie Héliou and Ning Niu and Shane Gu and Chenxi Pang and Yeqing Li and Nir Levine and Ariel Stolovich and Rebeca Santamaria-Fernandez and Sonam Goenka and Wenny Yustalim and Robin Strudel and Ali Elqursh and Charlie Deck and Hyo Lee and Zonglin Li and Kyle Levin and Raphael Hoffmann and Dan Holtmann-Rice and Olivier Bachem and Sho Arora and Christy Koh and Soheil Hassas Yeganeh and Siim Põder and Mukarram Tariq and Yanhua Sun and Lucian Ionita and Mojtaba Seyedhosseini and Pouya Tafti and Zhiyu Liu and Anmol Gulati and Jasmine Liu and Xinyu Ye and Bart Chrzaszcz and Lily Wang and Nikhil Sethi and Tianrun Li and Ben Brown and Shreya Singh and Wei Fan and Aaron Parisi and Joe Stanton and Vinod Koverkathu and Christopher A. Choquette-Choo and Yunjie Li and TJ Lu and Abe Ittycheriah and Prakash Shroff and Mani Varadarajan and Sanaz Bahargam and Rob Willoughby and David Gaddy and Guillaume Desjardins and Marco Cornero and Brona Robenek and Bhavishya Mittal and Ben Albrecht and Ashish Shenoy and Fedor Moiseev and Henrik Jacobsson and Alireza Ghaffarkhah and Morgane Rivière and Alanna Walton and Clément Crepy and Alicia Parrish and Zongwei Zhou and Clement Farabet and Carey Radebaugh and Praveen Srinivasan and Claudia van der Salm and Andreas Fidjeland and Salvatore Scellato and Eri Latorre-Chimoto and Hanna Klimczak-Plucińska and David Bridson and Dario de Cesare and Tom Hudson and Piermaria Mendolicchio and Lexi Walker and Alex Morris and Matthew Mauger and Alexey Guseynov and Alison Reid and Seth Odoom and Lucia Loher and Victor Cotruta and Madhavi Yenugula and Dominik Grewe and Anastasia Petrushkina and Tom Duerig and Antonio Sanchez and Steve Yadlowsky and Amy Shen and Amir Globerson and Lynette Webb and Sahil Dua and Dong Li and Surya Bhupatiraju and Dan Hurt and Haroon Qureshi and Ananth Agarwal and Tomer Shani and Matan Eyal and Anuj Khare and Shreyas Rammohan Belle and Lei Wang and Chetan Tekur and Mihir Sanjay Kale and Jinliang Wei and Ruoxin Sang and Brennan Saeta and Tyler Liechty and Yi Sun and Yao Zhao and Stephan Lee and Pandu Nayak and Doug Fritz and Manish Reddy Vuyyuru and John Aslanides and Nidhi Vyas and Martin Wicke and Xiao Ma and Evgenii Eltyshev and Nina Martin and Hardie Cate and James Manyika and Keyvan Amiri and Yelin Kim and Xi Xiong and Kai Kang and Florian Luisier and Nilesh Tripuraneni and David Madras and Mandy Guo and Austin Waters and Oliver Wang and Joshua Ainslie and Jason Baldridge and Han Zhang and Garima Pruthi and Jakob Bauer and Feng Yang and Riham Mansour and Jason Gelman and Yang Xu and George Polovets and Ji Liu and Honglong Cai and Warren Chen and XiangHai Sheng and Emily Xue and Sherjil Ozair and Christof Angermueller and Xiaowei Li and Anoop Sinha and Weiren Wang and Julia Wiesinger and Emmanouil Koukoumidis and Yuan Tian and Anand Iyer and Madhu Gurumurthy and Mark Goldenson and Parashar Shah and MK Blake and Hongkun Yu and Anthony Urbanowicz and Jennimaria Palomaki and Chrisantha Fernando and Ken Durden and Harsh Mehta and Nikola Momchev and Elahe Rahimtoroghi and Maria Georgaki and Amit Raul and Sebastian Ruder and Morgan Redshaw and Jinhyuk Lee and Denny Zhou and Komal Jalan and Dinghua Li and Blake Hechtman and Parker Schuh and Milad Nasr and Kieran Milan and Vladimir Mikulik and Juliana Franco and Tim Green and Nam Nguyen and Joe Kelley and Aroma Mahendru and Andrea Hu and Joshua Howland and Ben Vargas and Jeffrey Hui and Kshitij Bansal and Vikram Rao and Rakesh Ghiya and Emma Wang and Ke Ye and Jean Michel Sarr and Melanie Moranski Preston and Madeleine Elish and Steve Li and Aakash Kaku and Jigar Gupta and Ice Pasupat and Da-Cheng Juan and Milan Someswar and Tejvi M. and Xinyun Chen and Aida Amini and Alex Fabrikant and Eric Chu and Xuanyi Dong and Amruta Muthal and Senaka Buthpitiya and Sarthak Jauhari and Nan Hua and Urvashi Khandelwal and Ayal Hitron and Jie Ren and Larissa Rinaldi and Shahar Drath and Avigail Dabush and Nan-Jiang Jiang and Harshal Godhia and Uli Sachs and Anthony Chen and Yicheng Fan and Hagai Taitelbaum and Hila Noga and Zhuyun Dai and James Wang and Chen Liang and Jenny Hamer and Chun-Sung Ferng and Chenel Elkind and Aviel Atias and Paulina Lee and Vít Listík and Mathias Carlen and Jan van de Kerkhof and Marcin Pikus and Krunoslav Zaher and Paul Müller and Sasha Zykova and Richard Stefanec and Vitaly Gatsko and Christoph Hirnschall and Ashwin Sethi and Xingyu Federico Xu and Chetan Ahuja and Beth Tsai and Anca Stefanoiu and Bo Feng and Keshav Dhandhania and Manish Katyal and Akshay Gupta and Atharva Parulekar and Divya Pitta and Jing Zhao and Vivaan Bhatia and Yashodha Bhavnani and Omar Alhadlaq and Xiaolin Li and Peter Danenberg and Dennis Tu and Alex Pine and Vera Filippova and Abhipso Ghosh and Ben Limonchik and Bhargava Urala and Chaitanya Krishna Lanka and Derik Clive and Yi Sun and Edward Li and Hao Wu and Kevin Hongtongsak and Ianna Li and Kalind Thakkar and Kuanysh Omarov and Kushal Majmundar and Michael Alverson and Michael Kucharski and Mohak Patel and Mudit Jain and Maksim Zabelin and Paolo Pelagatti and Rohan Kohli and Saurabh Kumar and Joseph Kim and Swetha Sankar and Vineet Shah and Lakshmi Ramachandruni and Xiangkai Zeng and Ben Bariach and Laura Weidinger and Tu Vu and Alek Andreev and Antoine He and Kevin Hui and Sheleem Kashem and Amar Subramanya and Sissie Hsiao and Demis Hassabis and Koray Kavukcuoglu and Adam Sadovsky and Quoc Le and Trevor Strohman and Yonghui Wu and Slav Petrov and Jeffrey Dean and Oriol Vinyals},
      year={2025},
      eprint={2312.11805},
      archivePrefix={arXiv},
      primaryClass={cs.CL},
      url={https://arxiv.org/abs/2312.11805}, 
}

@misc{shi2022languagemodelsmultilingualchainofthought,
      title={Language Models are Multilingual Chain-of-Thought Reasoners}, 
      author={Freda Shi and Mirac Suzgun and Markus Freitag and Xuezhi Wang and Suraj Srivats and Soroush Vosoughi and Hyung Won Chung and Yi Tay and Sebastian Ruder and Denny Zhou and Dipanjan Das and Jason Wei},
      year={2022},
      eprint={2210.03057},
      archivePrefix={arXiv},
      primaryClass={cs.CL},
      url={https://arxiv.org/abs/2210.03057}, 
}

@inproceedings{koto-etal-2023-large,
    title = "Large Language Models Only Pass Primary School Exams in {I}ndonesia: A Comprehensive Test on {I}ndo{MMLU}",
    author = "Koto, Fajri  and
      Aisyah, Nurul  and
      Li, Haonan  and
      Baldwin, Timothy",
    editor = "Bouamor, Houda  and
      Pino, Juan  and
      Bali, Kalika",
    booktitle = "Proceedings of the 2023 Conference on Empirical Methods in Natural Language Processing",
    month = dec,
    year = "2023",
    address = "Singapore",
    publisher = "Association for Computational Linguistics",
    url = "https://aclanthology.org/2023.emnlp-main.760/",
    doi = "10.18653/v1/2023.emnlp-main.760",
    pages = "12359--12374"
}

@inproceedings{li-etal-2024-cmmlu,
    title = "{CMMLU}: Measuring massive multitask language understanding in {C}hinese",
    author = "Li, Haonan  and
      Zhang, Yixuan  and
      Koto, Fajri  and
      Yang, Yifei  and
      Zhao, Hai  and
      Gong, Yeyun  and
      Duan, Nan  and
      Baldwin, Timothy",
    editor = "Ku, Lun-Wei  and
      Martins, Andre  and
      Srikumar, Vivek",
    booktitle = "Findings of the Association for Computational Linguistics: ACL 2024",
    month = aug,
    year = "2024",
    address = "Bangkok, Thailand",
    publisher = "Association for Computational Linguistics",
    url = "https://aclanthology.org/2024.findings-acl.671/",
    doi = "10.18653/v1/2024.findings-acl.671",
    pages = "11260--11285"
}

@inproceedings{kwon2023efficient,
  title={Efficient Memory Management for Large Language Model Serving with PagedAttention},
  author={Woosuk Kwon and Zhuohan Li and Siyuan Zhuang and Ying Sheng and Lianmin Zheng and Cody Hao Yu and Joseph E. Gonzalez and Hao Zhang and Ion Stoica},
  booktitle={Proceedings of the ACM SIGOPS 29th Symposium on Operating Systems Principles},
  year={2023}
}

@misc{chiu2025culturalbenchrobustdiversechallenging,
      title={CulturalBench: A Robust, Diverse, and Challenging Cultural Benchmark by Human-AI CulturalTeaming}, 
      author={Yu Ying Chiu and Liwei Jiang and Bill Yuchen Lin and Chan Young Park and Shuyue Stella Li and Sahithya Ravi and Mehar Bhatia and Maria Antoniak and Yulia Tsvetkov and Vered Shwartz and Yejin Choi},
      year={2025},
      eprint={2410.02677},
      archivePrefix={arXiv},
      primaryClass={cs.CL},
      url={https://arxiv.org/abs/2410.02677}, 
}

@misc{hupkes2025multilokomultilinguallocalknowledge,
      title={MultiLoKo: a multilingual local knowledge benchmark for LLMs spanning 31 languages}, 
      author={Dieuwke Hupkes and Nikolay Bogoychev},
      year={2025},
      eprint={2504.10356},
      archivePrefix={arXiv},
      primaryClass={cs.CL},
      url={https://arxiv.org/abs/2504.10356}, 
}

@misc{eval-harness,
  author       = {Gao, Leo and Tow, Jonathan and Abbasi, Baber and Biderman, Stella and Black, Sid and DiPofi, Anthony and Foster, Charles and Golding, Laurence and Hsu, Jeffrey and Le Noac'h, Alain and Li, Haonan and McDonell, Kyle and Muennighoff, Niklas and Ociepa, Chris and Phang, Jason and Reynolds, Laria and Schoelkopf, Hailey and Skowron, Aviya and Sutawika, Lintang and Tang, Eric and Thite, Anish and Wang, Ben and Wang, Kevin and Zou, Andy},
  title        = {The Language Model Evaluation Harness},
  month        = 07,
  year         = 2024,
  publisher    = {Zenodo},
  version      = {v0.4.3},
  doi          = {10.5281/zenodo.12608602},
  url          = {https://zenodo.org/records/12608602}
}

@inproceedings{mohamed-etal-2024-culture,
    title = "No Culture Left Behind: {A}rt{EL}ingo-28, a Benchmark of {W}iki{A}rt with Captions in 28 Languages",
    author = "Mohamed, Youssef  and
      Li, Runjia  and
      Ahmad, Ibrahim Said  and
      Haydarov, Kilichbek  and
      Torr, Philip  and
      Church, Kenneth  and
      Elhoseiny, Mohamed",
    editor = "Al-Onaizan, Yaser  and
      Bansal, Mohit  and
      Chen, Yun-Nung",
    booktitle = "Proceedings of the 2024 Conference on Empirical Methods in Natural Language Processing",
    month = nov,
    year = "2024",
    address = "Miami, Florida, USA",
    publisher = "Association for Computational Linguistics",
    url = "https://aclanthology.org/2024.emnlp-main.1165/",
    doi = "10.18653/v1/2024.emnlp-main.1165",
    pages = "20939--20962"
}

@inproceedings{NEURIPS2024_1568882b,
 author = {Romero, David and Lyu, Chenyang and Wibowo, Haryo Akbarianto and Lynn, Teresa and Hamed, Injy and Kishore, Aditya Nanda and Mandal, Aishik and Dragonetti, Alina and Abzaliev, Artem and Tonja, Atnafu Lambebo and Balcha, Bontu Fufa and Whitehouse, Chenxi and Salamea, Christian and Velasco, Dan John and Adelani, David Ifeoluwa and Le Meur, David and Villa-Cueva, Emilio and Koto, Fajri and Farooqui, Fauzan and Belcavello, Frederico and Batnasan, Ganzorig and Vallejo, Gisela and Caulfield, Grainne and Ivetta, Guido and Song, Haiyue and Ademtew, Henok Biadglign and Maina, Hern\'{a}n and Lovenia, Holy and Azime, Israel Abebe and Cruz, Jan Christian Blaise and Gala, Jay and Geng, Jiahui and Ortiz-Barajas, Jesus-German and Baek, Jinheon and Dunstan, Jocelyn and Alemany, Laura Alonso and Nagasinghe, Kumaranage Ravindu Yasas and Benotti, Luciana and D\textquotesingle Haro, Luis Fernando and Viridiano, Marcelo and Estecha-Garitagoitia, Marcos and Cabrera, Maria Camila Buitrago and Rodr\'{\i}guez-Cantelar, Mario and Jouitteau, M\'{e}lanie and Mihaylov, Mihail and Etori, Naome and Imam, Mohamed Fazli Mohamed and Adilazuarda, Muhammad Farid and Gochoo, Munkhjargal and Otgonbold, Munkh-Erdene and Niyomugisha, Olivier and Silva, Paula M\'{o}nica and Chitale, Pranjal and Dabre, Raj and Chevi, Rendi and Zhang, Ruochen and Diandaru, Ryandito and Cahyawijaya, Samuel and G\'{o}ngora, Santiago and Jeong, Soyeong and Purkayastha, Sukannya and Kuribayashi, Tatsuki and Clifford, Teresa and Jayakumar, Thanmay and Torrent, Tiago Timponi and Ehsan, Toqeer and Araujo, Vladimir and Kementchedjhieva, Yova and Burzo, Zara and Lim, Zheng Wei and Yong, Zheng Xin and Ignat, Oana and Nwatu, Joan and Mihalcea, Rada and Solorio, Thamar and Aji, Alham Fikri},
 booktitle = {Advances in Neural Information Processing Systems},
 editor = {A. Globerson and L. Mackey and D. Belgrave and A. Fan and U. Paquet and J. Tomczak and C. Zhang},
 pages = {11479--11505},
 publisher = {Curran Associates, Inc.},
 title = {CVQA: Culturally-diverse Multilingual Visual Question Answering Benchmark},
 url = {https://proceedings.neurips.cc/paper_files/paper/2024/file/1568882ba1a50316e87852542523739c-Paper-Datasets_and_Benchmarks_Track.pdf},
 volume = {37},
 year = {2024}
}

@article{spearman1961proof,
  title={The proof and measurement of association between two things.},
  author={Spearman, Charles},
  year={1961},
  publisher={Appleton-Century-Crofts}
}

@article{pearson1895vii,
  title={VII. Note on regression and inheritance in the case of two parents},
  author={Pearson, Karl},
  journal={proceedings of the royal society of London},
  volume={58},
  number={347-352},
  pages={240--242},
  year={1895},
  publisher={The Royal Society London}
}

@ARTICLE{2020SciPy-NMeth,
  author  = {Virtanen, Pauli and Gommers, Ralf and Oliphant, Travis E. and
            Haberland, Matt and Reddy, Tyler and Cournapeau, David and
            Burovski, Evgeni and Peterson, Pearu and Weckesser, Warren and
            Bright, Jonathan and {van der Walt}, St{\'e}fan J. and
            Brett, Matthew and Wilson, Joshua and Millman, K. Jarrod and
            Mayorov, Nikolay and Nelson, Andrew R. J. and Jones, Eric and
            Kern, Robert and Larson, Eric and Carey, C J and
            Polat, {\.I}lhan and Feng, Yu and Moore, Eric W. and
            {VanderPlas}, Jake and Laxalde, Denis and Perktold, Josef and
            Cimrman, Robert and Henriksen, Ian and Quintero, E. A. and
            Harris, Charles R. and Archibald, Anne M. and
            Ribeiro, Ant{\^o}nio H. and Pedregosa, Fabian and
            {van Mulbregt}, Paul and {SciPy 1.0 Contributors}},
  title   = {{{SciPy} 1.0: Fundamental Algorithms for Scientific
            Computing in Python}},
  journal = {Nature Methods},
  year    = {2020},
  volume  = {17},
  pages   = {261--272},
  adsurl  = {https://rdcu.be/b08Wh},
  doi     = {10.1038/s41592-019-0686-2},
}

@misc{romanou2024includeevaluatingmultilinguallanguage,
      title={INCLUDE: Evaluating Multilingual Language Understanding with Regional Knowledge}, 
      author={Angelika Romanou and Negar Foroutan and Anna Sotnikova and Zeming Chen and Sree Harsha Nelaturu and Shivalika Singh and Rishabh Maheshwary and Micol Altomare and Mohamed A. Haggag and Snegha A and Alfonso Amayuelas and Azril Hafizi Amirudin and Viraat Aryabumi and Danylo Boiko and Michael Chang and Jenny Chim and Gal Cohen and Aditya Kumar Dalmia and Abraham Diress and Sharad Duwal and Daniil Dzenhaliou and Daniel Fernando Erazo Florez and Fabian Farestam and Joseph Marvin Imperial and Shayekh Bin Islam and Perttu Isotalo and Maral Jabbarishiviari and Börje F. Karlsson and Eldar Khalilov and Christopher Klamm and Fajri Koto and Dominik Krzemiński and Gabriel Adriano de Melo and Syrielle Montariol and Yiyang Nan and Joel Niklaus and Jekaterina Novikova and Johan Samir Obando Ceron and Debjit Paul and Esther Ploeger and Jebish Purbey and Swati Rajwal and Selvan Sunitha Ravi and Sara Rydell and Roshan Santhosh and Drishti Sharma and Marjana Prifti Skenduli and Arshia Soltani Moakhar and Bardia Soltani Moakhar and Ran Tamir and Ayush Kumar Tarun and Azmine Toushik Wasi and Thenuka Ovin Weerasinghe and Serhan Yilmaz and Mike Zhang and Imanol Schlag and Marzieh Fadaee and Sara Hooker and Antoine Bosselut},
      year={2024},
      eprint={2411.19799},
      archivePrefix={arXiv},
      primaryClass={cs.CL},
      url={https://arxiv.org/abs/2411.19799}, 
}

@misc{aakanksha2024multilingualalignmentprismaligning,
      title={The Multilingual Alignment Prism: Aligning Global and Local Preferences to Reduce Harm}, 
      author={Aakanksha and Arash Ahmadian and Beyza Ermis and Seraphina Goldfarb-Tarrant and Julia Kreutzer and Marzieh Fadaee and Sara Hooker},
      year={2024},
      eprint={2406.18682},
      archivePrefix={arXiv},
      primaryClass={cs.CL},
      url={https://arxiv.org/abs/2406.18682}, 
}

@misc{ramezani2023knowledgeculturalmoralnorms,
      title={Knowledge of cultural moral norms in large language models}, 
      author={Aida Ramezani and Yang Xu},
      year={2023},
      eprint={2306.01857},
      archivePrefix={arXiv},
      primaryClass={cs.CL},
      url={https://arxiv.org/abs/2306.01857}, 
}

@article{singh2024aya,
  title={Aya dataset: An open-access collection for multilingual instruction tuning},
  author={Singh, Shivalika and Vargus, Freddie and Dsouza, Daniel and Karlsson, B{\"o}rje F and Mahendiran, Abinaya and Ko, Wei-Yin and Shandilya, Herumb and Patel, Jay and Mataciunas, Deividas and OMahony, Laura and others},
  journal={arXiv preprint arXiv:2402.06619},
  year={2024}
}

@misc{workshop2023bloom176bparameteropenaccessmultilingual,
      title={BLOOM: A 176B-Parameter Open-Access Multilingual Language Model}, 
      author={BigScience Workshop and : and Teven Le Scao and Angela Fan and Christopher Akiki and Ellie Pavlick and Suzana Ilić and Daniel Hesslow and Roman Castagné and Alexandra Sasha Luccioni and François Yvon and Matthias Gallé and Jonathan Tow and Alexander M. Rush and Stella Biderman and Albert Webson and Pawan Sasanka Ammanamanchi and Thomas Wang and Benoît Sagot and Niklas Muennighoff and Albert Villanova del Moral and Olatunji Ruwase and Rachel Bawden and Stas Bekman and Angelina McMillan-Major and Iz Beltagy and Huu Nguyen and Lucile Saulnier and Samson Tan and Pedro Ortiz Suarez and Victor Sanh and Hugo Laurençon and Yacine Jernite and Julien Launay and Margaret Mitchell and Colin Raffel and Aaron Gokaslan and Adi Simhi and Aitor Soroa and Alham Fikri Aji and Amit Alfassy and Anna Rogers and Ariel Kreisberg Nitzav and Canwen Xu and Chenghao Mou and Chris Emezue and Christopher Klamm and Colin Leong and Daniel van Strien and David Ifeoluwa Adelani and Dragomir Radev and Eduardo González Ponferrada and Efrat Levkovizh and Ethan Kim and Eyal Bar Natan and Francesco De Toni and Gérard Dupont and Germán Kruszewski and Giada Pistilli and Hady Elsahar and Hamza Benyamina and Hieu Tran and Ian Yu and Idris Abdulmumin and Isaac Johnson and Itziar Gonzalez-Dios and Javier de la Rosa and Jenny Chim and Jesse Dodge and Jian Zhu and Jonathan Chang and Jörg Frohberg and Joseph Tobing and Joydeep Bhattacharjee and Khalid Almubarak and Kimbo Chen and Kyle Lo and Leandro Von Werra and Leon Weber and Long Phan and Loubna Ben allal and Ludovic Tanguy and Manan Dey and Manuel Romero Muñoz and Maraim Masoud and María Grandury and Mario Šaško and Max Huang and Maximin Coavoux and Mayank Singh and Mike Tian-Jian Jiang and Minh Chien Vu and Mohammad A. Jauhar and Mustafa Ghaleb and Nishant Subramani and Nora Kassner and Nurulaqilla Khamis and Olivier Nguyen and Omar Espejel and Ona de Gibert and Paulo Villegas and Peter Henderson and Pierre Colombo and Priscilla Amuok and Quentin Lhoest and Rheza Harliman and Rishi Bommasani and Roberto Luis López and Rui Ribeiro and Salomey Osei and Sampo Pyysalo and Sebastian Nagel and Shamik Bose and Shamsuddeen Hassan Muhammad and Shanya Sharma and Shayne Longpre and Somaieh Nikpoor and Stanislav Silberberg and Suhas Pai and Sydney Zink and Tiago Timponi Torrent and Timo Schick and Tristan Thrush and Valentin Danchev and Vassilina Nikoulina and Veronika Laippala and Violette Lepercq and Vrinda Prabhu and Zaid Alyafeai and Zeerak Talat and Arun Raja and Benjamin Heinzerling and Chenglei Si and Davut Emre Taşar and Elizabeth Salesky and Sabrina J. Mielke and Wilson Y. Lee and Abheesht Sharma and Andrea Santilli and Antoine Chaffin and Arnaud Stiegler and Debajyoti Datta and Eliza Szczechla and Gunjan Chhablani and Han Wang and Harshit Pandey and Hendrik Strobelt and Jason Alan Fries and Jos Rozen and Leo Gao and Lintang Sutawika and M Saiful Bari and Maged S. Al-shaibani and Matteo Manica and Nihal Nayak and Ryan Teehan and Samuel Albanie and Sheng Shen and Srulik Ben-David and Stephen H. Bach and Taewoon Kim and Tali Bers and Thibault Fevry and Trishala Neeraj and Urmish Thakker and Vikas Raunak and Xiangru Tang and Zheng-Xin Yong and Zhiqing Sun and Shaked Brody and Yallow Uri and Hadar Tojarieh and Adam Roberts and Hyung Won Chung and Jaesung Tae and Jason Phang and Ofir Press and Conglong Li and Deepak Narayanan and Hatim Bourfoune and Jared Casper and Jeff Rasley and Max Ryabinin and Mayank Mishra and Minjia Zhang and Mohammad Shoeybi and Myriam Peyrounette and Nicolas Patry and Nouamane Tazi and Omar Sanseviero and Patrick von Platen and Pierre Cornette and Pierre François Lavallée and Rémi Lacroix and Samyam Rajbhandari and Sanchit Gandhi and Shaden Smith and Stéphane Requena and Suraj Patil and Tim Dettmers and Ahmed Baruwa and Amanpreet Singh and Anastasia Cheveleva and Anne-Laure Ligozat and Arjun Subramonian and Aurélie Névéol and Charles Lovering and Dan Garrette and Deepak Tunuguntla and Ehud Reiter and Ekaterina Taktasheva and Ekaterina Voloshina and Eli Bogdanov and Genta Indra Winata and Hailey Schoelkopf and Jan-Christoph Kalo and Jekaterina Novikova and Jessica Zosa Forde and Jordan Clive and Jungo Kasai and Ken Kawamura and Liam Hazan and Marine Carpuat and Miruna Clinciu and Najoung Kim and Newton Cheng and Oleg Serikov and Omer Antverg and Oskar van der Wal and Rui Zhang and Ruochen Zhang and Sebastian Gehrmann and Shachar Mirkin and Shani Pais and Tatiana Shavrina and Thomas Scialom and Tian Yun and Tomasz Limisiewicz and Verena Rieser and Vitaly Protasov and Vladislav Mikhailov and Yada Pruksachatkun and Yonatan Belinkov and Zachary Bamberger and Zdeněk Kasner and Alice Rueda and Amanda Pestana and Amir Feizpour and Ammar Khan and Amy Faranak and Ana Santos and Anthony Hevia and Antigona Unldreaj and Arash Aghagol and Arezoo Abdollahi and Aycha Tammour and Azadeh HajiHosseini and Bahareh Behroozi and Benjamin Ajibade and Bharat Saxena and Carlos Muñoz Ferrandis and Daniel McDuff and Danish Contractor and David Lansky and Davis David and Douwe Kiela and Duong A. Nguyen and Edward Tan and Emi Baylor and Ezinwanne Ozoani and Fatima Mirza and Frankline Ononiwu and Habib Rezanejad and Hessie Jones and Indrani Bhattacharya and Irene Solaiman and Irina Sedenko and Isar Nejadgholi and Jesse Passmore and Josh Seltzer and Julio Bonis Sanz and Livia Dutra and Mairon Samagaio and Maraim Elbadri and Margot Mieskes and Marissa Gerchick and Martha Akinlolu and Michael McKenna and Mike Qiu and Muhammed Ghauri and Mykola Burynok and Nafis Abrar and Nazneen Rajani and Nour Elkott and Nour Fahmy and Olanrewaju Samuel and Ran An and Rasmus Kromann and Ryan Hao and Samira Alizadeh and Sarmad Shubber and Silas Wang and Sourav Roy and Sylvain Viguier and Thanh Le and Tobi Oyebade and Trieu Le and Yoyo Yang and Zach Nguyen and Abhinav Ramesh Kashyap and Alfredo Palasciano and Alison Callahan and Anima Shukla and Antonio Miranda-Escalada and Ayush Singh and Benjamin Beilharz and Bo Wang and Caio Brito and Chenxi Zhou and Chirag Jain and Chuxin Xu and Clémentine Fourrier and Daniel León Periñán and Daniel Molano and Dian Yu and Enrique Manjavacas and Fabio Barth and Florian Fuhrimann and Gabriel Altay and Giyaseddin Bayrak and Gully Burns and Helena U. Vrabec and Imane Bello and Ishani Dash and Jihyun Kang and John Giorgi and Jonas Golde and Jose David Posada and Karthik Rangasai Sivaraman and Lokesh Bulchandani and Lu Liu and Luisa Shinzato and Madeleine Hahn de Bykhovetz and Maiko Takeuchi and Marc Pàmies and Maria A Castillo and Marianna Nezhurina and Mario Sänger and Matthias Samwald and Michael Cullan and Michael Weinberg and Michiel De Wolf and Mina Mihaljcic and Minna Liu and Moritz Freidank and Myungsun Kang and Natasha Seelam and Nathan Dahlberg and Nicholas Michio Broad and Nikolaus Muellner and Pascale Fung and Patrick Haller and Ramya Chandrasekhar and Renata Eisenberg and Robert Martin and Rodrigo Canalli and Rosaline Su and Ruisi Su and Samuel Cahyawijaya and Samuele Garda and Shlok S Deshmukh and Shubhanshu Mishra and Sid Kiblawi and Simon Ott and Sinee Sang-aroonsiri and Srishti Kumar and Stefan Schweter and Sushil Bharati and Tanmay Laud and Théo Gigant and Tomoya Kainuma and Wojciech Kusa and Yanis Labrak and Yash Shailesh Bajaj and Yash Venkatraman and Yifan Xu and Yingxin Xu and Yu Xu and Zhe Tan and Zhongli Xie and Zifan Ye and Mathilde Bras and Younes Belkada and Thomas Wolf},
      year={2023},
      eprint={2211.05100},
      archivePrefix={arXiv},
      primaryClass={cs.CL},
      url={https://arxiv.org/abs/2211.05100}, 
}

@misc{xue2021mt5massivelymultilingualpretrained,
      title={mT5: A massively multilingual pre-trained text-to-text transformer}, 
      author={Linting Xue and Noah Constant and Adam Roberts and Mihir Kale and Rami Al-Rfou and Aditya Siddhant and Aditya Barua and Colin Raffel},
      year={2021},
      eprint={2010.11934},
      archivePrefix={arXiv},
      primaryClass={cs.CL},
      url={https://arxiv.org/abs/2010.11934}, 
}

@misc{ahuja2022staticmodelstestsets,
      title={Beyond Static Models and Test Sets: Benchmarking the Potential of Pre-trained Models Across Tasks and Languages}, 
      author={Kabir Ahuja and Sandipan Dandapat and Sunayana Sitaram and Monojit Choudhury},
      year={2022},
      eprint={2205.06356},
      archivePrefix={arXiv},
      primaryClass={cs.CL},
      url={https://arxiv.org/abs/2205.06356}, 
}

@misc{choenni2024evaluationpracticesmultilingualnlp,
      title={On the Evaluation Practices in Multilingual NLP: Can Machine Translation Offer an Alternative to Human Translations?}, 
      author={Rochelle Choenni and Sara Rajaee and Christof Monz and Ekaterina Shutova},
      year={2024},
      eprint={2406.14267},
      archivePrefix={arXiv},
      primaryClass={cs.CL},
      url={https://arxiv.org/abs/2406.14267}, 
}

@Article{info16090723,
AUTHOR = {Ataman, Duygu and Birch, Alexandra and Habash, Nizar and Federico, Marcello and Koehn, Philipp and Cho, Kyunghyun},
TITLE = {Machine Translation in the Era of Large Language Models:A Survey of Historical and Emerging Problems},
JOURNAL = {Information},
VOLUME = {16},
YEAR = {2025},
NUMBER = {9},
ARTICLE-NUMBER = {723},
URL = {https://www.mdpi.com/2078-2489/16/9/723},
ISSN = {2078-2489},
DOI = {10.3390/info16090723}
}

@article{liu2025culturally,
  title={Culturally aware and adapted nlp: A taxonomy and a survey of the state of the art},
  author={Liu, Chen Cecilia and Gurevych, Iryna and Korhonen, Anna},
  journal={Transactions of the Association for Computational Linguistics},
  volume={13},
  pages={652--689},
  year={2025},
  publisher={MIT Press 255 Main Street, 9th Floor, Cambridge, Massachusetts 02142, USA~…}
}

@misc{joshi2021statefatelinguisticdiversity,
      title={The State and Fate of Linguistic Diversity and Inclusion in the NLP World}, 
      author={Pratik Joshi and Sebastin Santy and Amar Budhiraja and Kalika Bali and Monojit Choudhury},
      year={2021},
      eprint={2004.09095},
      archivePrefix={arXiv},
      primaryClass={cs.CL},
      url={https://arxiv.org/abs/2004.09095}, 
}

@inproceedings{kojima-etal-2024-multilingual,
    title = "On the Multilingual Ability of Decoder-based Pre-trained Language Models: Finding and Controlling Language-Specific Neurons",
    author = "Kojima, Takeshi  and
      Okimura, Itsuki  and
      Iwasawa, Yusuke  and
      Yanaka, Hitomi  and
      Matsuo, Yutaka",
    editor = "Duh, Kevin  and
      Gomez, Helena  and
      Bethard, Steven",
    booktitle = "Proceedings of the 2024 Conference of the North American Chapter of the Association for Computational Linguistics: Human Language Technologies (Volume 1: Long Papers)",
    month = jun,
    year = "2024",
    address = "Mexico City, Mexico",
    publisher = "Association for Computational Linguistics",
    url = "https://aclanthology.org/2024.naacl-long.384/",
    doi = "10.18653/v1/2024.naacl-long.384",
    pages = "6919--6971",
}

@inproceedings{wendler-etal-2024-llamas,
    title = "Do Llamas Work in {E}nglish? On the Latent Language of Multilingual Transformers",
    author = "Wendler, Chris  and
      Veselovsky, Veniamin  and
      Monea, Giovanni  and
      West, Robert",
    editor = "Ku, Lun-Wei  and
      Martins, Andre  and
      Srikumar, Vivek",
    booktitle = "Proceedings of the 62nd Annual Meeting of the Association for Computational Linguistics (Volume 1: Long Papers)",
    month = aug,
    year = "2024",
    address = "Bangkok, Thailand",
    publisher = "Association for Computational Linguistics",
    url = "https://aclanthology.org/2024.acl-long.820/",
    doi = "10.18653/v1/2024.acl-long.820",
    pages = "15366--15394",
}

@inproceedings{tang-etal-2024-language,
    title = "Language-Specific Neurons: The Key to Multilingual Capabilities in Large Language Models",
    author = "Tang, Tianyi  and
      Luo, Wenyang  and
      Huang, Haoyang  and
      Zhang, Dongdong  and
      Wang, Xiaolei  and
      Zhao, Xin  and
      Wei, Furu  and
      Wen, Ji-Rong",
    editor = "Ku, Lun-Wei  and
      Martins, Andre  and
      Srikumar, Vivek",
    booktitle = "Proceedings of the 62nd Annual Meeting of the Association for Computational Linguistics (Volume 1: Long Papers)",
    month = aug,
    year = "2024",
    address = "Bangkok, Thailand",
    publisher = "Association for Computational Linguistics",
    url = "https://aclanthology.org/2024.acl-long.309/",
    doi = "10.18653/v1/2024.acl-long.309",
    pages = "5701--5715",
}

@inproceedings{alabi-etal-2024-hidden,
    title = "The Hidden Space of Transformer Language Adapters",
    author = "Alabi, Jesujoba  and
      Mosbach, Marius  and
      Eyal, Matan  and
      Klakow, Dietrich  and
      Geva, Mor",
    editor = "Ku, Lun-Wei  and
      Martins, Andre  and
      Srikumar, Vivek",
    booktitle = "Proceedings of the 62nd Annual Meeting of the Association for Computational Linguistics (Volume 1: Long Papers)",
    month = aug,
    year = "2024",
    address = "Bangkok, Thailand",
    publisher = "Association for Computational Linguistics",
    url = "https://aclanthology.org/2024.acl-long.356/",
    doi = "10.18653/v1/2024.acl-long.356",
    pages = "6588--6607",
}

@inproceedings{
bandarkar2025layer,
title={Layer Swapping for Zero-Shot Cross-Lingual Transfer in Large Language Models},
author={Lucas Bandarkar and Benjamin Muller and Pritish Yuvraj and Rui Hou and Nayan Singhal and Hongjiang Lv and Bing Liu},
booktitle={The Thirteenth International Conference on Learning Representations},
year={2025},
url={https://openreview.net/forum?id=vQhn4wrQ6j}
}

@article{lcm2024,
  author = {{LCM team} and Lo\"{i}c Barrault and Paul-Ambroise Duquenne and Maha Elbayad and Artyom Kozhevnikov and Belen Alastruey and Pierre Andrews and Mariano Coria and Guillaume Couairon and Marta R. Costa-juss\`{a} and David Dale and Hady Elsahar and Kevin Heffernan and Jo\~{a}o Maria Janeiro and Tuan Tran and Christophe Ropers and Eduardo Sánchez and Robin San Roman and Alexandre Mourachko and Safiyyah Saleem and Holger Schwenk},
  title = {{Large Concept Models}: Language Modeling in a Sentence Representation Space},
  publisher = {arXiv},
  year = {2024},
  url = {https://arxiv.org/abs/2412.08821},
}

@inproceedings{
    wu2025the,
    title={The Semantic Hub Hypothesis: Language Models Share Semantic Representations Across Languages and Modalities},
    author={Zhaofeng Wu and Xinyan Velocity Yu and Dani Yogatama and Jiasen Lu and Yoon Kim},
    booktitle={The Thirteenth International Conference on Learning Representations},
    year={2025},
    url={https://openreview.net/forum?id=FrFQpAgnGE}
}

@article{romanou2024include,
  title={INCLUDE: Evaluating Multilingual Language Understanding with Regional Knowledge},
  author={Romanou, Angelika and Foroutan, Negar and Sotnikova, Anna and Chen, Zeming and Nelaturu, Sree Harsha and Singh, Shivalika and Maheshwary, Rishabh and Altomare, Micol and Haggag, Mohamed A and Amayuelas, Alfonso and others},
  journal={arXiv preprint arXiv:2411.19799},
  year={2024}
}

@misc{issaka2025africanlanguageslabcollaborative,
      title={The African Languages Lab: A Collaborative Approach to Advancing Low-Resource African NLP}, 
      author={Sheriff Issaka and Keyi Wang and Yinka Ajibola and Oluwatumininu Samuel-Ipaye and Zhaoyi Zhang and Nicte Aguillon Jimenez and Evans Kofi Agyei and Abraham Lin and Rohan Ramachandran and Sadick Abdul Mumin and Faith Nchifor and Mohammed Shuraim and Lieqi Liu and Erick Rosas Gonzalez and Sylvester Kpei and Jemimah Osei and Carlene Ajeneza and Persis Boateng and Prisca Adwoa Dufie Yeboah and Saadia Gabriel},
      year={2025},
      eprint={2510.05644},
      archivePrefix={arXiv},
      primaryClass={cs.CL},
      url={https://arxiv.org/abs/2510.05644}, 
}

@misc{issaka2024ghanaiannlplandscapelook,
      title={The Ghanaian NLP Landscape: A First Look}, 
      author={Sheriff Issaka and Zhaoyi Zhang and Mihir Heda and Keyi Wang and Yinka Ajibola and Ryan DeMar and Xuefeng Du},
      year={2024},
      eprint={2405.06818},
      archivePrefix={arXiv},
      primaryClass={cs.CL},
      url={https://arxiv.org/abs/2405.06818}, 
}

\appendix
\newpage
\clearpage

\startcontents[sections]
\printcontents[sections]{l}{1}{\setcounter{tocdepth}{3}}

\newpage
\clearpage

\section{Discussion}
Our results support a simple but consequential claim: across a broad set of multilingual benchmarks and model families, translation quality provides a strong first-pass signal of downstream multilingual performance (Table~\ref{tab:transposed_correlations}; Figure~\ref{fig:scaling}).
This section attempts to explore \emph{why} that relationship emerges, \emph{when} it weakens, and \emph{how} to use translation-based proxies responsibly in multilingual evaluation.

\subsection{Translation as a capability-access signal}
Translation is a demanding NLU/NLG task because it requires mapping between non-isomorphic linguistic systems while preserving meaning across semantics and discourse.
From the perspective of multilingual model internals, recent work suggests a functional separation between language-specific processing and language-agnostic computation: multilingual information is largely mediated by the lower and upper transformer layers, while intermediate layers behave more language-universal \citep{kojima-etal-2024-multilingual, wendler-etal-2024-llamas, tang-etal-2024-language, alabi-etal-2024-hidden, wu2025the}.
This structure is consistent with a modular view in which the model must (i) encode an input language into a shared representation space, (ii) apply core reasoning/knowledge within that space, and (iii) decode into the target language \citep{bandarkar2025layer, lcm2024}.

Under this view, MT benchmarks stress-test the linguistic interfaces (encoding/decoding) that gate access to the model’s internal capabilities.
If a model translates a language well, it is evidence that the model can reliably move information into and out of the shared computation space for that language, making downstream success on many language-understanding tasks more likely.
This interpretation aligns with our empirical findings: stronger translation quality tends to coincide with stronger benchmark performance across languages and models (Figure~\ref{fig:scaling}).

\subsection{Why translation quality correlates with downstream performance}
Two patterns in our results are particularly informative.
First, the strongest translation-benchmark alignment occurs on benchmarks that are primarily gated by broad semantic understanding and faithful mapping from text to decisions.
For example, median correlations are highest for \textsc{AfriMMLU} (median \(r=0.92\)), \textsc{HellaSwag} (median \(r=0.87\)), and \textsc{Belebele} (median \(r=0.83\); Table~\ref{tab:transposed_correlations}).
Second, the proxy remains strong across model sizes (1B-72B), suggesting it reflects a stable capability relationship rather than a quirk of one model family (Figure~\ref{fig:scaling}).

At the language level, our findings further illustrate that translation quality can track benchmark performance with tight rank and linear agreement across many languages (Figures~\ref{fig:belebele_vs_ssa-comet} and~\ref{fig:hellaswag_vs_xcomet}).
Together, these observations support the hypothesis that a large component of multilingual benchmark performance is mediated by the same cross-lingual access mechanisms measured by translation quality.

\subsection{When translation is an incomplete proxy}
While translation quality is informative overall, our results also show clear boundary conditions.
Benchmarks that emphasize more specialized competencies or evaluation formats exhibit weaker alignment with translation scores.
In particular, \textsc{INCLUDE} has the lowest median correlation (median \(r=0.62\)), and \textsc{MGSM} shows only moderate alignment (median \(r=0.72\); Table~\ref{tab:transposed_correlations}).
These cases are consistent with the idea that downstream performance can be limited by factors other than linguistic access (e.g., specialized reasoning demands, task-specific constraints, or domain mismatch) so translation quality alone cannot serve as a complete substitute for task-centric evaluation.

We also observe that the relationship is not uniform across metrics within a task.
For example, \textsc{MLQA} exhibits substantial metric sensitivity (Table~\ref{tab:transposed_correlations}), indicating that even when translation is a useful proxy, metric choice can materially change conclusions.
This reinforces a practical takeaway: translation-first screening should be paired with targeted downstream evaluation wherever the task is high-stakes, correlations are weaker, or metric behavior is unstable.

\subsection{Metric choice: what to use for translation-first screening}
Our results consistently rank neural MT metrics as the most correlated overall, with median correlations of \(r=0.91\) for \textsc{xCOMET}, \(r=0.89\) for \textsc{MetricX}, and \(r=0.87\) for \textsc{SSA-COMET} (Table~\ref{tab:transposed_correlations}).
At the same time, lexical metrics remain surprisingly informative (e.g., \textsc{BLEU} median \(r=0.79\)), which provides a useful sanity check that the signal is not solely an artifact of learned evaluators.

In practice, these findings suggest a simple evaluation recipe.
\textsc{xCOMET} is a strong default for proxy-based screening due to its top aggregate performance, while \textsc{MetricX} can be particularly informative on benchmarks where lexical/semantic alignment behaves differently (e.g., \textsc{MLQA}).
Where compute or tooling is constrained, \textsc{BLEU} can serve as a coarse baseline, but it should not be treated as interchangeable with neural metrics for all tasks.

\subsection{Task-centric analysis, language overlap, and resource stratification}
A natural question is whether correlations are primarily driven by task demands or language-level factors (e.g., resource availability).
In principle, language-centric analyses (computing correlations per language across tasks) could help disentangle these effects.
In practice, the multilingual benchmark ecosystem provides limited language overlap across tasks, making robust per-language correlation estimates infeasible in many settings; our own analysis therefore remains task-centric, using all available languages per benchmark to recover stable estimates.

As a partial alternative, we report coarse stratification by resource class for Phi-4 (Table~\ref{tab:correlation_segments}).
While this view can hint at differences across resource tiers, it necessarily reduces sample sizes and aggregates heterogeneous languages, so it should be interpreted cautiously and not as definitive evidence of resource-specific effects.

\subsection{Data contamination and robustness of correlation findings}
As with most LLM evaluations, we cannot fully rule out data contamination because model training corpora are not disclosed in detail.
Contamination could inflate absolute benchmark scores, and in principle could also affect correlations if exposure is systematically uneven across languages or datasets.
However, the breadth of our analysis (multiple translation datasets, multiple metrics, diverse benchmarks, and many model families) makes it less plausible that a single contamination source is the sole driver of the consistent alignment we observe.
We therefore treat contamination as a remaining confound rather than a complete explanation, and view controlled causal studies as an important direction for future work.

\subsection{Implications for efficient multilingual evaluation}
Taken together, our findings motivate a staged evaluation workflow.
Translation-based proxy metrics can provide an inexpensive first-pass assessment across many languages, enabling researchers to prioritize which model-language pairs warrant deeper benchmarking.
Downstream task suites remain necessary when evaluating competencies that translation does not reliably capture (as reflected by weaker correlations on benchmarks such as \textsc{INCLUDE} and \textsc{MGSM}) and whenever deployment decisions require task-specific validation.

\section{Ethical Implications}
Using translation quality as a proxy for multilingual evaluation can materially change who is able to measure model performance and which languages receive attention.
This section summarizes the primary benefits, risks, and responsible-use considerations implied by our findings.

\subsection{Democratizing multilingual assessment}
A central motivation for translation-first evaluation is to reduce the cost and expertise barrier of multilingual benchmarking.
Because high-quality parallel corpora and MT metrics often cover far more languages than task suites, translation-based proxies can enable low-cost, reproducible screening for languages that otherwise remain unevaluated.
In this sense, proxy evaluation can broaden participation: communities and researchers who lack the resources to build and maintain task benchmarks can still obtain informative, comparable signals of model behavior.

\subsection{Risk of translation-centric bias}
Proxy evaluation also risks over-weighting a single notion of ``language competence''.
Translation benchmarks tend to emphasize written, standardized language and meaning preservation, and our results show that translation quality is not equally correlated for all downstream competencies (e.g., weaker correlations on \textsc{INCLUDE} and \textsc{MGSM}; Table~\ref{tab:transposed_correlations}).
Over-reliance on translation scores could therefore mischaracterize readiness for culturally grounded, domain-specific, or reasoning-intensive applications, reinforcing a translation-centric bias in what the field chooses to measure and improve.

\subsection{Impacts on low-resource language communities}
For low-resource languages, translation-first screening can be empowering by providing a first visible signal of model support.
However, it also carries a failure mode: declaring a model ``good enough'' based only on proxy metrics can prematurely reduce investment in richer, community-defined evaluations, or conceal harms that do not manifest in translation quality.
Responsible use therefore requires aligning evaluation criteria with community priorities and treating translation as an entry point rather than a final verdict.

\subsection{Transparency and accountability}
Because proxy scores can be easy to compute and tempting to over-interpret, transparency is essential.
Researchers and practitioners should clearly state when translation metrics are used as substitutes for task benchmarks, report the translation datasets/metrics used, and communicate known boundary conditions.
For high-stakes deployment (e.g., education, healthcare, legal settings), translation-first evaluation should be supplemented with task-specific validation representative of the intended use case.

\subsection{Environmental considerations}
A practical benefit of translation-first evaluation is efficiency: computing translation metrics can be substantially cheaper than running large benchmark suites across many languages.
This can reduce computational cost and associated environmental impact.
Nevertheless, efficiency should not be used to justify under-evaluation in safety- or fairness-critical contexts; reduced cost is most valuable when it enables broader measurement while preserving targeted follow-up where necessary.

\subsection{Recommendations for responsible use}
Based on the empirical behavior observed in this work, we recommend the following principles:
(i) use translation metrics for rapid screening and comparative analysis, not as the sole basis for deployment decisions;
(ii) pair proxy evaluation with targeted benchmarks for tasks where correlations are weaker or unstable (e.g., \textsc{INCLUDE}, \textsc{MGSM});
and (iii) report proxy methodology and limitations clearly to avoid overstating multilingual competence.

\section{Additional Correlation Coefficient Table on WMT24++ and NTREX}
\begin{table*}[t]
\centering
\small
\setlength{\tabcolsep}{6pt}
\renewcommand{\arraystretch}{1}

\begin{adjustbox}{max width=\textwidth}
\begin{tabular}{|l|l|c|c|c|c|c|c|c|c|c|c|}
    \hline
    \textbf{Metric} & \textbf{Dataset} &
    \textbf{AfriMMLU} & \textbf{AfriXNLI} & \textbf{Belebele} & \textbf{Global MMLU} &
    \textbf{HellaSwag} & \textbf{INCLUDE} & \textbf{MGSM} & \textbf{MLQA} & \textbf{TruthfulQA} & \textbf{Median}\\
    \hline
    \multirow{3}{*}{\textbf{BLEU}} & FLORES & 0.95 (17) & 0.68 (17) & 0.89 (115) & 0.78 (14) & 0.88 (30) & 0.67 (44) & 0.68 (10) & 0.79 (6) & 0.79 (31) & \multirow{3}{*}{0.79} \\
    & WMT24++ & 1.00 (3) & 0.96 (3) & 0.69 (51) & 0.78 (13) & 0.87 (27) & 0.58 (32) & 0.80 (10) & 0.79 (6) & 0.70 (28) & \\
    & NTREX & 0.95 (13) & 0.72 (13) & 0.86 (85) & 0.83 (14) & 0.90 (29) & 0.67 (39) & 0.74 (10) & 0.76 (6) & 0.76 (30)  & \\
    \hline
    
    \multirow{3}{*}{\textbf{ChrF++}} & FLORES & 0.88 (17) & 0.64 (17) & 0.85 (115) & 0.63 (14) & 0.77 (30) & 0.56 (44) & 0.63 (10) & 0.51 (6) & 0.62 (31)  & \multirow{3}{*}{0.63}\\
    & WMT24++ & 0.99 (3) & 0.99 (3) & 0.44 (51) & 0.48 (13) & 0.71 (27) & 0.34 (32) & 0.66 (10) & 0.73 (6) & 0.55 (28) &\\
    & NTREX & 0.82 (13) & 0.57 (13) & 0.77 (85) & 0.61 (14) & 0.77 (29) & 0.49 (39) & 0.59 (10) & 0.51 (6) & 0.57 (30) & \\
    \hline
    
    \multirow{3}{*}{\textbf{METEOR}} & FLORES & 0.92 (17) & 0.65 (17) & 0.77 (115) & 0.41 (14) & 0.89 (30) & 0.53 (44) & 0.45 (10) & 0.26 (6) & 0.67 (31) & \multirow{3}{*}{0.62} \\
    & WMT24++ & 0.99 (3) & 1.00 (3) & 0.47 (51) & 0.44 (13) & 0.86 (27) & 0.37 (32) & 0.56 (10) & 0.56 (6) & 0.62 (28) & \\
    & NTREX & 0.88 (13) & 0.64 (13) & 0.72 (85) & 0.43 (14) & 0.90 (29) & 0.51 (39) & 0.47 (10) & 0.36 (6) & 0.65 (30) & \\
    \hline
    \multirow{3}{*}{\textbf{ROUGE-L}} & FLORES & 0.91 (17) & 0.71 (17) & 0.64 (115) & 0.47 (14) & 0.77 (30) & 0.49 (44) & 0.78 (10) & 0.63 (6) & 0.60 (31) & \multirow{3}{*}{0.66} \\
    & WMT24++ & 0.98 (3) & 1.00 (3) & 0.56 (51) & 0.66 (13) & 0.77 (27) & 0.46 (32) & 0.81 (10) & 0.64 (6) & 0.58 (28) & \\
    & NTREX & 0.91 (13) & 0.75 (13) & 0.68 (85) & 0.57 (14) & 0.80 (29) & 0.50 (39) & 0.80 (10) & 0.63 (6) & 0.61 (30)  & \\
    \hline
    
    \multirow{3}{*}{\textbf{MetricX}} & FLORES & 0.90 (17) & 0.74 (17) & 0.91 (115) & 0.94 (14) & 0.86 (30) & 0.69 (44) & 0.64 (10) & 0.95 (6) & 0.91 (31)  & \multirow{3}{*}{0.89}\\
    & WMT24++ & 0.97 (3) & 1.00 (3) & 0.71 (51) & 0.89 (13) & 0.83 (27) & 0.68 (32) & 0.64 (10) & 0.93 (6) & 0.81 (28) & \\
    & NTREX & 0.91 (13) & 0.79 (13) & 0.92 (85) & 0.94 (14) & 0.87 (29) & 0.71 (39) & 0.65 (10) & 0.98 (6) & 0.91 (30) & \\
    \hline
    
    \multirow{3}{*}{\textbf{SSA-COMET}} & FLORES & 0.81 (17) & 0.42 (17) & 0.94 (115) & 0.93 (14) & 0.92 (30) & 0.66 (44) & 0.72 (10) & 0.84 (6) & 0.92 (31) & \multirow{3}{*}{0.87}\\
    & WMT24++ & 0.91 (3) & 0.98 (3) & 0.85 (51) & 0.87 (13) & 0.89 (27) & 0.62 (32) & 0.73 (10) & 0.87 (6) & 0.85 (28) & \\
    & NTREX & 0.79 (13) & 0.37 (13) & 0.93 (85) & 0.93 (14) & 0.92 (29) & 0.66 (39) & 0.72 (10) & 0.87 (6) & 0.90 (30)  & \\
    \hline
    
    \multirow{3}{*}{\textbf{xCOMET}} & FLORES & 0.95 (17) & 0.68 (17) & 0.93 (115) & 0.96 (14) & 0.97 (30) & 0.77 (44) & 0.72 (10) & 0.76 (6) & 0.90 (31) & \multirow{3}{*}{0.91} \\
    & WMT24++ & 1.00 (3) & 0.95 (3) & 0.83 (51) & 0.92 (13) & 0.97 (27) & 0.73 (32) & 0.75 (10) & 0.91 (6) & 0.90 (28)  & \\
    & NTREX & 0.95 (13) & 0.68 (13) & 0.91 (85) & 0.95 (14) & 0.98 (29) & 0.78 (39) & 0.74 (10) & 0.81 (6) & 0.91 (30) &  \\
    \hline

    \multirow{1}{*}{\textbf{Statistics}} & Median & 0.92 & 0.72& 0.83 & 0.78 & 0.87 & 0.62 & 0.72 & 0.76 & 0.76 & --- \\ \hline
\end{tabular}
\end{adjustbox}

\caption{Pearson correlations between benchmark tasks and MT metrics grouped by dataset. Numbers inside parentheses (n) indicate the number of data points used in the correlation calculation.}
\label{tab:transposed_correlations}
\end{table*}

\begin{table*}[t]
    \centering
    \small 
    \setlength{\tabcolsep}{6pt} 
    
    \begin{adjustbox}{max width=\textwidth}
        \begin{tabular}{l|ccccccccc}
        \hline
            \textbf{Metric} & \textbf{AfriMMLU} & \textbf{AfriXNLI} & \textbf{Belebele} & \textbf{Global MMLU} & \textbf{HellaSwag} & \textbf{Include} & \textbf{MGSM} & \textbf{MLQA} & \textbf{TruthfulQA} \\ \hline
            High & --- & --- & 0.35 & 0.7 & 0.78 & 0.42 & 0.5 & 0.76 & 0.44 \\ 
            Medium & 0.815 & 0.44 & 0.88 & 0.98 & 0.91 & 0.65 & -0.3 & --- & 0.88 \\ 
            Low & -0.075 & -0.15 & 0.86 & --- & 0.86 & 0.915 & --- & --- & 0.46 \\ 
            All & 0.92 & 0.72 & 0.83 & 0.78 & 0.87 & 0.62 & 0.72 & 0.76 & 0.76 \\ \hline
        \end{tabular}
    \end{adjustbox}
    
    \caption{Medians for Phi-4. Grouped by language resources.}
    \label{tab:correlation_segments}
\end{table*}

Here we provide correlation matrices in the form of heatmaps of Phi-4 model on two of the three datasets used in this study, namely WMT24++ and NTREX. They contain a dash iff \( n \le 3\). We also include the correlation matrices calculated based on Spearman's rank on FLORES-200 dataset. 

\begin{figure}[H]
  \centering
  \begin{minipage}[t]{0.49\textwidth}
    \centering
    \includegraphics[width=\linewidth]{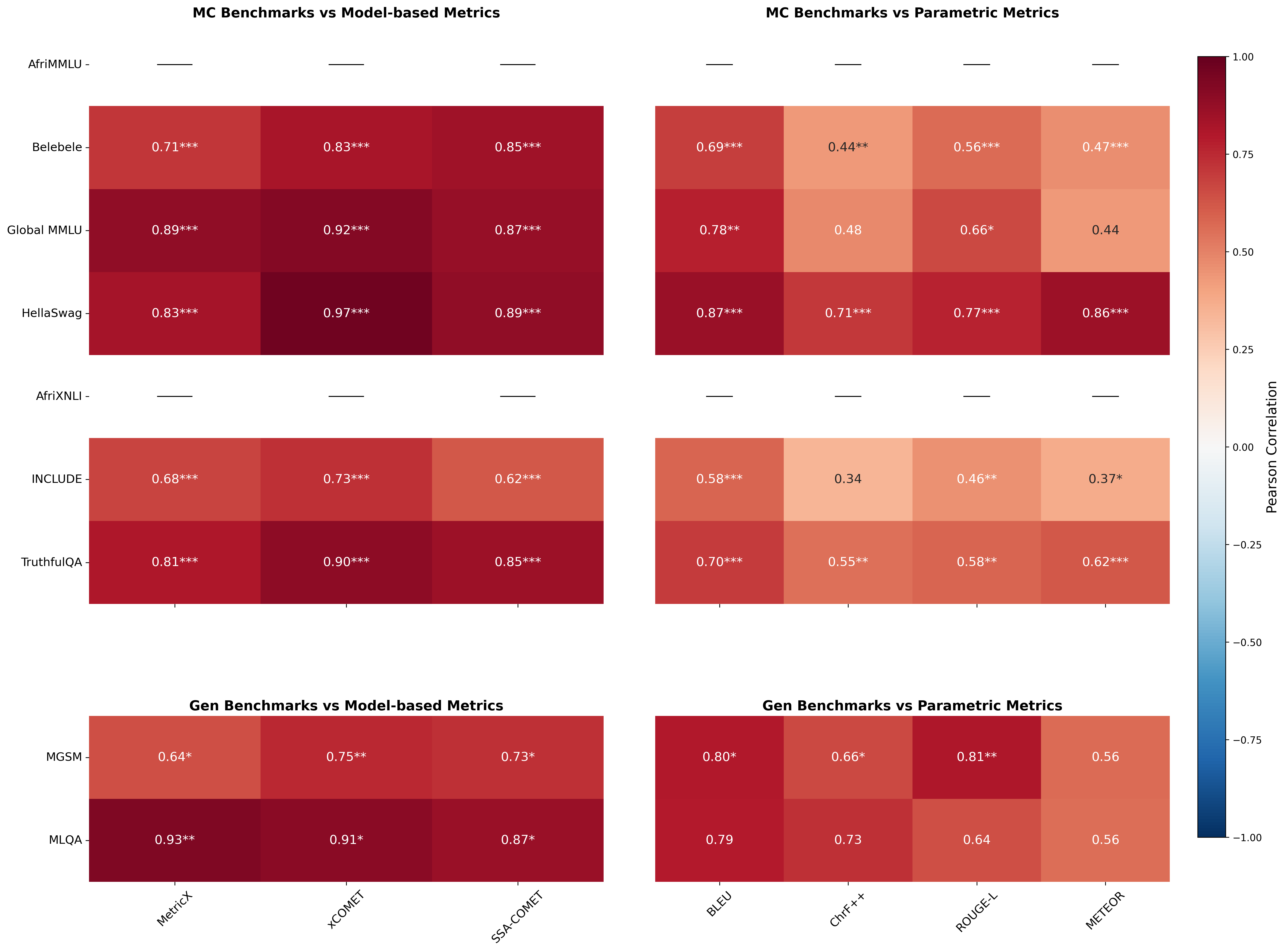}
    \captionof{figure}{WMT Correlation}\label{fig:left}
  \end{minipage}\hfill
  \begin{minipage}[t]{0.49\textwidth}
    \centering
    \includegraphics[width=\linewidth]{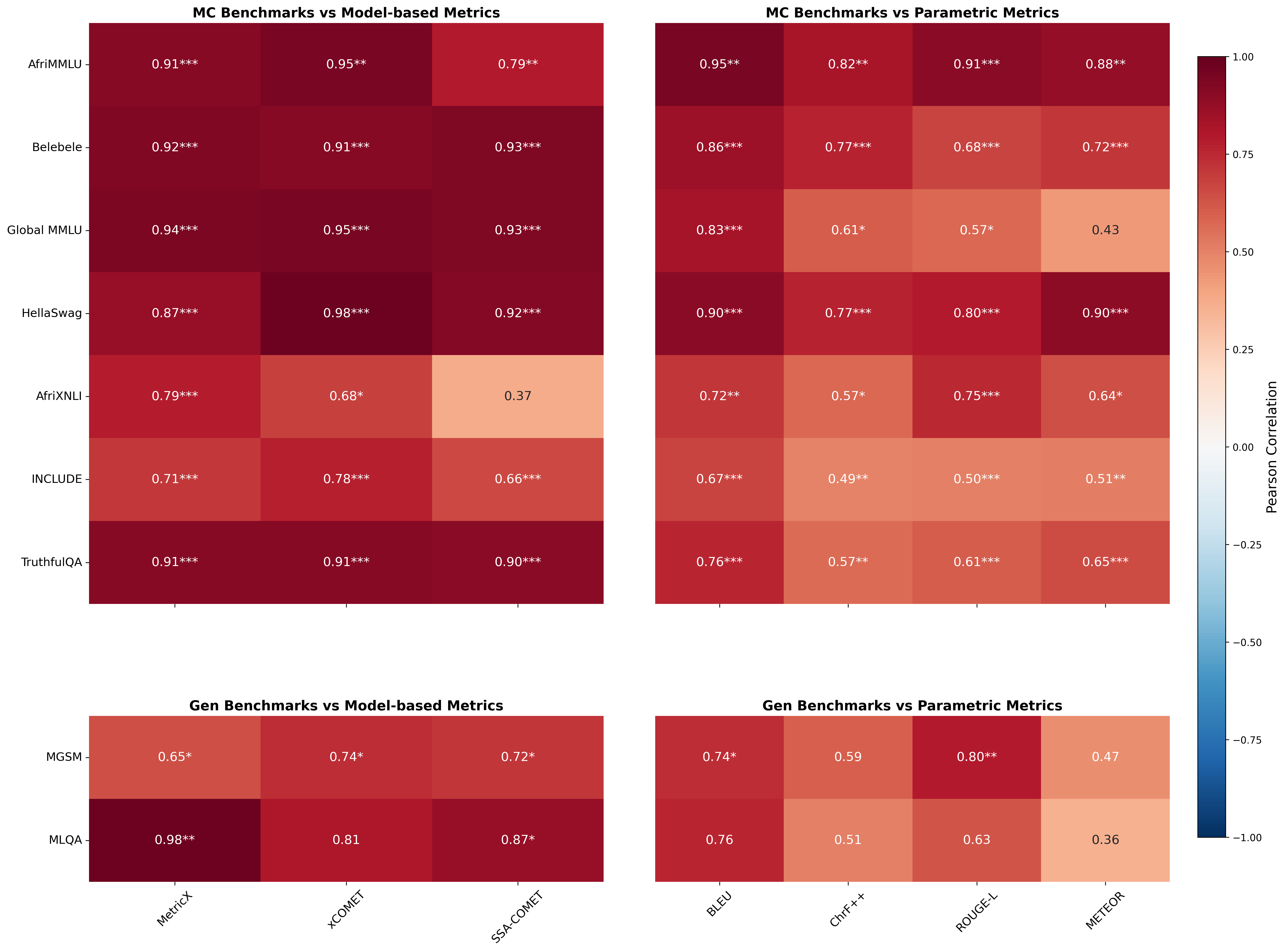}
    \captionof{figure}{NTREX Correlation}\label{fig:right}
  \end{minipage}
\end{figure}

\begin{figure}[H]
    \centering
    \includegraphics[width=\linewidth]{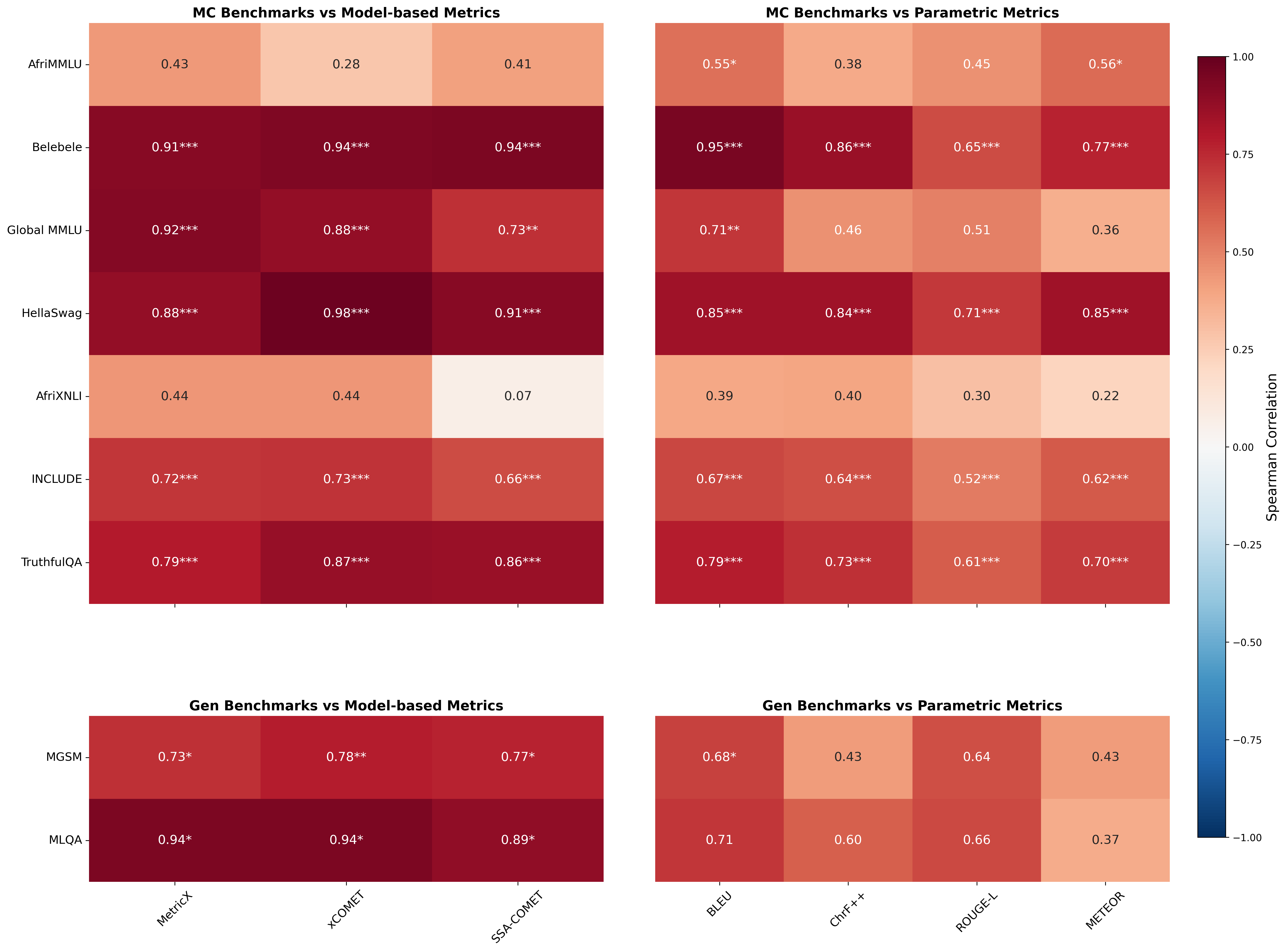}
    \caption{\textbf{Spearman} correlation matrix for Phi-4 model performance across evaluation metrics. Color intensity indicates the strength of correlation, with darker red representing stronger positive correlations. }
    \label{fig:placeholder}
\end{figure}

\section{Models Investigated}
\begin{table}[H]
\centering
\begin{tabular}{|l|l|}
\hline
\textbf{Model family} & \textbf{Model} \\
\hline
\multirow{6}{*}{Qwen}
  & Qwen3-4B \\
  & Qwen3-8B \\
  & Qwen3-14B \\
  & Qwen3-32B \\
  & Qwen3-30B-A3B(MOE) \\
  & Qwen2.5-72B-Instruct \\
\hline
\multirow{4}{*}{Gemma}
  & Gemma-3-1B-it \\
  & Gemma-3-4B-it \\
  & Gemma-3-12B-it \\
  & Gemma-3-27B-it \\
\hline
\multirow{1}{*}{Llama}
  & Llama-3.3-70B-Instruct \\
\hline
\multirow{2}{*}{DeepSeek}
  & DeepSeek-R1-Distill-Qwen-32B \\
  & DeepSeek-R1-Distill-Llama-70B \\
\hline
\multirow{1}{*}{Phi}
  & Phi-4 (14B) \\
\hline
\end{tabular}
\caption{Model families and corresponding models.}
\label{tab:model-families}
\end{table}

\section{Correlation Analysis Across All Models}
\label{app:additional_models}
\subsection{Llama Family}
\begin{figure}[H]
  \centering
  \begin{subfigure}[t]{\linewidth}
    \centering
    \includegraphics[width=\linewidth]{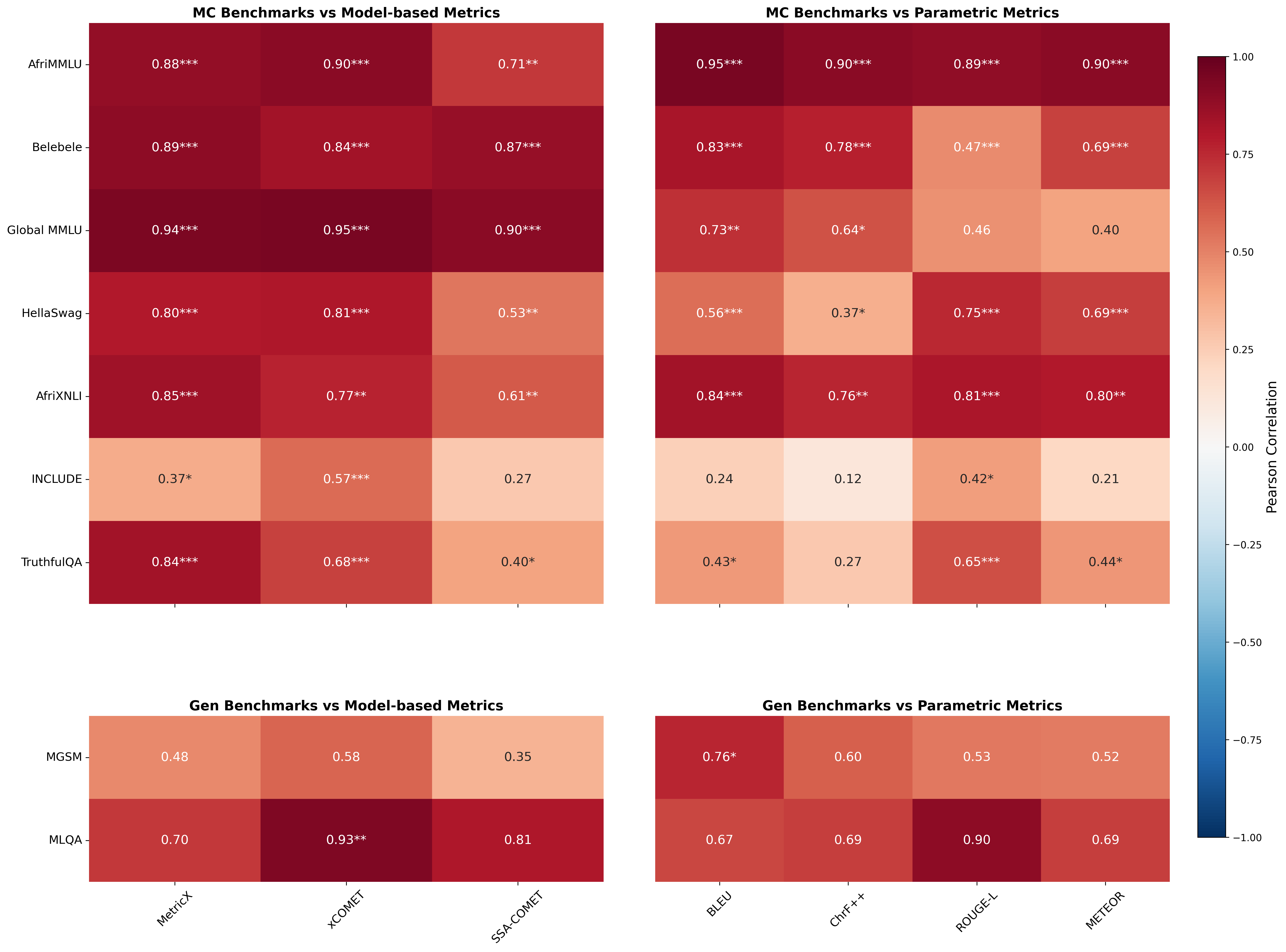}
    \caption{Llama3-70B-Instruct}
    \label{fig:llama-70b}
  \end{subfigure}

  \caption{Pearson correlations between MT metrics and benchmark tasks.
  Stars indicate significance: * $p<0.05$, ** $p<0.01$, *** $p<0.001$.}
  \label{fig:llama-family}
\end{figure}

The correlation patterns for Llama-70B are consistent with those of the Phi-4 model, indicating that the observed relationships are not unique to a single model family.

\noindent \textbf{Similar Trends}: Benchmarks like AfriMMLU and AfriXNLI show strong positive correlations with nearly all MT metrics. For instance, the correlation between AfriXNLI and MetricX is high (r=0.85, p < 0.001) closely mirroring the Qwen-72B model (r=0.74, p < 0.01). As with Phi-4, MGSM and INCLUDE continues to show the weakest correlations, most of which are not statistically significant.

\noindent \textbf{Differences}: The main difference is seen in the MLQA benchmark. For Llama-70B we see strong correlations Rouge-L (r=0.9), and like Phi-4, with XComet (r=0.93, p < 0.01).

\subsection{Deepseek Family}
\begin{figure}[H]
  \centering
  \begin{subfigure}[t]{\linewidth}
    \centering
    \includegraphics[width=\linewidth]{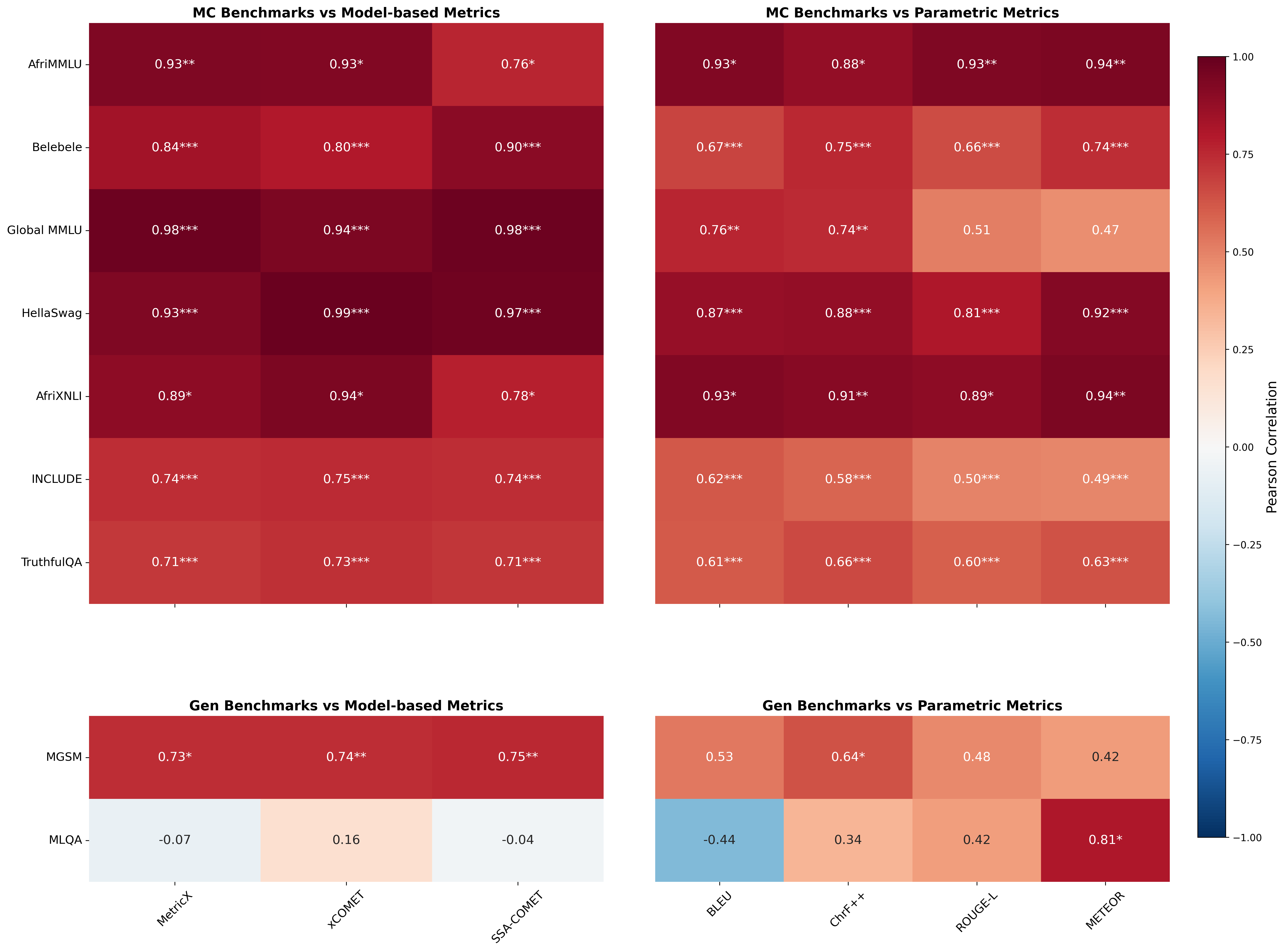}
    \caption{DeepSeek-R1-Distill-Qwen-32B}
    \label{fig:ds-32b}
  \end{subfigure}

    \vspace{0.6em}
  
  \begin{subfigure}[t]{\linewidth}
    \centering
    \includegraphics[width=\linewidth]{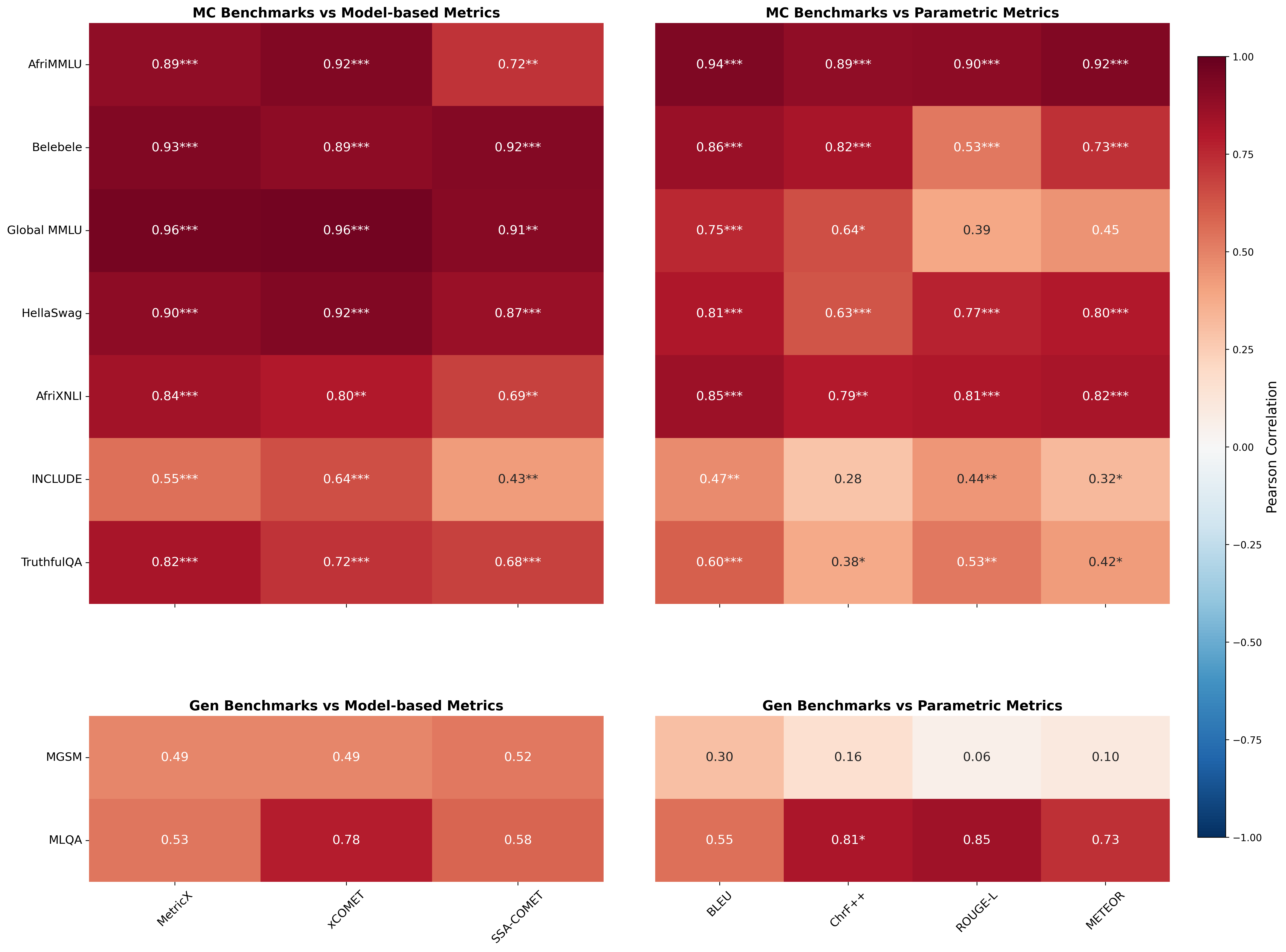}
    \caption{DeepSeek-R1-Distill-Llama-70B}
    \label{fig:ds-70b}
  \end{subfigure}

  \caption{Pearson correlations between MT metrics and benchmark tasks.
  Stars indicate significance: * $p<0.05$, ** $p<0.01$, *** $p<0.001$.}
  \label{fig:deepseek-family}
\end{figure}

Deepseek models align with general trends (low MGSM scores) but also present unique patterns.

\noindent\textbf{Similar Trends}: Both the 70B and 32B models show high and significant correlations for the African-centric benchmarks (AfriMMLU and AfriXNLI) and HellaSwag.

\subsection{Qwen3 Family}
\begin{figure}[!t]
  \centering
    \begin{adjustbox}{max totalsize={\textwidth}{0.9\textheight},center}
    \begin{minipage}{\linewidth}

  \begin{subfigure}[t]{\linewidth}
    \centering
    \includegraphics[width=\linewidth]{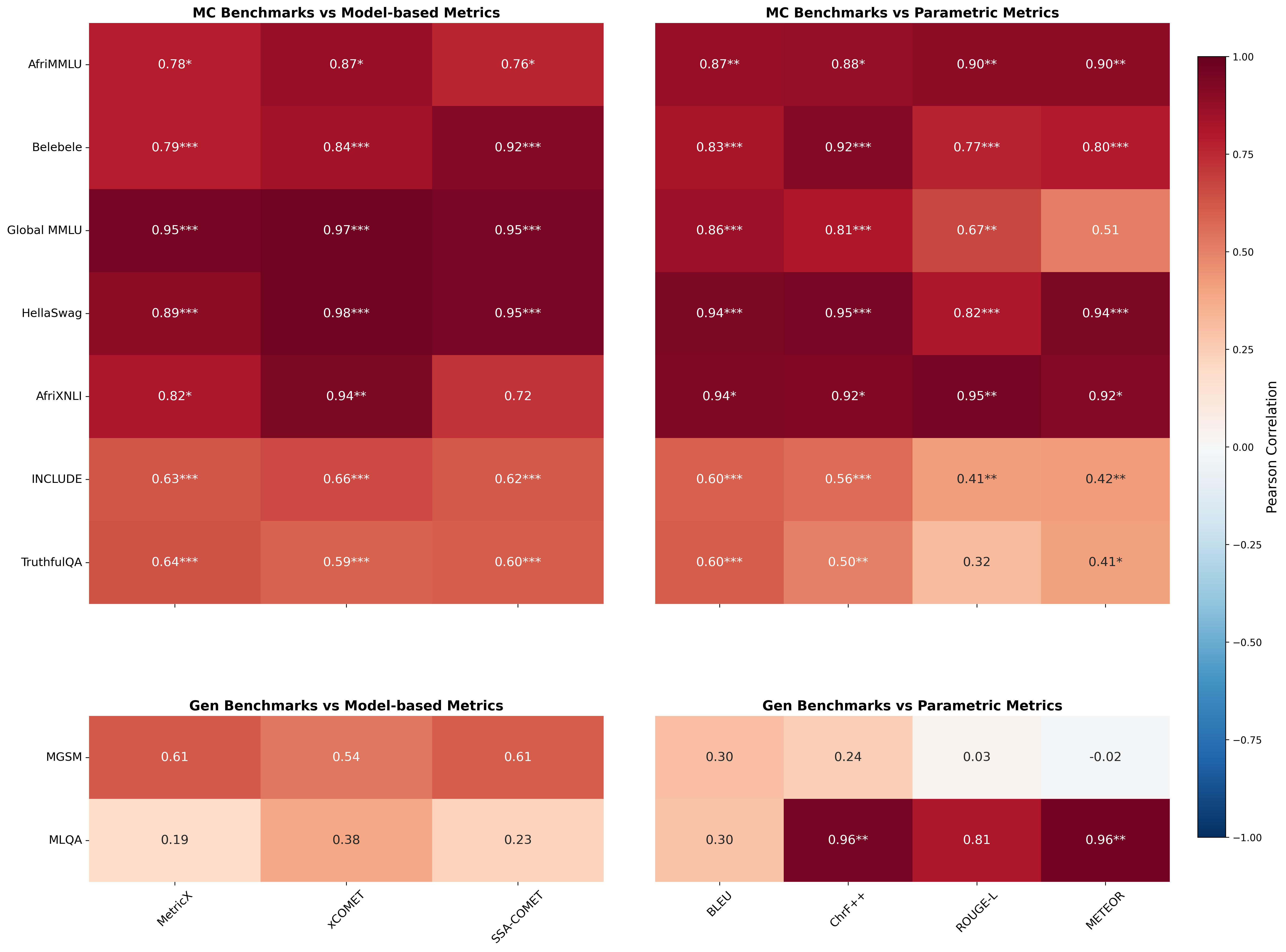}
    \caption{Qwen3-4B}
    \label{fig:qwen-4b-flores}
  \end{subfigure}

  \vspace{0.6em}

  \begin{subfigure}[t]{\linewidth}
    \centering
    \includegraphics[width=\linewidth]{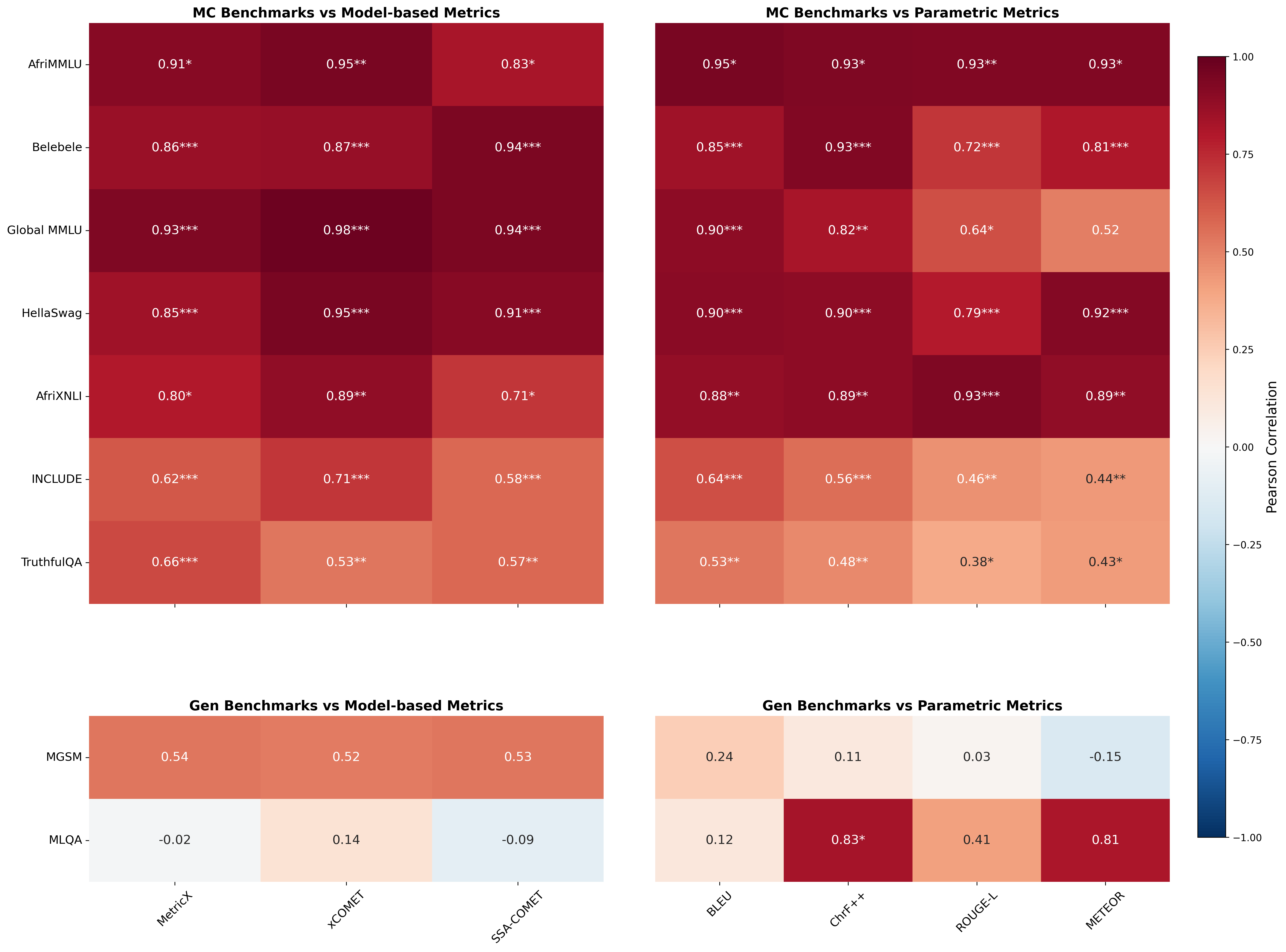}
    \caption{Qwen3-8B}
    \label{fig:qwen-8b}
  \end{subfigure}

  \vspace{0.6em}

  \begin{subfigure}[t]{\linewidth}
    \centering
    \includegraphics[width=\linewidth]{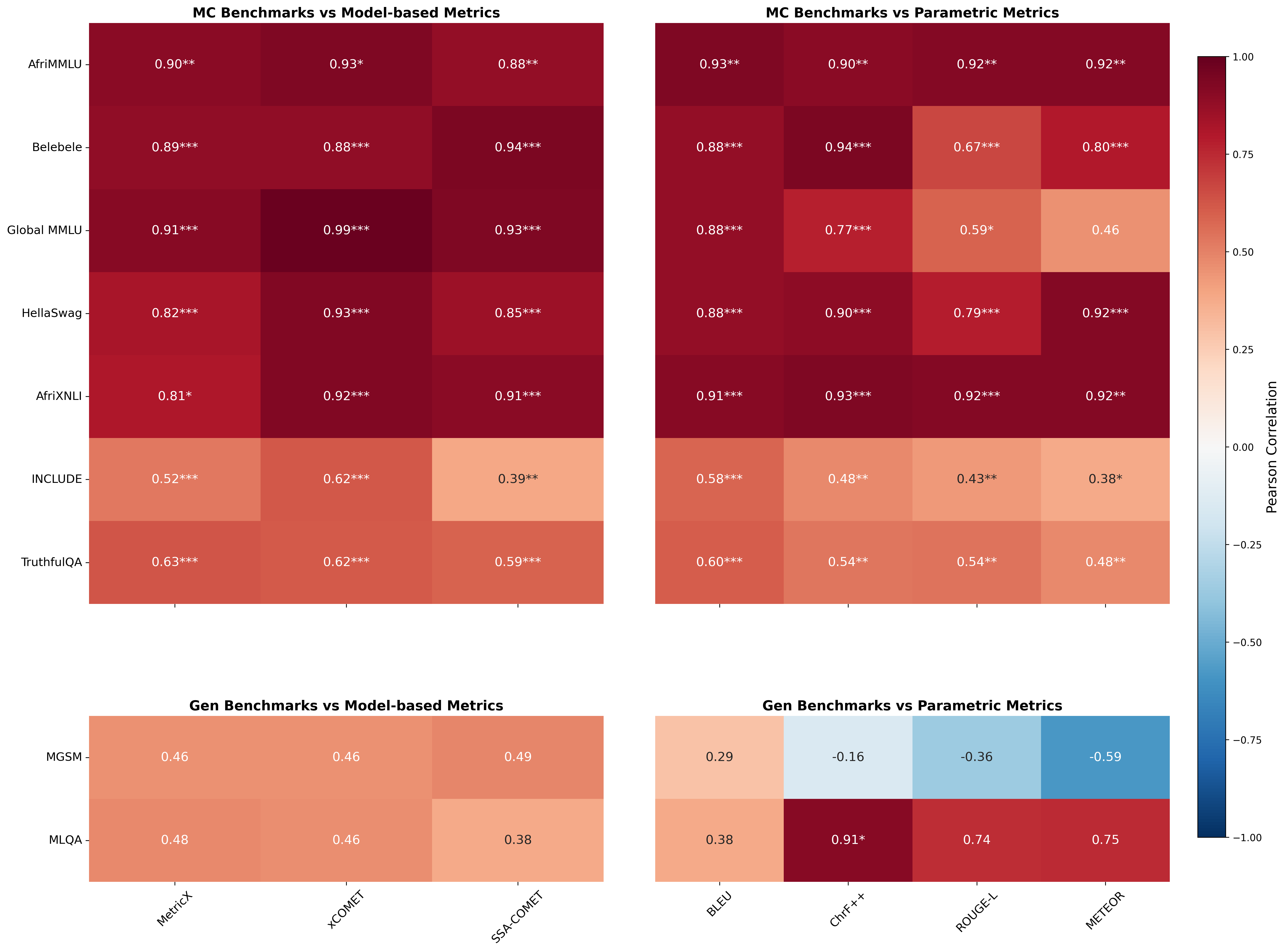}
    \caption{Qwen3-14B}
    \label{fig:qwen-14b}
  \end{subfigure}

    \vspace{0.6em}

  \begin{subfigure}[t]{\linewidth}
    \centering
    \includegraphics[width=\linewidth]{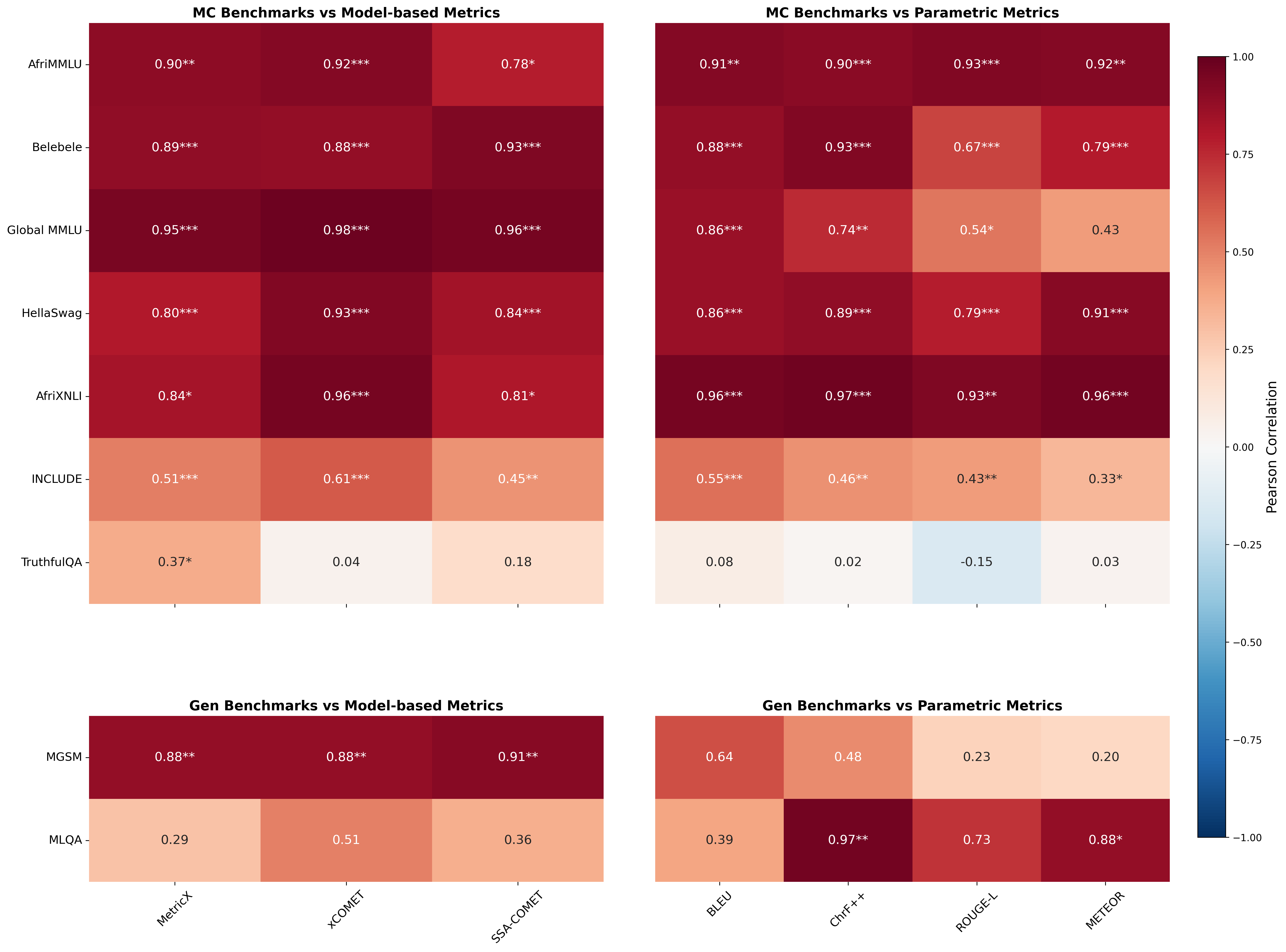}
    \caption{Qwen3-30B-A3B}
    \label{fig:qwen-30b-a3b-flores}
  \end{subfigure}

  \vspace{0.6em}

    \begin{subfigure}[t]{\linewidth}
    \centering
    \includegraphics[width=\linewidth]{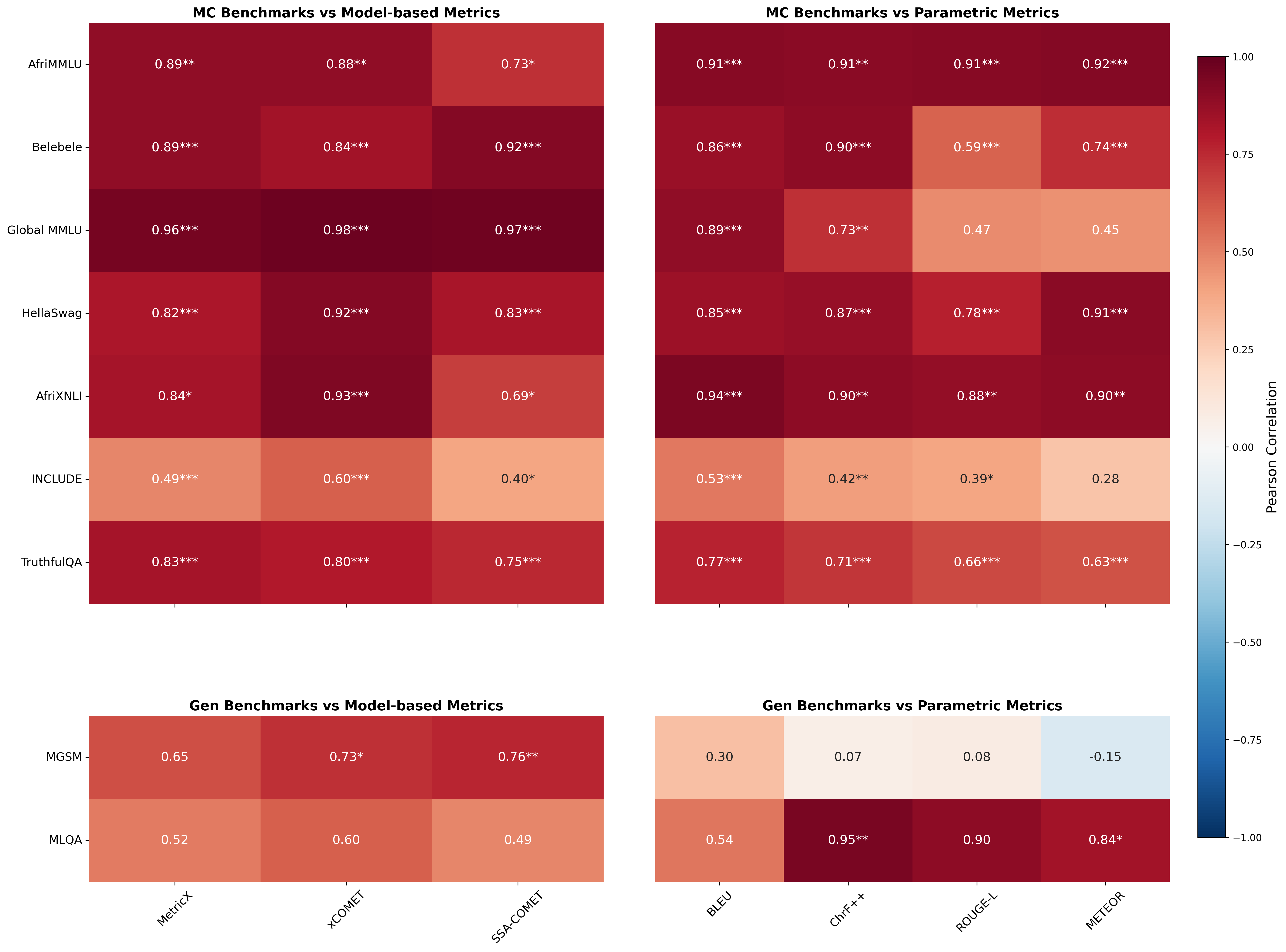}
    \caption{Qwen3-32B}
    \label{fig:qwen-32b-flores}
  \end{subfigure}

    \end{minipage}
    \end{adjustbox}
    
  \caption{Pearson correlations between MT metrics and benchmark tasks.
  Stars indicate significance: * $p<0.05$, ** $p<0.01$, *** $p<0.001$.}
  \label{fig:qwen-family}
\end{figure}
The smaller models in the Qwen family replicate the trends of Phi-4, demonstrating that these correlations are stable within the model family, though some weakening is observed at the smallest scale.

\noindent \textbf{Similar Trends}: The African-centric benchmarks (AfriMMLU and AfriXNLI) maintain extremely high and significant correlations (r > 0.9 in many cases). For XComet and those benchmarks, it emerges as a top correlation across the entire Qwen Family. It also follows a similar trend with MGSM, where it continues to show the weakest correlations, none of which are statistically significant. For the 32B model, metrics like ChrF++ and Bleu show moderately strong positive correlations with most benchmarks, like in AfriMMLU (r=0.91,p< 0.001)and AfriXNLI (r=0.94, p< 0.001).

\subsection{Gemma3 Family}

\begin{figure}[!t]
  \centering
  \begin{adjustbox}{max totalsize={\textwidth}{0.9\textheight},center}
    \begin{minipage}{\linewidth}

  \begin{subfigure}[t]{\linewidth}
    \centering
    \includegraphics[width=\linewidth]{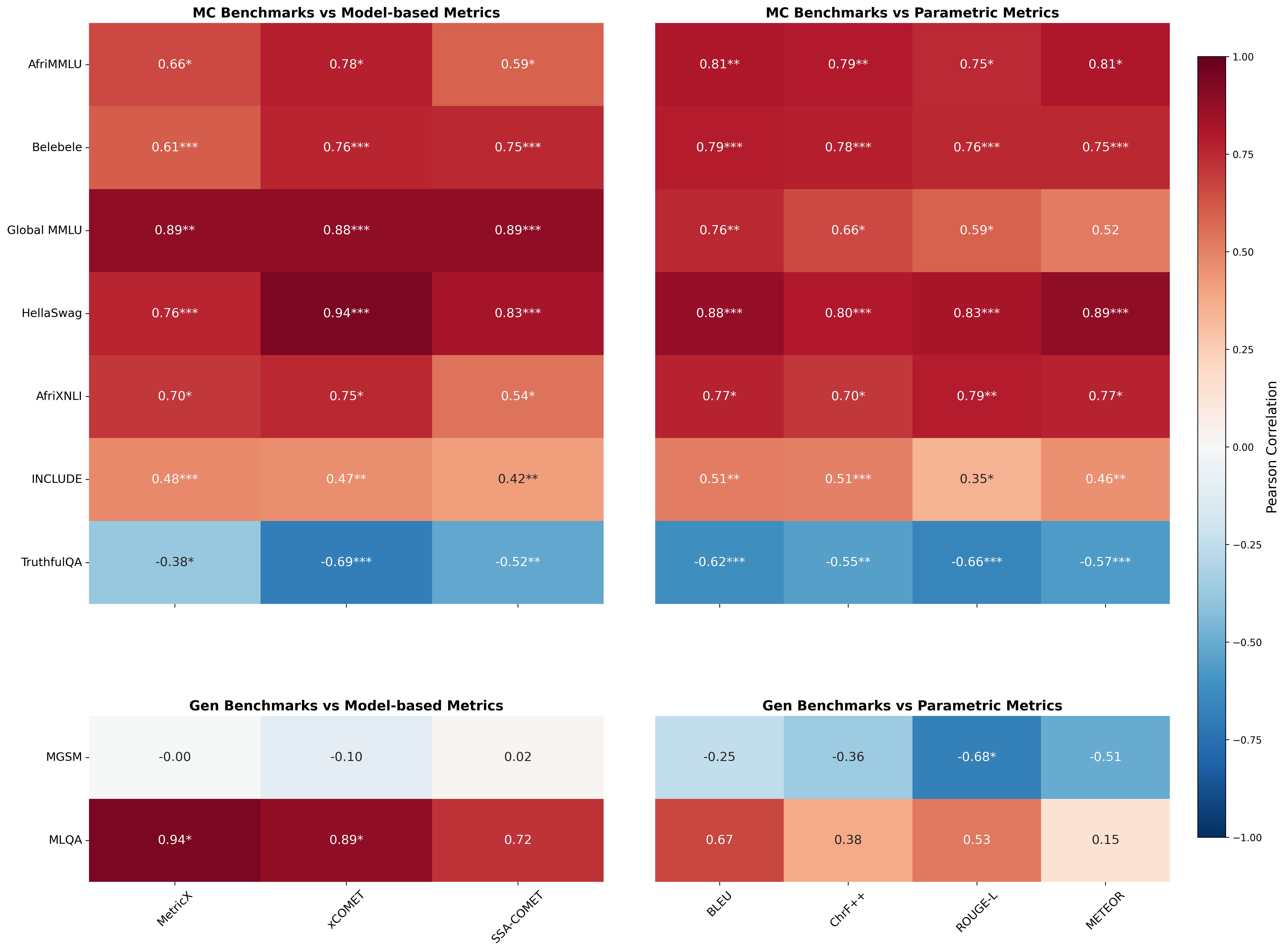}
    \caption{Gemma 3 1B-IT}
    \label{fig:gemma3-1b}
  \end{subfigure}

    \vspace{0.6em}

    \begin{subfigure}[t]{\linewidth}
    \centering
    \includegraphics[width=\linewidth]{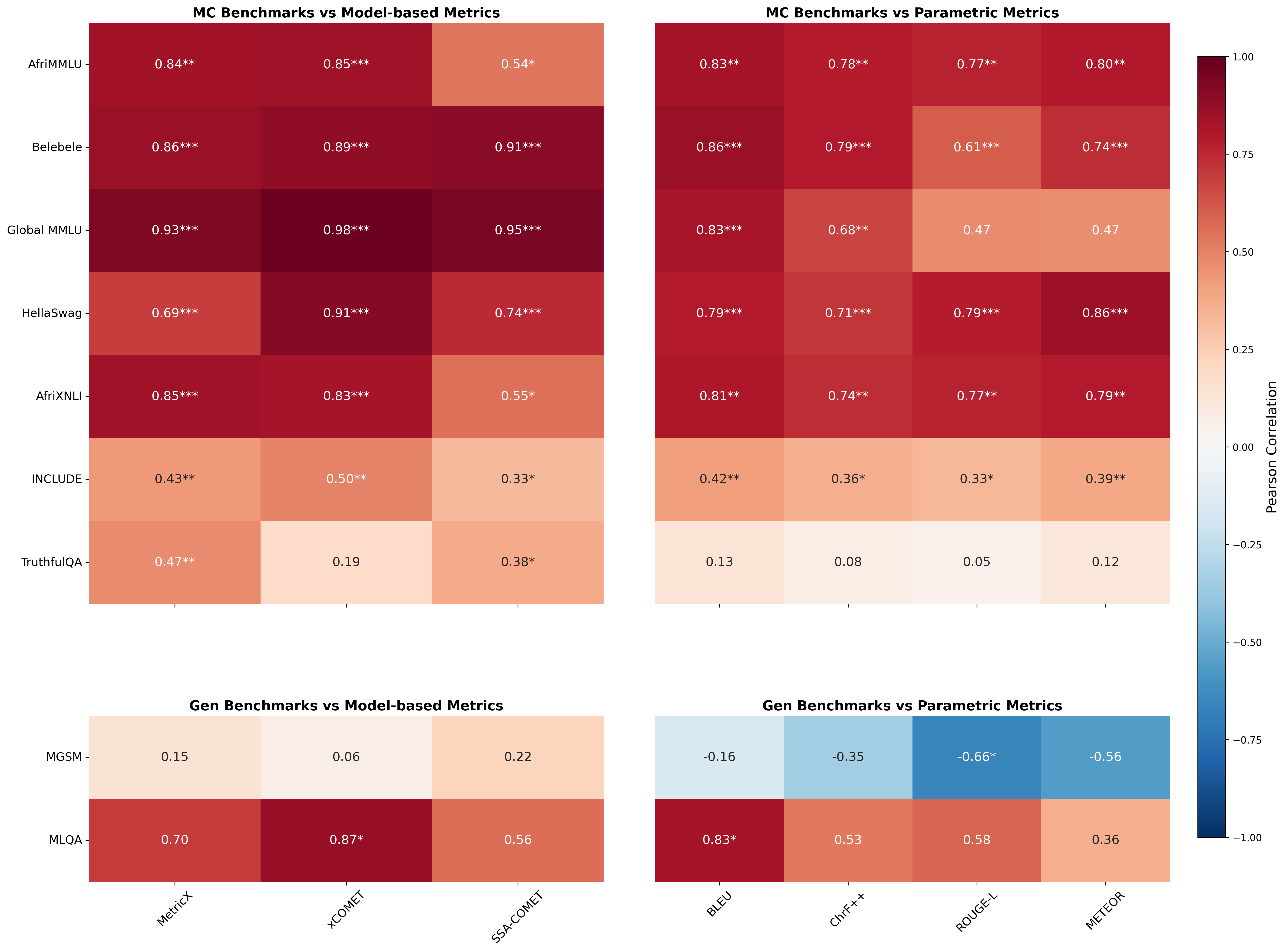}
    \caption{Gemma 3 4B-IT}
    \label{fig:gemma3-4b}
  \end{subfigure}
  
  \vspace{0.6em}
  
    \begin{subfigure}[t]{\linewidth}
    \centering
    \includegraphics[width=\linewidth]{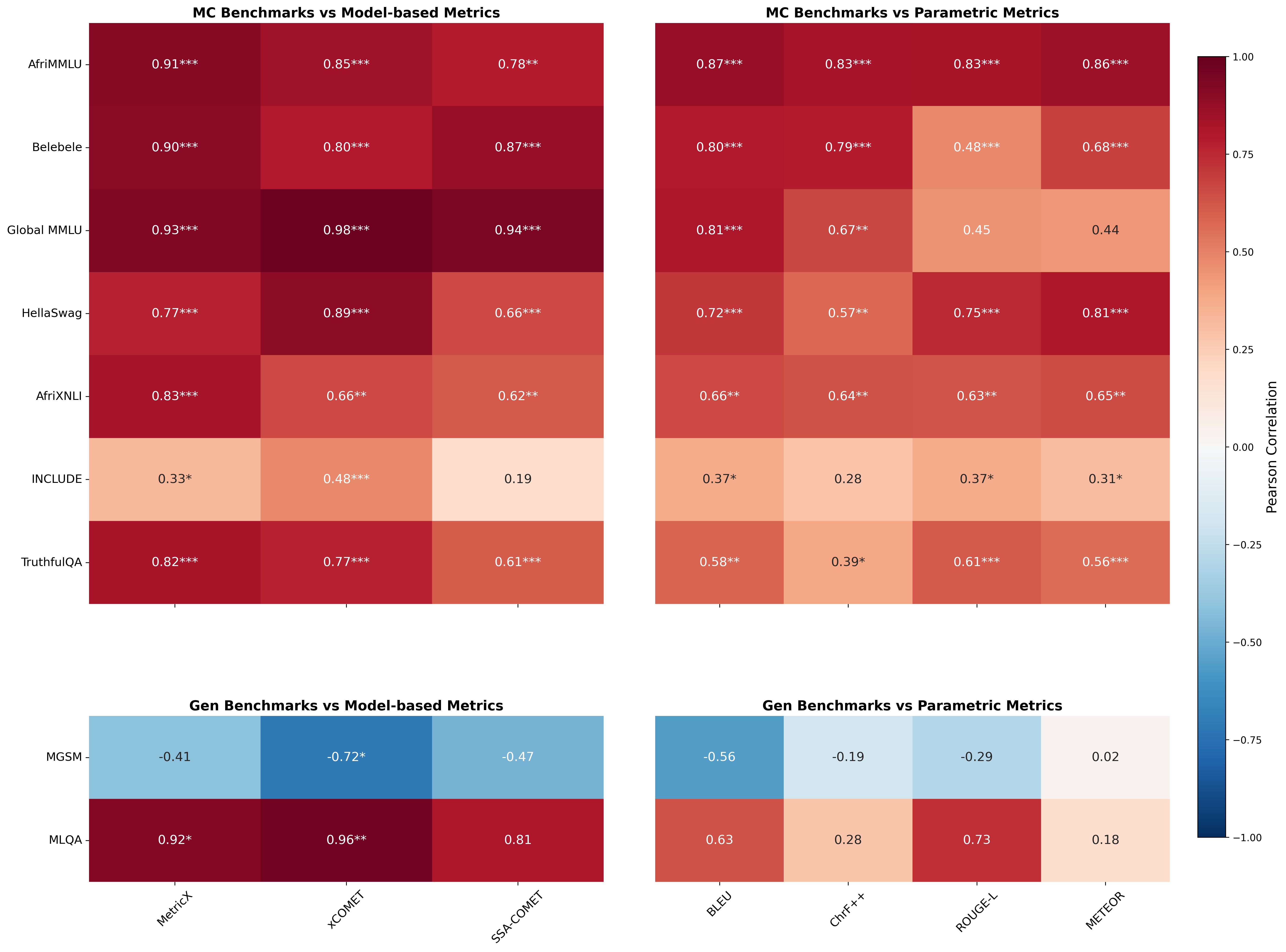}
    \caption{Gemma 3 12B-IT}
    \label{fig:gemma3-12b}
  \end{subfigure}

  \vspace{0.6em}

    \begin{subfigure}[t]{\linewidth}
    \centering
    \includegraphics[width=\linewidth]{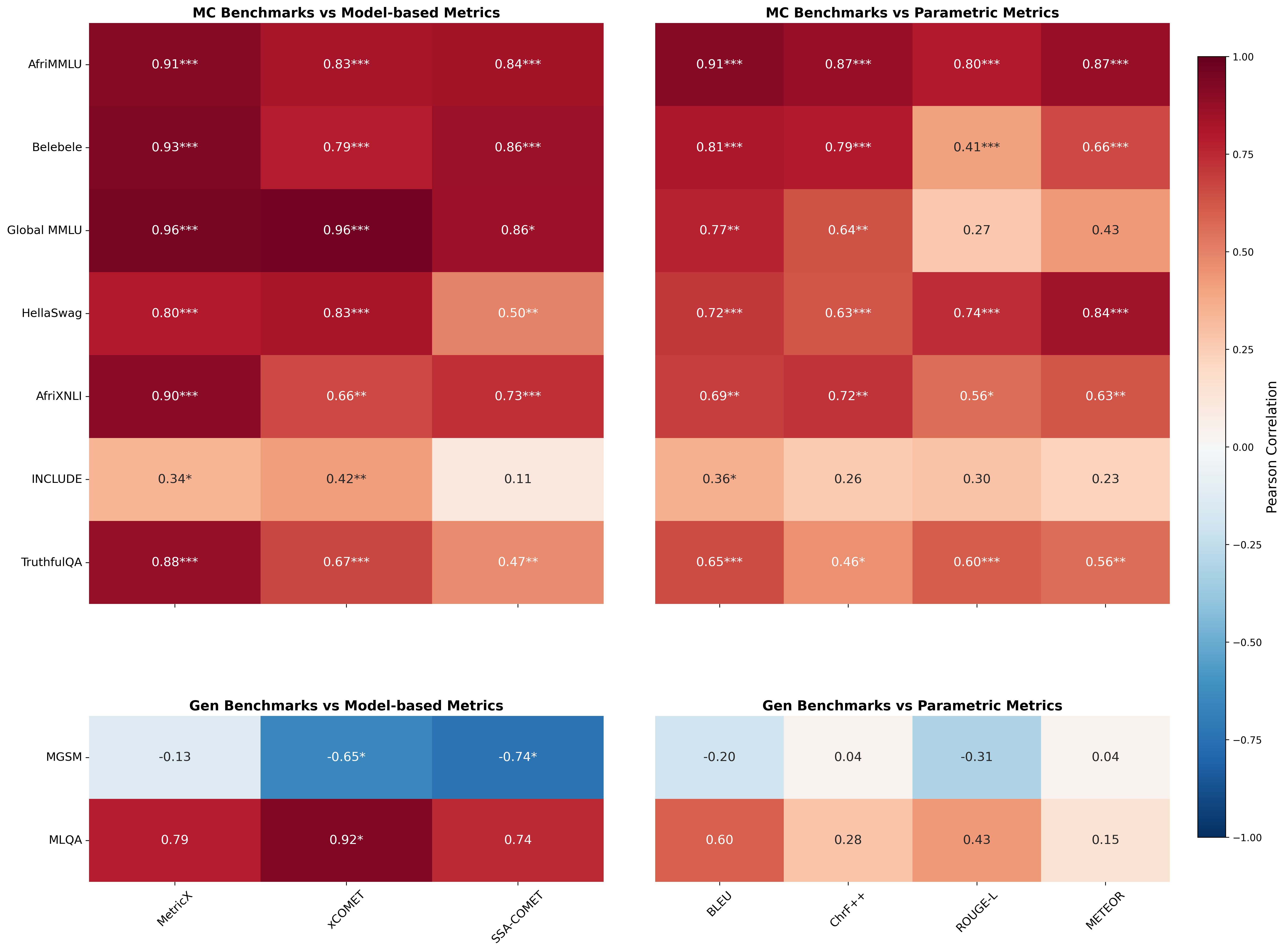}
    \caption{Gemma 3 27B-IT}
    \label{fig:gemma3-27b}
  \end{subfigure}

    \end{minipage}
    \end{adjustbox}
  \caption{Pearson correlations between MT metrics and benchmark tasks.
  Stars indicate significance: * $p<0.05$, ** $p<0.01$, *** $p<0.001$.}
  \label{fig:gemma3-family}
\end{figure}

When looking at the Gemma-3 Family, it provides at some insight into the effect of model scaling. The trends observed in the flagship model and other large models gold for the mid-sized Gemma models but degraded as the model shrinks. Something interesting happens when it we get to the 27B model however, because it returns to chaotic results that mirror the 1B model.

\noindent \textbf{Similar Trends}: For the Gemma3-12B model, it behaves similary to Qwen with benchmarks like AfriMMLU and AfriXNLI having strong, significant positive correlations. 

\noindent \textbf{Differences}: The Gemma3-1B model represents a near complete breakdown of the trends. In Gemma3-27B, we see that SSA-Comet and MGSM have a low, negative score (r= -0.74, p<0.05).

\subsection{Phi4}
\begin{figure}[H]
  \centering
  \begin{subfigure}[t]{\linewidth}
    \centering
    \includegraphics[width=\linewidth]{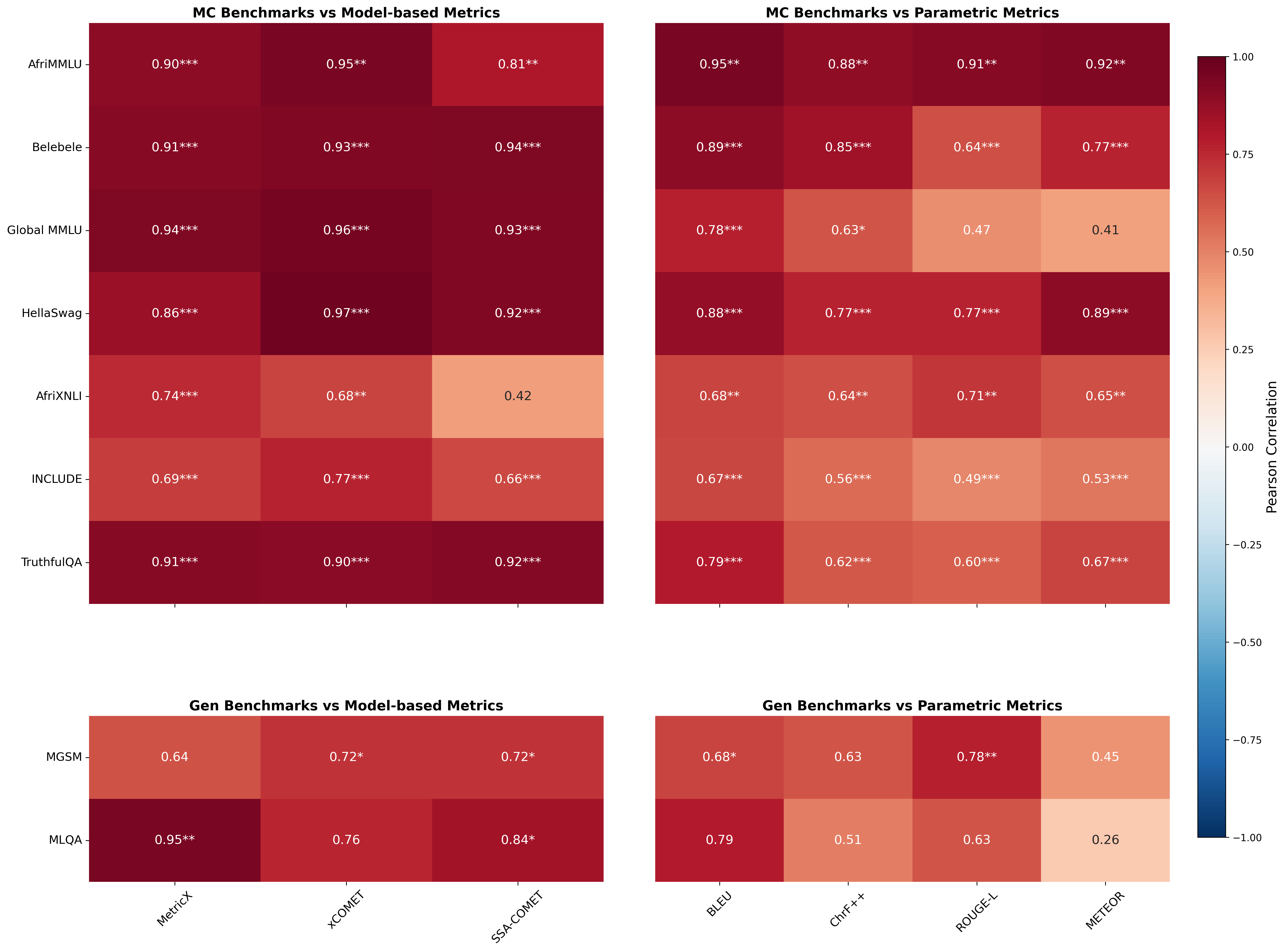}
    \caption{Phi-4}
    \label{fig:phi4-flores}
  \end{subfigure}

  \caption{Pearson correlations between MT metrics and benchmark tasks.
  Stars indicate significance: * $p<0.05$, ** $p<0.01$, *** $p<0.001$.}
  \label{fig:phi4}
\end{figure}

The Phi-4 model is an interesting case, showing strong and consistent correlations across nearly all benchmarks, including those that were challenging for other models 

\noindent \textbf{Similar Trends}: Like usual, Phi-4 shows strong correlations with benchmarks like AfriMMLU and AfriXNLI. 

\noindent \textbf{Differences}: The most significant outlier is its performance on MGSM. Unlike almost every other model, Phi-4 shows a very strong and statistically significant correlation between MLQA and SSA-Comet (r=0.84, p < 0.05). Furthermore, MGSM also shows remarkably strong and consistent correlations across multiple metrics (r=0.78 with Rough-L), which was not the case for Qwen-72B.

\section{Additional Correlations Between Individual Metric and Benchmark}

Here we provide detailed individual translation metrics against performance benchmark scatter plots for further insights and analysis. For each benchmark, we select the metric that yield the best correlation result and present them below. 
    
\onecolumn

\begin{landscape}
    \section{Supported Languages per Benchmark}
    The table lists the 120 unique language included in our final evaluation set, which were selected based on their presence across the various benchmarks. While some sources such as Flores-200 offer a much broader range, we focused on the intersection of languages available in the parallel translation corpora (Flores-200, WMT24++, or NTRE) and the other benchmarks (AfriMMLU, Afrixnli, Belebele, etc.). Languages that did not appear in both a parallel resource and a benchmark task were exclude as translation was required for our evaluation setup.
    The final column, Resources indicates the resource class as defined from \citet{joshi2021statefatelinguisticdiversity}. This scale ranges from Class 0 (lowest resource) up to Class 5 (highest resource). The character \(*\) was put down for languages that did not appear.

    For language that appeared with multiple script variants (e.g.  Hindi with Deva and Latn scripts), we prioritized the native script and excluded the Romanized version. 
                
    { 
    \setlength{\tabcolsep}{6pt}

    \begin{longtable}{lccccccccccccc}
        \caption{All Languages Included in Experiments.}
        \label{tab:supported_languages} \\
        \hline
        Language & 
        \adjustbox{angle=90}{Flores-200} & 
        \adjustbox{angle=90}{WMT24++} & 
        \adjustbox{angle=90}{NTREX} & 
        \adjustbox{angle=90}{AfriMMLU} & 
        \adjustbox{angle=90}{Afrixnli} & 
        \adjustbox{angle=90}{Belebele} & 
        \adjustbox{angle=90}{GlobalMMLU} & 
        \adjustbox{angle=90}{HellaSwag} & 
        \adjustbox{angle=90}{MGSM} &
        \adjustbox{angle=90}{MLQA} & 
        \adjustbox{angle=90}{TruthfulQA} & 
        \adjustbox{angle=90}{INCLUDE} & 
        \adjustbox{angle=90}{Resources} 
        \\ 
        \hline
        Afrikaans & afr\_Latn & --- & afr\_Latn & --- & --- & afr\_Latn & --- & --- & --- & --- & --- & --- & 3 \\
        Amharic & amh\_Ethi & --- & amh\_Ethi & amh & amh & amh\_Ethi & --- & --- & --- & --- & --- & --- & 2 \\
        Armenian & hye\_Armn & --- & hye\_Armn & --- & --- & hye\_Armn & --- & hy & --- & --- & hy & Armenian & 1 \\
        Assamese & asm\_Beng & --- & --- & --- & --- & asm\_Beng & --- & --- & --- & --- & --- & --- & 1 \\
        Bambara & bam\_Latn & --- & --- & --- & --- & bam\_Latn & --- & --- & --- & --- & --- & --- & 1 \\
        Basque & eus\_Latn & --- & eus\_Latn & --- & --- & eus\_Latn & --- & eu & --- & --- & eu & Basque & 4 \\
        Belarusian & bel\_Cyrl & --- & --- & --- & --- & --- & --- & --- & --- & --- & --- & Belarusian & 3 \\
        Bengali & ben\_Beng & bn\_IN & ben\_Beng & --- & --- & ben\_Beng & bn & bn & bn & --- & bn & Bengali & 3 \\
        Bulgarian & bul\_Cyrl & bg\_BG & bul\_Cyrl & --- & --- & bul\_Cyrl & --- & --- & --- & --- & --- & Bulgarian & 3 \\
        Burmese & mya\_Mymr & --- & mya\_Mymr & --- & --- & mya\_Mymr & --- & --- & --- & --- & --- & --- & 1 \\
        Catalan & cat\_Latn & ca\_ES & cat\_Latn & --- & --- & cat\_Latn & --- & ca & --- & --- & ca & --- & 4 \\
        Cebuano & ceb\_Latn & --- & --- & --- & --- & ceb\_Latn & --- & --- & --- & --- & --- & --- & 3 \\
        Central Kurdish & ckb\_Arab & --- & ckb\_Arab & --- & --- & ckb\_Arab & --- & --- & --- & --- & --- & --- & 1 \\
        Chinese (Simplified) & zho\_Hans & zh\_CN & zho\_Hans & --- & --- & zho\_Hans & zh & --- & zh & zh & zh & Chinese & 5 \\
        Chinese (Traditional) & zho\_Hant & zh\_TW & zho\_Hant & --- & --- & zho\_Hant & --- & --- & --- & --- & --- & --- & 5 \\
        Croatian & hrv\_Latn & hr\_HR & hrv\_Latn & --- & --- & hrv\_Latn & --- & hr & --- & --- & hr & Croatian & 4 \\
        Czech & ces\_Latn & cs\_CZ & ces\_Latn & --- & --- & ces\_Latn & --- & --- & --- & --- & --- & --- & 4 \\
        Danish & dan\_Latn & da\_DK & dan\_Latn & --- & --- & dan\_Latn & --- & da & --- & --- & da & --- & 3 \\
        Dutch & nld\_Latn & nl\_NL & nld\_Latn & --- & --- & nld\_Latn & --- & nl & --- & --- & nl & Dutch & 4 \\
        Eastern Panjabi & pan\_Guru & pa\_IN & pan\_Guru & --- & --- & pan\_Guru & --- & --- & --- & --- & --- & --- & 2 \\
        Egyptian Arabic & arz\_Arab & ar\_EG & --- & --- & --- & arz\_Arab & --- & --- & --- & --- & --- & --- & 3 \\
        English & eng\_Latn & --- & eng\_Latn & eng & eng & eng\_Latn & en & --- & --- & en & --- & --- & 5 \\
        Estonian & est\_Latn & et\_EE & est\_Latn & --- & --- & est\_Latn & --- & --- & --- & --- & --- & Estonian & 3 \\
        Ewe & ewe\_Latn & --- & ewe\_Latn & ewe & ewe & --- & --- & --- & --- & --- & --- & --- & 1 \\
        Finnish & fin\_Latn & fi\_FI & fin\_Latn & --- & --- & fin\_Latn & --- & --- & --- & --- & --- & Finnish & 4 \\
        French & fra\_Latn & fr\_FR & fra\_Latn & fra & fra & fra\_Latn & fr & fr & fr & --- & fr & French & 5 \\
        Georgian & kat\_Geor & --- & kat\_Geor & --- & --- & kat\_Geor & --- & --- & --- & --- & --- & Georgian & 3 \\
        German & deu\_Latn & de\_DE & deu\_Latn & --- & --- & deu\_Latn & de & de & de & de & de & German & 5 \\
        Greek & ell\_Grek & el\_GR & ell\_Grek & --- & --- & ell\_Grek & --- & --- & --- & --- & --- & Greek & 3 \\
        Guarani & grn\_Latn & --- & --- & --- & --- & grn\_Latn & --- & --- & --- & --- & --- & --- & 1 \\
        Gujarati & guj\_Gujr & gu\_IN & guj\_Gujr & --- & --- & guj\_Gujr & --- & gu & --- & --- & gu & --- & 1 \\
        Haitian Creole & hat\_Latn & --- & --- & --- & --- & hat\_Latn & --- & --- & --- & --- & --- & --- & 0 \\
        Halh Mongolian & khk\_Cyrl & --- & mon\_Mong & --- & --- & khk\_Cyrl & --- & --- & --- & --- & --- & --- & 1 \\
        Hausa & hau\_Latn & --- & hau\_Latn & hau & hau & hau\_Latn & --- & --- & --- & --- & --- & --- & 2 \\
        Hebrew & heb\_Hebr & he\_IL & heb\_Hebr & --- & --- & heb\_Hebr & --- & --- & --- & --- & --- & Hebrew & 3 \\
        Hindi & hin\_Deva & hi\_IN & hin\_Deva & --- & --- & hin\_Deva & hi & hi & --- & hi & hi & Hindi & 4 \\
        Hungarian & hun\_Latn & hu\_HU & hun\_Latn & --- & --- & hun\_Latn & --- & hu & --- & --- & hu & Hungarian & 4 \\
        Icelandic & isl\_Latn & is\_IS & isl\_Latn & --- & --- & isl\_Latn & --- & --- & --- & --- & --- & --- & 2 \\
        Igbo & ibo\_Latn & --- & ibo\_Latn & ibo & ibo & ibo\_Latn & --- & --- & --- & --- & --- & --- & 1 \\
        Ilocano & ilo\_Latn & --- & --- & --- & --- & ilo\_Latn & --- & --- & --- & --- & --- & --- & 1 \\
        Indonesian & ind\_Latn & id\_ID & ind\_Latn & --- & --- & ind\_Latn & id & id & --- & --- & id & Indonesian & 3 \\
        Italian & ita\_Latn & it\_IT & ita\_Latn & --- & --- & ita\_Latn & it & it & --- & --- & it & Italian & 4 \\
        Japanese & jpn\_Jpan & ja\_JP & jpn\_Jpan & --- & --- & jpn\_Jpan & ja & --- & ja & --- & --- & Japanese & 5 \\
        Javanese & jav\_Latn & --- & --- & --- & --- & jav\_Latn & --- & --- & --- & --- & --- & --- & 1 \\
        Jingpho & kac\_Latn & --- & --- & --- & --- & kac\_Latn & --- & --- & --- & --- & --- & --- & 0 \\
        Kabuverdianu & kea\_Latn & --- & --- & --- & --- & kea\_Latn & --- & --- & --- & --- & --- & --- & 0 \\
        Kannada & kan\_Knda & kn\_IN & kan\_Knda & --- & --- & kan\_Knda & --- & kn & --- & --- & kn & --- & 1 \\
        Kazakh & kaz\_Cyrl & --- & kaz\_Cyrl & --- & --- & kaz\_Cyrl & --- & --- & --- & --- & --- & Kazakh & 3 \\
        Khmer & khm\_Khmr & --- & khm\_Khmr & --- & --- & khm\_Khmr & --- & --- & --- & --- & --- & --- & 1 \\
        Kinyarwanda & kin\_Latn & --- & kin\_Latn & kin & kin & kin\_Latn & --- & --- & --- & --- & --- & --- & 1 \\
        Korean & kor\_Hang & ko\_KR & kor\_Hang & --- & --- & kor\_Hang & ko & --- & --- & --- & --- & Korean & 4 \\
        Kyrgyz & kir\_Cyrl & --- & kir\_Cyrl & --- & --- & kir\_Cyrl & --- & --- & --- & --- & --- & --- & 1 \\
        Lao & lao\_Laoo & --- & lao\_Laoo & --- & --- & lao\_Laoo & --- & --- & --- & --- & --- & --- & 2 \\
        Lingala & lin\_Latn & --- & --- & lin & lin & lin\_Latn & --- & --- & --- & --- & --- & --- & 1 \\
        Lithuanian & lit\_Latn & lt\_LT & lit\_Latn & --- & --- & lit\_Latn & --- & --- & --- & --- & --- & Lithuanian & 3 \\
        Luganda & lug\_Latn & --- & --- & lug & lug & lug\_Latn & --- & --- & --- & --- & --- & --- & 1 \\
        Luo & luo\_Latn & --- & --- & --- & --- & luo\_Latn & --- & --- & --- & --- & --- & --- & * \\
        Macedonian & mkd\_Cyrl & --- & mkd\_Cyrl & --- & --- & mkd\_Cyrl & --- & --- & --- & --- & --- & North Macedonian & 1 \\
        Malayalam & mal\_Mlym & ml\_IN & mal\_Mlym & --- & --- & mal\_Mlym & --- & ml & --- & --- & ml & Malayalam & 1 \\
        Maltese & mlt\_Latn & --- & mlt\_Latn & --- & --- & mlt\_Latn & --- & --- & --- & --- & --- & --- & 2 \\
        Maori & mri\_Latn & --- & mri\_Latn & --- & --- & mri\_Latn & --- & --- & --- & --- & --- & --- & 1 \\
        Marathi & mar\_Deva & mr\_IN & mar\_Deva & --- & --- & mar\_Deva & --- & mr & --- & --- & mr & --- & 2 \\
        Mesopotamian Arabic & acm\_Arab & --- & --- & --- & --- & acm\_Arab & --- & --- & --- & --- & --- & --- & * \\
        Modern Standard Arabic & arb\_Arab & ar\_SA & arb\_Arab & --- & --- & arb\_Arab & ar & ar & --- & ar & ar & Arabic & 5 \\
        Moroccan Arabic & ary\_Arab & --- & --- & --- & --- & ary\_Arab & --- & --- & --- & --- & --- & --- & * \\
        Najdi Arabic & ars\_Arab & --- & --- & --- & --- & ars\_Arab & --- & --- & --- & --- & --- & --- & * \\
        Nepali & npi\_Deva & --- & --- & --- & --- & npi\_Deva & --- & ne & --- & --- & ne & Nepali & 1 \\
        Nigerian Fulfulde & fuv\_Latn & --- & --- & --- & --- & fuv\_Latn & --- & --- & --- & --- & --- & --- & 0 \\
        North Azerbaijani & azj\_Latn & --- & aze\_Latn & --- & --- & azj\_Latn & --- & --- & --- & --- & --- & Azerbaijani & 1 \\
        North Levantine Arabic & apc\_Arab & --- & --- & --- & --- & apc\_Arab & --- & --- & --- & --- & --- & --- & * \\
        Northern Sotho & nso\_Latn & --- & nso\_Latn & --- & --- & nso\_Latn & --- & --- & --- & --- & --- & --- & 1 \\
        Northern Uzbek & uzn\_Latn & --- & --- & --- & --- & uzn\_Latn & --- & --- & --- & --- & --- & Uzbek & 3 \\
        Norwegian Bokmål & nob\_Latn & no\_NO & nob\_Latn & --- & --- & nob\_Latn & --- & --- & --- & --- & --- & --- & 1 \\
        Nyanja & nya\_Latn & --- & nya\_Latn & --- & --- & nya\_Latn & --- & --- & --- & --- & --- & --- & 1 \\
        Oriya & ory\_Orya & --- & --- & --- & --- & ory\_Orya & --- & --- & --- & --- & --- & --- & 1 \\
        Plateau Malagasy & plt\_Latn & --- & --- & --- & --- & plt\_Latn & --- & --- & --- & --- & --- & --- & 1 \\
        Polish & pol\_Latn & pl\_PL & pol\_Latn & --- & --- & pol\_Latn & --- & --- & --- & --- & --- & Polish & 4 \\
        Portuguese & por\_Latn & pt\_PT & por\_Latn & --- & --- & por\_Latn & pt & pt & --- & --- & pt & Portuguese & 4 \\
        Romanian & ron\_Latn & ro\_RO & ron\_Latn & --- & --- & ron\_Latn & --- & ro & --- & --- & ro & --- & 3 \\
        Russian & rus\_Cyrl & ru\_RU & rus\_Cyrl & --- & --- & rus\_Cyrl & --- & ru & ru & --- & ru & Russian & 4 \\
        Serbian & srp\_Cyrl & sr\_RS & srp\_Cyrl & --- & --- & srp\_Cyrl & --- & sr & --- & --- & sr & Serbian & 4 \\
        Shan & shn\_Mymr & --- & --- & --- & --- & shn\_Mymr & --- & --- & --- & --- & --- & --- & 0 \\
        Shona & sna\_Latn & --- & sna\_Latn & sna & sna & sna\_Latn & --- & --- & --- & --- & --- & --- & 1 \\
        Sindhi & snd\_Arab & --- & snd\_Arab & --- & --- & snd\_Arab & --- & --- & --- & --- & --- & --- & 1 \\
        Sinhala & sin\_Sinh & --- & sin\_Sinh & --- & --- & sin\_Sinh & --- & --- & --- & --- & --- & --- & 0 \\
        Slovak & slk\_Latn & sk\_SK & slk\_Latn & --- & --- & slk\_Latn & --- & sk & --- & --- & sk & --- & 3 \\
        Slovenian & slv\_Latn & sl\_SI & slv\_Latn & --- & --- & slv\_Latn & --- & --- & --- & --- & --- & --- & 3 \\
        Somali & som\_Latn & --- & som\_Latn & --- & --- & som\_Latn & --- & --- & --- & --- & --- & --- & 1 \\
        Southern Pashto & pbt\_Arab & --- & --- & --- & --- & pbt\_Arab & --- & --- & --- & --- & --- & --- & 1 \\
        Southern Sotho & sot\_Latn & --- & --- & sot & sot & sot\_Latn & --- & --- & --- & --- & --- & --- & 1 \\
        Spanish & spa\_Latn & es\_MX & spa\_Latn & --- & --- & spa\_Latn & es & es & es & es & es & Spanish & 5 \\
        Standard Latvian & lvs\_Latn & lv\_LV & lav\_Latn & --- & --- & lvs\_Latn & --- & --- & --- & --- & --- & --- & 3 \\
        Standard Malay & zsm\_Latn & --- & msa\_Latn & --- & --- & zsm\_Latn & --- & --- & --- & --- & --- & Malay & 3 \\
        Standard Tibetan & bod\_Tibt & --- & bod\_Tibt & --- & --- & bod\_Tibt & --- & --- & --- & --- & --- & --- & 1 \\
        Sundanese & sun\_Latn & --- & --- & --- & --- & sun\_Latn & --- & --- & --- & --- & --- & --- & 1 \\
        Swahili & swh\_Latn & sw\_KE & swa\_Latn & swa & swa & swh\_Latn & sw & --- & sw & --- & --- & --- & 2 \\
        Swati & ssw\_Latn & --- & ssw\_Latn & --- & --- & ssw\_Latn & --- & --- & --- & --- & --- & --- & 1 \\
        Swedish & swe\_Latn & sv\_SE & swe\_Latn & --- & --- & swe\_Latn & --- & sv & --- & --- & sv & --- & 4 \\
        Tagalog & tgl\_Latn & --- & --- & --- & --- & tgl\_Latn & --- & --- & --- & --- & --- & Tagalog & 3 \\
        Tajik & tgk\_Cyrl & --- & tgk\_Cyrl & --- & --- & tgk\_Cyrl & --- & --- & --- & --- & --- & --- & 1 \\
        Tamil & tam\_Taml & ta\_IN & tam\_Taml & --- & --- & tam\_Taml & --- & ta & --- & --- & ta & Tamil & 3 \\
        Telugu & tel\_Telu & te\_IN & tel\_Telu & --- & --- & tel\_Telu & --- & te & te & --- & te & Telugu & 1 \\
        Thai & tha\_Thai & th\_TH & tha\_Thai & --- & --- & tha\_Thai & --- & --- & th & --- & --- & --- & 3 \\
        Tigrinya & tir\_Ethi & --- & tir\_Ethi & --- & --- & tir\_Ethi & --- & --- & --- & --- & --- & --- & 2 \\
        Tosk Albanian & als\_Latn & --- & sqi\_Latn & --- & --- & als\_Latn & --- & --- & --- & --- & --- & Albanian & 1 \\
        Tsonga & tso\_Latn & --- & --- & --- & --- & tso\_Latn & --- & --- & --- & --- & --- & --- & 1 \\
        Tswana & tsn\_Latn & --- & tsn\_Latn & --- & --- & tsn\_Latn & --- & --- & --- & --- & --- & --- & 2 \\
        Turkish & tur\_Latn & tr\_TR & tur\_Latn & --- & --- & tur\_Latn & --- & --- & --- & --- & --- & Turkish & 4 \\
        Twi & twi\_Latn & --- & --- & twi & twi & --- & --- & --- & --- & --- & --- & --- & 1 \\
        Ukrainian & ukr\_Cyrl & uk\_UA & ukr\_Cyrl & --- & --- & ukr\_Cyrl & --- & uk & --- & --- & uk & Ukrainian & 3 \\
        Urdu & urd\_Arab & ur\_PK & urd\_Arab & --- & --- & urd\_Arab & --- & --- & --- & --- & --- & Urdu & 3 \\
        Vietnamese & vie\_Latn & vi\_VN & vie\_Latn & --- & --- & vie\_Latn & --- & vi & --- & vi & vi & Vietnamese & 4 \\
        Waray & war\_Latn & --- & --- & --- & --- & war\_Latn & --- & --- & --- & --- & --- & --- & 1 \\
        Welsh & cym\_Latn & --- & cym\_Latn & --- & --- & --- & --- & --- & --- & --- & --- & --- & 1 \\
        West Central Oromo & gaz\_Latn & --- & orm\_Ethi & orm & orm & gaz\_Latn & --- & --- & --- & --- & --- & --- & 1 \\
        Western Persian & pes\_Arab & fa\_IR & --- & --- & --- & pes\_Arab & --- & --- & --- & --- & --- & Persian & 4 \\
        Wolof & wol\_Latn & --- & wol\_Latn & wol & wol & wol\_Latn & --- & --- & --- & --- & --- & --- & 2 \\
        Xhosa & xho\_Latn & --- & xho\_Latn & xho & xho & xho\_Latn & --- & --- & --- & --- & --- & --- & 2 \\
        Yorùbá & yor\_Latn & --- & yor\_Latn & yor & yor & yor\_Latn & yo & --- & --- & --- & --- & --- & 2 \\
        Zulu & zul\_Latn & zu\_ZA & zul\_Latn & zul & zul & zul\_Latn & --- & --- & --- & --- & --- & --- & 2 \\ \hline
        Count & 120 & 51 & 88 & 18 & 18 & 116 & 15 & 30 & 10 & 7 & 31 & 44 & N/A \\ \hline

    \end{longtable}
    }
\end{landscape}

\begin{landscape}
\section{Raw Performance Metrics Scores of Phi-4}
\setlength{\tabcolsep}{6pt}
\begin{scriptsize} 
\begin{longtable}{lcccccccccccccccc}
        \caption{Raw Performance Metrics Scores of Phi-4.}
        \label{tab:phi_raw_scores} \\
\toprule
        Lang. & AfriMMLU & AfriXNLI & Belebele & Global MMLU & HellaSwag & TruthfulQA & MGSM & MLQA & INCLUDE & BLEU & ChrF++ & ROUGE-L & METEOR & xCOMET & SSA-COMET & MetricX \\ 
\hline
        acm & --- & --- & 0.63 & --- & --- & --- & --- & --- & --- & 19.13 & 39.97 & 0.15 & 0.26 & 0.62 & 0.48 & 18.18 \\ 
        afr & --- & --- & 0.78 & --- & --- & --- & --- & --- & --- & 31.70 & 57.58 & 0.59 & 0.56 & 0.82 & 0.59 & 20.40 \\ 
        als & --- & --- & 0.73 & --- & --- & --- & --- & --- & 0.68 & 21.01 & 47.23 & 0.46 & 0.43 & 0.60 & 0.56 & 14.92 \\ 
        amh & 0.30 & 0.40 & 0.29 & --- & --- & --- & --- & --- & --- & 1.11 & 10.82 & 0.15 & 0.04 & 0.22 & 0.08 & 3.25 \\ 
        apc & --- & --- & 0.69 & --- & --- & --- & --- & --- & --- & 19.81 & 41.05 & 0.16 & 0.28 & 0.64 & 0.49 & 18.52 \\ 
        arb & --- & --- & 0.87 & 0.61 & 0.58 & 0.55 & --- & 0.00 & 0.60 & 30.06 & 49.75 & 0.15 & 0.41 & 0.81 & 0.57 & 21.19 \\ 
        ars & --- & --- & 0.70 & --- & --- & --- & --- & --- & --- & 23.16 & 42.74 & 0.10 & 0.32 & 0.68 & 0.51 & 18.77 \\ 
        ary & --- & --- & 0.55 & --- & --- & --- & --- & --- & --- & 12.43 & 33.25 & 0.16 & 0.19 & 0.43 & 0.42 & 16.33 \\ 
        arz & --- & --- & 0.69 & --- & --- & --- & --- & --- & --- & 18.90 & 39.29 & 0.14 & 0.27 & 0.67 & 0.51 & 18.96 \\ 
        asm & --- & --- & 0.53 & --- & --- & --- & --- & --- & --- & 4.24 & 22.58 & 0.02 & 0.08 & 0.32 & 0.45 & 16.72 \\ 
        azj & --- & --- & 0.59 & --- & --- & --- & --- & --- & 0.41 & 6.88 & 30.98 & 0.24 & 0.19 & 0.33 & 0.43 & 9.46 \\ 
        bam & --- & --- & 0.32 & --- & --- & --- & --- & --- & --- & 2.27 & 17.57 & 0.14 & 0.10 & 0.23 & 0.27 & 5.85 \\ 
        bel & --- & --- & --- & --- & --- & --- & --- & --- & 0.29 & 13.03 & 33.97 & 0.17 & 0.24 & 0.52 & 0.49 & 13.92 \\ 
        ben & --- & --- & 0.68 & 0.49 & 0.40 & 0.49 & 0.02 & --- & 0.46 & 18.44 & 40.55 & 0.03 & 0.28 & 0.51 & 0.51 & 19.24 \\ 
        bod & --- & --- & 0.29 & --- & --- & --- & --- & --- & --- & 0.54 & 28.16 & 0.08 & 0.02 & 0.34 & 0.24 & 12.57 \\ 
        bul & --- & --- & 0.83 & --- & --- & --- & --- & --- & 0.67 & 32.22 & 55.31 & 0.19 & 0.52 & 0.80 & 0.55 & 20.23 \\ 
        cat & --- & --- & 0.82 & --- & 0.62 & 0.54 & --- & --- & --- & 36.40 & 59.75 & 0.58 & 0.60 & 0.86 & 0.58 & 20.56 \\ 
        ceb & --- & --- & 0.57 & --- & --- & --- & --- & --- & --- & 13.08 & 40.84 & 0.37 & 0.36 & 0.23 & 0.38 & 10.99 \\ 
        ces & --- & --- & 0.85 & --- & --- & --- & --- & --- & --- & 32.04 & 52.84 & 0.53 & 0.50 & 0.85 & 0.62 & 19.55 \\ 
        ckb & --- & --- & 0.44 & --- & --- & --- & --- & --- & --- & 1.63 & 22.43 & 0.05 & 0.10 & 0.22 & 0.34 & 8.40 \\ 
        dan & --- & --- & 0.85 & --- & 0.65 & 0.55 & --- & --- & --- & 19.24 & 43.11 & 0.41 & 0.42 & 0.49 & 0.51 & 14.06 \\ 
        deu & --- & --- & 0.83 & 0.74 & 0.69 & 0.57 & 0.30 & 0.03 & 0.55 & 42.01 & 64.08 & 0.63 & 0.65 & 0.91 & 0.63 & 22.09 \\ 
        ell & --- & --- & 0.83 & --- & --- & --- & --- & --- & 0.42 & 43.04 & 63.89 & 0.60 & 0.62 & 0.97 & 0.63 & 24.06 \\ 
        est & --- & --- & 0.67 & --- & --- & --- & --- & --- & 0.44 & 23.14 & 43.64 & 0.19 & 0.40 & 0.69 & 0.54 & 17.90 \\ 
        eus & --- & --- & 0.59 & --- & 0.32 & 0.42 & --- & --- & 0.37 & 10.50 & 39.50 & 0.29 & 0.29 & 0.39 & 0.47 & 8.11 \\ 
        ewe & 0.30 & 0.34 & --- & --- & --- & --- & --- & --- & --- & 8.63 & 39.51 & 0.21 & 0.23 & 0.33 & 0.44 & 9.70 \\ 
        fin & --- & --- & 0.83 & --- & --- & --- & --- & --- & 0.45 & 1.81 & 16.42 & 0.11 & 0.09 & 0.22 & 0.31 & 2.59 \\ 
        fra & 0.65 & 0.45 & 0.85 & 0.72 & 0.73 & 0.55 & 0.36 & --- & 0.69 & 23.15 & 51.38 & 0.41 & 0.40 & 0.80 & 0.62 & 18.53 \\ 
        fuv & --- & --- & 0.28 & --- & --- & --- & --- & --- & --- & 52.35 & 69.61 & 0.70 & 0.69 & 0.93 & 0.63 & 22.76 \\ 
        gaz & 0.36 & 0.38 & 0.31 & --- & --- & --- & --- & --- & --- & 2.10 & 19.55 & 0.07 & 0.09 & 0.22 & 0.17 & 9.14 \\ 
        grn & --- & --- & 0.35 & --- & --- & --- & --- & --- & --- & 1.15 & 22.16 & 0.04 & 0.07 & 0.23 & 0.27 & 3.37 \\ 
        guj & --- & --- & 0.63 & --- & 0.38 & 0.50 & --- & --- & --- & 2.50 & 20.50 & 0.12 & 0.12 & 0.23 & 0.34 & 6.76 \\ 
        hat & --- & --- & 0.49 & --- & --- & --- & --- & --- & --- & 14.51 & 33.49 & 0.15 & 0.29 & 0.58 & 0.56 & 17.71 \\ 
        hau & 0.33 & 0.39 & 0.38 & --- & --- & --- & --- & --- & --- & 6.30 & 31.62 & 0.26 & 0.26 & 0.27 & 0.44 & 7.52 \\ 
        heb & --- & --- & 0.79 & --- & --- & --- & --- & --- & 0.63 & 6.56 & 32.12 & 0.23 & 0.24 & 0.24 & 0.12 & 9.80 \\ 
        hin & --- & --- & 0.67 & 0.61 & 0.50 & 0.49 & --- & 0.01 & 0.61 & 27.47 & 47.77 & 0.19 & 0.42 & 0.75 & 0.50 & 19.87 \\ 
        hrv & --- & --- & 0.81 & --- & 0.56 & 0.54 & --- & --- & 0.75 & 28.40 & 49.73 & 0.18 & 0.47 & 0.66 & 0.57 & 21.49 \\ 
        hun & --- & --- & 0.80 & --- & 0.53 & 0.52 & --- & --- & 0.48 & 24.46 & 50.63 & 0.45 & 0.44 & 0.80 & 0.60 & 23.06 \\ 
        hye & --- & --- & 0.61 & --- & 0.30 & 0.45 & --- & --- & 0.35 & 24.71 & 49.85 & 0.46 & 0.42 & 0.78 & 0.58 & 19.21 \\ 
        ibo & 0.34 & 0.37 & 0.31 & --- & --- & --- & --- & --- & --- & 12.46 & 35.61 & 0.17 & 0.21 & 0.36 & 0.44 & 11.28 \\ 
        ilo & --- & --- & 0.40 & --- & --- & --- & --- & --- & --- & 3.07 & 19.41 & 0.16 & 0.13 & 0.21 & 0.16 & 1.80 \\ 
        ind & --- & --- & 0.84 & 0.62 & 0.65 & 0.55 & --- & --- & 0.64 & 5.63 & 32.96 & 0.24 & 0.24 & 0.22 & 0.32 & 9.36 \\ 
        isl & --- & --- & 0.72 & --- & --- & --- & --- & --- & --- & 40.21 & 65.60 & 0.65 & 0.66 & 0.89 & 0.63 & 22.57 \\ 
        ita & --- & --- & 0.83 & 0.77 & 0.71 & 0.54 & --- & --- & 0.76 & 12.15 & 35.51 & 0.32 & 0.31 & 0.48 & 0.43 & 11.19 \\ 
        jav & --- & --- & 0.66 & --- & --- & --- & --- & --- & --- & 35.08 & 58.44 & 0.54 & 0.56 & 0.94 & 0.62 & 23.10 \\ 
        jpn & --- & --- & 0.79 & 0.67 & --- & --- & 0.29 & --- & 0.75 & 15.05 & 43.92 & 0.39 & 0.39 & 0.45 & 0.45 & 13.76 \\ 
        kac & --- & --- & 0.32 & --- & --- & --- & --- & --- & --- & 29.27 & 40.52 & 0.27 & 0.00 & 0.89 & 0.64 & 21.75 \\ 
        kan & --- & --- & 0.67 & --- & 0.35 & 0.50 & --- & --- & --- & 0.43 & 1.43 & 0.02 & 0.01 & 0.21 & 0.20 & 7.02 \\ 
        kat & --- & --- & 0.48 & --- & --- & --- & --- & --- & 0.39 & 12.59 & 35.62 & 0.17 & 0.21 & 0.41 & 0.50 & 15.84 \\ 
        kaz & --- & --- & 0.58 & --- & --- & --- & --- & --- & 0.41 & 8.59 & 31.46 & 0.16 & 0.16 & 0.23 & 0.29 & 6.86 \\ 
        kea & --- & --- & 0.46 & --- & --- & --- & --- & --- & --- & 8.90 & 34.91 & 0.18 & 0.21 & 0.37 & 0.47 & 11.13 \\ 
        khk & --- & --- & 0.47 & --- & --- & --- & --- & --- & --- & 5.04 & 29.82 & 0.15 & 0.16 & 0.41 & 0.47 & 14.65 \\ 
        khm & --- & --- & 0.47 & --- & --- & --- & --- & --- & --- & 3.90 & 27.63 & 0.16 & 0.14 & 0.25 & 0.36 & 7.65 \\ 
        kin & 0.31 & 0.34 & 0.36 & --- & --- & --- & --- & --- & --- & 3.17 & 23.28 & 0.10 & 0.03 & 0.21 & 0.27 & 6.49 \\ 
        kir & --- & --- & 0.57 & --- & --- & --- & --- & --- & --- & 2.76 & 25.09 & 0.10 & 0.12 & 0.24 & 0.24 & 1.36 \\ 
        kor & --- & --- & 0.86 & 0.65 & --- & --- & --- & --- & 0.58 & 7.47 & 31.81 & 0.18 & 0.18 & 0.32 & 0.44 & 10.85 \\ 
        lao & --- & --- & 0.33 & --- & --- & --- & --- & --- & --- & 21.21 & 31.50 & 0.22 & 0.31 & 0.82 & 0.61 & 21.23 \\ 
        lin & 0.36 & 0.33 & 0.33 & --- & --- & --- & --- & --- & --- & 1.80 & 20.57 & 0.17 & 0.17 & 0.20 & 0.21 & 2.79 \\ 
        lit & --- & --- & 0.77 & --- & --- & --- & --- & --- & 0.51 & 3.19 & 23.62 & 0.16 & 0.16 & 0.23 & 0.23 & 5.48 \\ 
        lug & 0.29 & 0.37 & 0.33 & --- & --- & --- & --- & --- & --- & 15.06 & 43.39 & 0.33 & 0.34 & 0.56 & 0.54 & 13.58 \\ 
        luo & --- & --- & 0.33 & --- & --- & --- & --- & --- & --- & 2.58 & 24.92 & 0.09 & 0.12 & 0.23 & 0.26 & 4.84 \\ 
        lvs & --- & --- & 0.74 & --- & --- & --- & --- & --- & --- & 2.73 & 19.99 & 0.06 & 0.09 & 0.23 & 0.34 & 6.85 \\ 
        mal & --- & --- & 0.64 & --- & 0.32 & 0.48 & --- & --- & 0.38 & 15.09 & 42.17 & 0.36 & 0.35 & 0.46 & 0.50 & 10.15 \\ 
        mar & --- & --- & 0.62 & --- & 0.37 & 0.48 & --- & --- & --- & 11.29 & 33.62 & 0.15 & 0.18 & 0.42 & 0.46 & 14.83 \\ 
        mkd & --- & --- & 0.78 & --- & --- & --- & --- & --- & 0.80 & 13.38 & 38.63 & 0.06 & 0.27 & 0.31 & 0.48 & 18.57 \\ 
        mlt & --- & --- & 0.50 & --- & --- & --- & --- & --- & --- & 28.48 & 53.51 & 0.19 & 0.49 & 0.79 & 0.55 & 19.72 \\ 
        mri & --- & --- & 0.35 & --- & --- & --- & --- & --- & --- & 6.24 & 28.63 & 0.20 & 0.18 & 0.21 & 0.39 & 5.88 \\ 
        mya & --- & --- & 0.41 & --- & --- & --- & --- & --- & --- & 4.99 & 30.73 & 0.26 & 0.25 & 0.20 & 0.38 & 3.66 \\ 
        nld & --- & --- & 0.85 & --- & 0.66 & 0.57 & --- & --- & 0.73 & 2.30 & 29.38 & 0.04 & 0.02 & 0.22 & 0.31 & 6.43 \\ 
        nob & --- & --- & 0.85 & --- & --- & --- & --- & --- & --- & 29.65 & 55.18 & 0.49 & 0.51 & 0.91 & 0.62 & 22.92 \\ 
        npi & --- & --- & 0.56 & --- & 0.37 & 0.47 & --- & --- & 0.47 & 31.48 & 56.84 & 0.54 & 0.55 & 0.90 & 0.62 & 21.69 \\ 
        nso & --- & --- & 0.32 & --- & --- & --- & --- & --- & --- & 11.81 & 37.72 & 0.05 & 0.22 & 0.54 & 0.50 & 17.26 \\ 
        nya & --- & --- & 0.32 & --- & --- & --- & --- & --- & --- & 3.70 & 23.84 & 0.16 & 0.16 & 0.22 & 0.21 & 2.46 \\ 
        ory & --- & --- & 0.53 & --- & --- & --- & --- & --- & --- & 2.81 & 23.50 & 0.10 & 0.13 & 0.23 & 0.30 & 4.55 \\ 
        pan & --- & --- & 0.69 & --- & --- & --- & --- & --- & --- & 9.72 & 28.97 & 0.10 & 0.16 & 0.42 & 0.45 & 17.36 \\ 
        pbt & --- & --- & 0.50 & --- & --- & --- & --- & --- & --- & 18.22 & 35.90 & 0.18 & 0.33 & 0.47 & 0.52 & 17.17 \\ 
        pes & --- & --- & 0.79 & --- & --- & --- & --- & --- & 0.42 & 4.15 & 22.85 & 0.09 & 0.18 & 0.26 & 0.34 & 8.89 \\ 
        plt & --- & --- & 0.46 & --- & --- & --- & --- & --- & --- & 18.85 & 41.82 & 0.05 & 0.38 & 0.63 & 0.52 & 19.30 \\ 
        pol & --- & --- & 0.83 & --- & --- & --- & --- & --- & 0.50 & 4.38 & 32.60 & 0.25 & 0.20 & 0.24 & 0.26 & 5.99 \\ 
        por & --- & --- & 0.86 & 0.65 & 0.72 & 0.56 & --- & --- & 0.66 & 26.59 & 48.56 & 0.42 & 0.42 & 0.86 & 0.63 & 19.82 \\ 
        ron & --- & --- & 0.82 & --- & 0.57 & 0.52 & --- & --- & --- & 51.70 & 69.34 & 0.70 & 0.71 & 0.95 & 0.64 & 22.84 \\ 
        rus & --- & --- & 0.88 & --- & 0.66 & 0.56 & 0.02 & --- & 0.59 & 34.04 & 56.68 & 0.54 & 0.54 & 0.82 & 0.61 & 19.57 \\ 
        shn & --- & --- & 0.25 & --- & --- & --- & --- & --- & --- & 36.27 & 55.76 & 0.19 & 0.52 & 0.92 & 0.64 & 22.77 \\ 
        sin & --- & --- & 0.30 & --- & --- & --- & --- & --- & --- & 0.40 & 9.20 & 0.07 & 0.02 & 0.21 & 0.22 & 6.01 \\ 
        slk & --- & --- & 0.82 & --- & 0.54 & 0.50 & --- & --- & --- & 2.20 & 19.26 & 0.17 & 0.09 & 0.21 & 0.22 & 4.96 \\ 
        slv & --- & --- & 0.78 & --- & --- & --- & --- & --- & --- & 22.98 & 47.56 & 0.43 & 0.42 & 0.73 & 0.57 & 16.87 \\ 
        sna & 0.35 & 0.38 & 0.37 & --- & --- & --- & --- & --- & --- & 20.28 & 46.79 & 0.39 & 0.41 & 0.68 & 0.56 & 18.64 \\ 
        snd & --- & --- & 0.48 & --- & --- & --- & --- & --- & --- & 2.60 & 25.69 & 0.06 & 0.11 & 0.23 & 0.25 & 2.87 \\ 
        som & --- & --- & 0.31 & --- & --- & --- & --- & --- & --- & 3.49 & 21.56 & 0.16 & 0.13 & 0.24 & 0.33 & 9.47 \\ 
        sot & 0.32 & 0.36 & 0.32 & --- & --- & --- & --- & --- & --- & 2.96 & 26.37 & 0.11 & 0.13 & 0.20 & 0.14 & 2.85 \\ 
        spa & --- & --- & 0.83 & 0.73 & 0.74 & 0.54 & 0.39 & 0.02 & 0.72 & 3.39 & 25.34 & 0.17 & 0.17 & 0.23 & 0.21 & 2.46 \\ 
        srp & --- & --- & 0.80 & --- & 0.56 & 0.54 & --- & --- & 0.64 & 32.43 & 56.11 & 0.55 & 0.55 & 0.95 & 0.65 & 22.90 \\ 
        ssw & --- & --- & 0.35 & --- & --- & --- & --- & --- & --- & 8.18 & 13.24 & 0.06 & 0.17 & 0.74 & 0.60 & 22.09 \\ 
        sun & --- & --- & 0.49 & --- & --- & --- & --- & --- & --- & 2.82 & 24.25 & 0.05 & 0.10 & 0.22 & 0.14 & 3.75 \\ 
        swe & --- & --- & 0.87 & --- & 0.65 & 0.55 & --- & --- & --- & 6.42 & 31.96 & 0.17 & 0.20 & 0.40 & 0.41 & 12.21 \\ 
        swh & 0.44 & 0.42 & 0.68 & 0.45 & --- & --- & 0.00 & --- & --- & 42.48 & 64.04 & 0.63 & 0.63 & 0.90 & 0.63 & 22.06 \\ 
        tam & --- & --- & 0.65 & --- & 0.31 & 0.50 & --- & --- & 0.37 & 18.33 & 47.67 & 0.42 & 0.43 & 0.52 & 0.48 & 14.64 \\ 
        tel & --- & --- & 0.61 & --- & 0.34 & 0.48 & 0.02 & --- & 0.37 & 14.93 & 43.47 & 0.18 & 0.24 & 0.38 & 0.56 & 17.29 \\ 
        tgk & --- & --- & 0.43 & --- & --- & --- & --- & --- & --- & 15.54 & 39.56 & 0.19 & 0.26 & 0.46 & 0.48 & 16.25 \\ 
        tgl & --- & --- & 0.70 & --- & --- & --- & --- & --- & 0.67 & 4.10 & 25.69 & 0.15 & 0.16 & 0.19 & 0.39 & 5.89 \\ 
        tha & --- & --- & 0.74 & --- & --- & --- & 0.00 & --- & --- & 21.96 & 51.53 & 0.49 & 0.49 & 0.60 & 0.47 & 17.85 \\ 
        tir & --- & --- & 0.30 & --- & --- & --- & --- & --- & --- & 26.70 & 44.15 & 0.19 & 0.09 & 0.68 & 0.53 & 19.10 \\ 
        tsn & --- & --- & 0.33 & --- & --- & --- & --- & --- & --- & 0.62 & 8.47 & 0.12 & 0.01 & 0.21 & 0.05 & 2.44 \\ 
        tso & --- & --- & 0.38 & --- & --- & --- & --- & --- & --- & 2.76 & 24.29 & 0.16 & 0.16 & 0.22 & 0.21 & 2.79 \\ 
        tur & --- & --- & 0.81 & --- & --- & --- & --- & --- & 0.60 & 3.01 & 22.98 & 0.10 & 0.11 & 0.23 & 0.25 & 5.88 \\ 
        twi & 0.32 & 0.35 & --- & --- & --- & --- & --- & --- & --- & 28.82 & 53.90 & 0.47 & 0.45 & 0.79 & 0.61 & 19.36 \\ 
        ukr & --- & --- & 0.86 & --- & 0.61 & 0.55 & --- & --- & 0.66 & 2.72 & 20.64 & 0.16 & 0.13 & 0.22 & 0.20 & 3.55 \\ 
        urd & --- & --- & 0.71 & --- & --- & --- & --- & --- & 0.36 & 31.49 & 53.46 & 0.19 & 0.48 & 0.87 & 0.64 & 21.11 \\ 
        uzn & --- & --- & 0.53 & --- & --- & --- & --- & --- & 0.39 & 15.75 & 38.05 & 0.17 & 0.34 & 0.50 & 0.49 & 18.11 \\ 
        vie & --- & --- & 0.85 & --- & 0.60 & 0.53 & --- & 0.01 & 0.57 & 4.77 & 30.54 & 0.15 & 0.16 & 0.26 & 0.40 & 6.31 \\ 
        war & --- & --- & 0.49 & --- & --- & --- & --- & --- & --- & 35.19 & 53.72 & 0.65 & 0.63 & 0.84 & 0.60 & 21.70 \\ 
        wol & 0.31 & 0.34 & 0.26 & --- & --- & --- & --- & --- & --- & 7.48 & 34.32 & 0.21 & 0.21 & 0.23 & 0.36 & 11.26 \\ 
        xho & 0.29 & 0.38 & 0.35 & --- & --- & --- & --- & --- & --- & 1.96 & 16.69 & 0.10 & 0.11 & 0.22 & 0.36 & 5.57 \\ 
        yor & 0.39 & 0.42 & 0.33 & 0.31 & --- & --- & --- & --- & --- & 3.13 & 27.62 & 0.06 & 0.11 & 0.23 & 0.12 & 2.63 \\ 
        zho & --- & --- & 0.90 & 0.67 & --- & 0.54 & 0.06 & 0.01 & 0.57 & 3.35 & 19.30 & 0.25 & 0.14 & 0.22 & 0.30 & 9.32 \\ 
        zho\_trad & --- & --- & 0.88 & --- & --- & --- & --- & --- & --- & 32.83 & 35.12 & 0.19 & 0.03 & 0.86 & 0.61 & 23.01 \\ 
        zsm & --- & --- & 0.81 & --- & --- & --- & --- & --- & 0.63 & 26.02 & 29.15 & 0.18 & 0.00 & 0.83 & 0.61 & 22.70 \\ 
        zul & 0.34 & 0.37 & 0.35 & --- & --- & --- & --- & --- & --- & 25.75 & 56.30 & 0.54 & 0.54 & 0.80 & 0.59 & 20.61 \\ 
        ~ & --- & --- & --- & --- & --- & --- & --- & --- & --- & 3.31 & 28.55 & 0.07 & 0.11 & 0.23 & 0.12 & 2.21 \\ 
\bottomrule

\end{longtable}
\end{scriptsize}
\end{landscape}
\label{tab:raw_scores}

\section{Multilingual Abstract}
\label{app:translations}
Here we present the translation of abstract in 10 different langagues, generated using Mansa and ChatGPT5.


\section*{Supplementary material}

## 1. Chinese (Simplified)

大型语言模型（LLMs）的迅速扩散带来了一个关键的评估悖论：尽管LLMs宣称具备多语言能力，但目前存在的、非机器翻译构建的全面基准测试覆盖的语言不到30种，使全球约7000种语言中的98%以上处于实证空白之中。传统基准构建面临诸如成本高、领域专家稀缺以及数据污染等扩展性挑战。我们评估了一种更简单替代方案的有效性：仅凭翻译质量是否可以指示模型更广泛的多语言能力？通过在9个多样化基准和7种翻译指标上系统评估14个模型（参数规模1B–72B），我们发现翻译性能是下游任务成功的良好指标（例如 Phi-4 的中位数皮尔逊相关系数：METRICX = 0.89，XCOMET = 0.91，SSA-COMET = 0.87）。这些结果表明，支持忠实翻译的表征能力与多语言理解所需的能力高度重叠。因此，翻译质量可作为一种强大且低成本的多语言性能初筛指标，并辅以针对具体任务的后续评估。

## 2. Spanish

La rápida proliferación de los LLM ha generado una paradoja crítica de evaluación: aunque los LLM afirman poseer competencia multilingüe, existen menos de 30 idiomas con conjuntos de evaluación amplios que no estén traducidos automáticamente, dejando a más del 98% de las 7 000 lenguas del mundo en un vacío empírico. La construcción tradicional de benchmarks enfrenta problemas de escalabilidad como el costo, la escasez de expertos de dominio y la contaminación de datos. Evaluamos la validez de una alternativa más simple: ¿puede la calidad de la traducción por sí sola indicar las capacidades multilingües más amplias de un modelo? Mediante una evaluación sistemática de 14 modelos (1B–72B parámetros) en 9 benchmarks diversos y 7 métricas de traducción, encontramos que el desempeño en traducción es un buen indicador del éxito en tareas posteriores (por ejemplo, Phi-4, mediana de correlación de Pearson: METRICX = 0.89, XCOMET = 0.91, SSA-COMET = 0.87). Estos resultados sugieren que las capacidades representacionales que sustentan una traducción fiel se superponen con las necesarias para la comprensión multilingüe. Así, la calidad de traducción emerge como un proxy fuerte y de bajo costo para el desempeño multilingüe, permitiendo un filtrado inicial basado en traducción con seguimientos dirigidos a tareas específicas.

## 3. Hausa

Yaɗuwar LLMs cikin sauri ya haifar da matsalar tantancewa mai mahimmanci: yayin da LLMs ke da'awar iya yare da yawa, akwai ma'auni na gama gari da ba na fassarar injin ba don yaruka ƙasa da 30 kawai, yana barin fiye da kashi 98% na yaruka 7,000 na duniya a cikin babu tabbaci na gwaji. Gina ma'auni na gargajiya yana fuskantar ƙalubalen haɓakawa kamar farashi, ƙarancin ƙwararrun fannin, da gurɓataccen bayanai. Muna tantance ingancin zaɓi mafi sauƙi: shin ingancin fassara kaɗai zai iya nuna iyawar ƙirar harsuna da yawa?

Ta hanyar tantancewa na tsari na samfura 14 (1B–72B parameters) a cikin ma'auni daban-daban 9 da ma'aunin fassara 7, mun gano cewa aikin fassara alama ce mai kyau ta nasarar aiki na ƙasa (misali, Phi-4, median Pearson r: METRICX = 0.89, XCOMET = 0.91, SSA-COMET = 0.87). Waɗannan sakamakon suna nuna cewa iyawar wakilci da ke goyan bayan fassara ta aminci tana haɗuwa da waɗanda ake buƙata don fahimtar yaruka da yawa. Ingancin fassara don haka ya bayyana a matsayin wakili mai ƙarfi, mai arha na farko don aikin yaruka da yawa, yana ba da damar tantancewar fassara-ta-farko tare da bin daidai don takamaiman ayyuka.

## 4. Amharic

የLLMዎች ፈጣን መስፋፋት ወሳኝ የግምገማ ተቃርኖ ፈጥሯል፦ LLMዎች ብዙ ቋንቋ ብቃት እንዳላቸው ቢገልጹም፣ ሁሉ አቀፍ በማሽን-ያልተተረጎሙ መለኪያዎች ከ30 ያነሱ ቋንቋዎች ብቻ አሏቸው፣ ይህም ከዓለም 7,000 ቋንቋዎች ከ98% በላይ የሚሆኑትን በተጨባጭ ባዶነት ውስጥ ይተዋል። ባህላዊ መለኪያ ግንባታ እንዲ ወጪ፣ የባለሙያዎች እጥረት እና የመረጃ መበከል ያሉ የመጠን ፈተናዎች ይገጥሙታል። የቀለል አማራጭ ትክክለኛነት እንገመግማለን፦ የትርጉም ጥራት ብቻ የሞዴል ሰፊ ብዙ ቋንቋ ችሎታዎችን ሊያመለክት ይችላል?

14 ሞዴሎችን (1B–72B ፓራሜትሮች) በ9 የተለያዩ መለኪያዎች እና 7 የትርጉም መለኪያዎች ላይ በስርዓታዊ ግምገማ፣ የትርጉም አፈጻጸም የታችኛው ተግባር ስኬት ጥሩ አመልካች መሆኑን አግኝተናል (ለምሳሌ፣ Phi-4፣ መካከለኛ Pearson r: METRICX = 0.89, XCOMET = 0.91, SSA-COMET = 0.87)። እነዚህ ውጤቶች ታማኝ ትርጉምን የሚደግፉ የውክልና ችሎታዎች ለብዙ ቋንቋ ግንዛቤ ከሚያስፈልጉት ጋር መደራረባቸውን ይጠቁማሉ። ስለዚህ የትርጉም ጥራት እንደ ጠንካራ፣ ርካሽ የመጀመሪያ-ደረጃ ተወካይ ለብዙ ቋንቋ አቅም ይወጣል፣ ይህም ለተወሰኑ ተግባራት ያነጣጠረ ክትትል ያለው ትርጉም-መጀመሪያ ምርምራን ያስችላል።

## 5. Basque

LLMen hedapen azkarrak ebaluazio-paradoxa kritiko bat sortu du: LLMek gaitasun eleaniztuna dutela aldarrikatzen duten arren, 30 hizkuntza baino gutxiagorentzat baino ez daude automatikoki itzuli gabeko benchmark zabalak, eta horrek munduko 7.000 hizkuntzen %98 baino gehiago hutsune enpirikoan uzten ditu. Benchmark tradizionalen eraikuntzak eskalagarritasun-arazoak ditu, hala nola kostua, aditu faltak eta datuen kutsadura. Alternatiba sinpleago baten balioa ebaluatzen dugu: itzulpenaren kalitateak bakarrik adieraz al ditzake eredu baten gaitasun eleaniztun zabalagoak? 14 eredu (1B–72B parametro) 9 benchmarketan eta 7 itzulpen-neurritan ebaluatuz, aurkitu dugu itzulpen-errendimendua downstream zereginen arrakastaren adierazle ona dela. Horrek iradokitzen du itzulpen fidela ahalbidetzen duten gaitasunak eleaniztasun-ulermenarekin gainjartzen direla.

## 6. Bengali

LLM-গুলোর দ্রুত বিস্তার একটি গুরুত্বপূর্ণ মূল্যায়ন-বিরোধ সৃষ্টি করেছে: যদিও LLM-গুলো বহুভাষিক দক্ষতার দাবি করে, ৩০টিরও কম ভাষার জন্য এমন বিস্তৃত বেঞ্চমার্ক আছে যা মেশিন-অনুবাদিত নয়, ফলে বিশ্বের প্রায় ৭,০০০ ভাষার ৯৮%-এর বেশি পরীক্ষামূলক শূন্যতায় রয়ে গেছে। প্রচলিত বেঞ্চমার্ক নির্মাণ ব্যয়, বিশেষজ্ঞের অভাব এবং ডেটা দূষণের মতো স্কেলিং সমস্যার সম্মুখীন হয়। আমরা একটি সহজ বিকল্পের বৈধতা মূল্যায়ন করি: শুধুমাত্র অনুবাদের মান কি একটি মডেলের বিস্তৃত বহুভাষিক সক্ষমতার সূচক হতে পারে? ১৪টি মডেল (১B–৭২B প্যারামিটার) ৯টি বেঞ্চমার্ক ও ৭টি অনুবাদ মেট্রিক জুড়ে মূল্যায়নের মাধ্যমে আমরা পাই যে অনুবাদ কর্মক্ষমতা পরবর্তী কাজের সাফল্যের একটি ভালো সূচক। অতএব অনুবাদের মান একটি শক্তিশালী ও সাশ্রয়ী প্রথম-ধাপের বহুভাষিক কর্মক্ষমতা নির্দেশক।

## 7. Georgian

LLM-ების სწრაფმა გავრცელებამ მნიშვნელოვანი შეფასების პარადოქსი შექმნა: მიუხედავად იმისა, რომ LLM-ები მრავალენოვან უნარებს აცხადებენ, ავტომატურად თარგმნილი olmayan სრულფასოვანი ბენჩმარკები არსებობს 30-ზე ნაკლებ ენაზე, რის შედეგადაც მსოფლიოს 7,000 ენის 98%-ზე მეტი ემპირიულ სიცარიელეში რჩება. ტრადიციული ბენჩმარკების აგება აწყდება მასშტაბირების პრობლემებს, როგორიცაა ღირებულება, ექსპერტების სიმცირე და მონაცემთა დაბინძურება. ჩვენ ვაფასებთ უფრო მარტივი ალტერნატივის ვალიდობას: შეუძლია თუ არა მხოლოდ თარგმნის ხარისხს მიუთითოს მოდელის უფრო ფართო მრავალენოვან შესაძლებლობებზე? 14 მოდელის (1B–72B პარამეტრი) 9 ბენჩმარკსა და 7 თარგმანის მეტრიკაში შეფასებით ვპოულობთ, რომ თარგმანის შესრულება კარგი ინდიკატორია შემდგომი ამოცანების წარმატებისთვის.

## 8. Japanese

LLMの急速な普及は重要な評価パラドックスを生み出している。LLMは多言語能力を主張している一方で、機械翻訳に依存しない包括的なベンチマークが存在する言語は30未満であり、世界の約7,000言語の98%以上が実証的空白に置かれている。従来のベンチマーク構築は、コスト、専門家不足、データ汚染といったスケーリング上の課題に直面している。本研究ではより単純な代替案の妥当性を検証する。すなわち、翻訳品質だけでモデルの広範な多言語能力を示せるのか。14モデル（1B〜72B）を9種のベンチマークと7種の翻訳指標で体系的に評価した結果、翻訳性能は下流タスク成功の良い指標であることが分かった。したがって翻訳品質は低コストで有効な第一段階の代理指標となる。

## 9. Guarani

LLM-kuéra oñemombarete pya'e rupi ohechauka peteĩ jeikuaa-paradójha: LLM-kuéra he'i oguerekoha katupyry heta ñe'ẽme, péro oĩ mbovymi (sa'i 30-gui) ñe'ẽ oguerekóva hag̃ua mba'eha'ãrõ tuicháva ndaha'éiva traducción automática, ha upéicha 98% rupi umi 7.000 ñe'ẽ yvypórape oĩ hag̃ua vacío empírico-pe. Umi benchmark tradicional ojuhúma apañuãi tuicha hag̃ua hag̃ua hag̃ua, techapyrãrõ jehepyme'ẽ, experto'ỹre, ha datos oñembyaíva. Rohecha hag̃ua peteĩ alternativa isãsovéva: ikatúpa traducción rekóinte ohechauka modelo katupyry tuichavéva? Rohesa'ỹijo 14 modelo (1B–72B parámetro) 9 benchmark ha 7 métrica traducción rehe, ha rotopa traducción apopyre ha'eha indicador porã umi tembiapo oútavape g̃uarã.

## 10. Zulu

Ukunyuka okusheshayo kwama-LLM kudale iphuzu eliyinkinga yokuhlola: nakuba ama-LLM ethi ayakwazi izilimi eziningi, izilinganiso ezibanzi ezingahunyushwanga umshini zikhona kuphela ezilimini ezingaphansi kuka-30, okushiya ngaphezu kuka-98% yezilimi eziyizi-7,000 zomhlaba zingena bufakazi obubonakalayo. Ukwakhiwa kwezilinganiso kwendabuko kubhekene nezinselelo zokwandisa ezifana nezindleko, ukuswela izazi ezikhethekile, kanye nokungcoliswa kwedatha. Sihlola ubuqiniso benye indlela elula: ingabe ikhwalithi yokuhumusha yodwa ingakhombisa amakhono alolu hlobo olukhulu lwezilimi eziningi zomodeli?

Ngokuhlola ngendlela ehlelekile amamodeli angu-14 (1B–72B parameters) ezilinganisweni eziyisishiyagalolunye ezihlukahlukene nezilinganiso zokuhumusha eziyisikhombisa, sithole ukuthi

ukusebenza ekuhumusheni kuyisikhombi esihle sempumelelo yemisebenzi elandelayo (isb., Phi-4, i-median Pearson r: METRICX = 0.89, XCOMET = 0.91, SSA-COMET = 0.87). Le miphumela ikhombisa ukuthi amakhono okumela asiza ekuhumusheni okuthembekile ahlangana nalawo adingekayo ekuqondeni izilimi eziningi. Ngakho-ke ikhwalithi yokuhumusha ivelela njengesikhombi sokuqala esiqinile, esingabizi kakhulu sokusebenza ngezilimi eziningi, okuvumela ukuhlolwa kokuqala ngokuhumusha nokulandelwa okuqondiswe emisebenzini ethile.



\end{document}